\documentclass[twocolumn,journal]{IEEEtran}
\usepackage[T1]{fontenc}
\usepackage[latin9]{inputenc}
\usepackage{array}
\usepackage{mathtools}
\usepackage{multirow}
\usepackage{amsmath}
\usepackage{amssymb}
\usepackage{stackrel}
\usepackage{graphicx}
\usepackage[unicode=true,
 bookmarks=true,bookmarksnumbered=true,bookmarksopen=true,bookmarksopenlevel=1,
 breaklinks=false,pdfborder={0 0 0},pdfborderstyle={},backref=false,colorlinks=false]
 {hyperref}
\hypersetup{pdftitle={Your Title},
 pdfauthor={Your Name},
 pdfpagelayout=OneColumn, pdfnewwindow=true, pdfstartview=XYZ, plainpages=false}

\makeatletter

\providecommand{\tabularnewline}{\\}

\let\oldforeign@language\foreign@language
\DeclareRobustCommand{\foreign@language}[1]{%
  \lowercase{\oldforeign@language{#1}}}

\usepackage[caption=false,font=footnotesize]{subfig}
\usepackage[linesnumbered, ruled]{algorithm2e}
\usepackage{booktabs}
\usepackage{mathrsfs}

\@ifundefined{showcaptionsetup}{}{%
 \PassOptionsToPackage{caption=false}{subfig}}
\usepackage{subfig}
\makeatother

\begin{document}
\title{\textbf{Graph Convolutional Subspace Clustering: A Robust Subspace
Clustering Framework for Hyperspectral Image}}
\author{Yaoming~Cai,~\IEEEmembership{Student Member,~IEEE,} Zijia Zhang,~Zhihua~Cai,~Xiaobo~Liu,~\IEEEmembership{Member,~IEEE,}
Xinwei Jiang,~and~Qin Yan \thanks{This work was supported in part by the National Natural Science Foundation
of China (NSFC) under Grant 61973285, 61773355 and 61603355, in part
by the Fundamental Research Founds for National University, China
University of Geosciences(Wuhan) under Grant G1323541717 and 1910491T06,
and in part by the National Nature Science Foundation of Hubei Province
under Grant 2018CFB528. \emph{(Corresponding author: Z. Cai.)}}\thanks{Y. Cai, Z. Zhang, Z. Cai, X. Jiang, and Q. Yan are with the School
of Computer Science, China University of Geosciences, Wuhan 430074,
China (e-mail: caiyaom@cug.edu.cn; zhangzijia@cug.edu.cn; zhcai@cug.edu.cn;
ysjxw@hotmail.com; yanqin@cug.edu.cn).}\thanks{X. Liu is with the School of Automation, and also with the Hubei Key
Laboratory of Advanced Control and Intelligent Automation for Complex
Systems, China University of Geosciences, Wuhan 430074, China (e-mail:
xbliu@cug.edu.cn). }}
\markboth{IEEE Transactions on XXX}{}
\maketitle
\begin{abstract}
Hyperspectral image (HSI) clustering is a challenging task due to
the high complexity of HSI data. Subspace clustering has been proven
to be powerful for exploiting the intrinsic relationship between data
points. Despite the impressive performance in the HSI clustering,
traditional subspace clustering methods often ignore the inherent
structural information among data. In this paper, we revisit the subspace
clustering with graph convolution and present a novel subspace clustering
framework called Graph Convolutional Subspace Clustering (GCSC) for
robust HSI clustering. Specifically, the framework recasts the self-expressiveness
property of the data into the non-Euclidean domain, which results
in a more robust graph embedding dictionary. We show that traditional
subspace clustering models are the special forms of our framework
with the Euclidean data. Basing on the framework, we further propose
two novel subspace clustering models by using the Frobenius norm,
namely Efficient GCSC (EGCSC) and Efficient Kernel GCSC (EKGCSC).
Both models have a globally optimal closed-form solution, which makes
them easier to implement, train, and apply in practice. Extensive
experiments on three popular HSI datasets demonstrate that EGCSC and
EKGCSC can achieve state-of-the-art clustering performance and dramatically
outperforms many existing methods with significant margins. 
\end{abstract}

\begin{IEEEkeywords}
Hyperspectral Image Clustering, Graph Convolutional Networks, Subspace
Clustering, Kernel Method 
\end{IEEEkeywords}

\section{Introduction}

\IEEEPARstart{H}{yperspectral} images (HSIs) acquired by remote sensors
contain rich spectral and spatial information, which enables us to
accurately recognize the region of interest. Over the past decade,
HSIs have been widely applied to various fields, ranging from geological
exploration, marine monitoring, military reconnaissance to medical
imaging and forensics \cite{HSI-Advances-in-Hyperspectral-Image-Signal-Processing-Overview-Ghamisi-GRSM-2017,BS-Net-TGRS-2020-Cai,HSIC-DeepForeast-TGRS-2019-XLiu}.

HSI classification, which aims to classify every pixel with a certain
label, is the foundation for the application of HSI \cite{HSIC-DeepForeast-TGRS-2019-XLiu,Cai-WHISPERS-2019}.
The most commonly used HSI classification method is the supervised
classification \cite{HSIC-New-Frontier-Ghamisi-IGRSM-2018,HSIC-Semi-GraphCNN-Qin-GRSL-2019}
based on label information. In recent years, the supervised HSI classification
has made great progress. For several popular HSI datasets, such as
Indian Pines, Salinas and Pavia University images \cite{HSI-Advances-in-Hyperspectral-Image-Signal-Processing-Overview-Ghamisi-GRSM-2017,HSI-Advances-in-Spectral-Spatial-Classification-IEEEProceedings-2013},
the supervised methods have achieved excellent classification accuracy.
Particularly, deep learning models \cite{HSIC-overview-Li-TGRS-2019,HSI-Advances-in-Hyperspectral-Image-Signal-Processing-Overview-Ghamisi-GRSM-2017,HSI-Recent-Advances-on-Spectral-Spatial-Classification-He-TGRS-2018,HE-ELM-CYM-PRL-2018},
such as Convolutional Neural Networks (CNNs) \cite{CNN-overview-GuJX-PR-2018,Deep-learning-LeCun-Nature-2015},
have extremely narrowed the gap between human and machine. Unfortunately,
the supervised method typically requires a large amount of labeled
data, which cannot be satisfied in HSI scenarios due to the high cost
of labeling training data. Furthermore, the supervised methods are
difficult to deal with unknown objects, since they are modeled by
the known classes. 
\begin{center}
\begin{figure}[t]
\begin{centering}
\includegraphics[width=1\columnwidth]{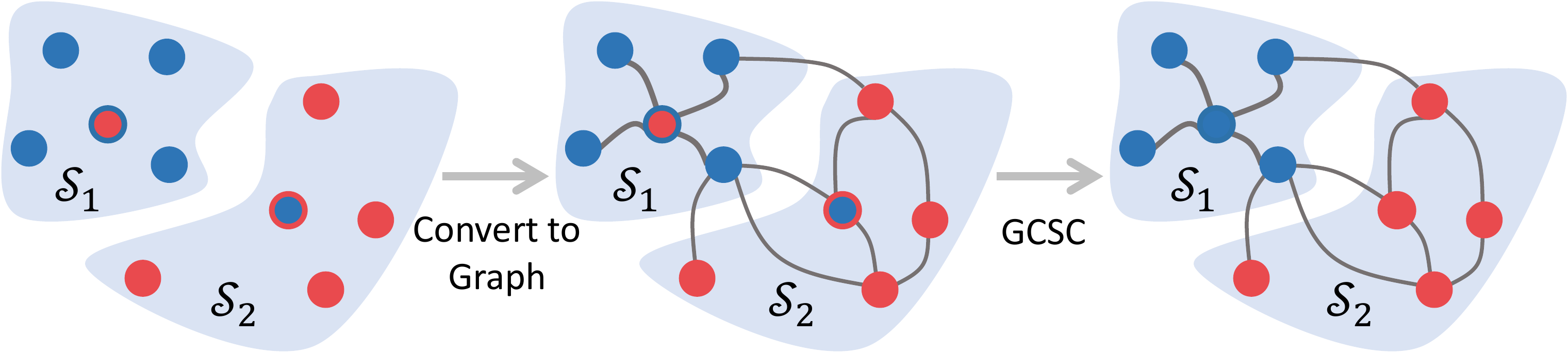}
\par\end{centering}
\caption{The motivation of GCSC. In the figure, the blue and red points signify
that two different classes that lie in the subspace $\mathcal{S}_{1}$
and $\mathcal{S}_{2}$, respectively. The red point with a blue outline
and the blue point with a red outline denote the misclassified points.
Intuitively, GCSC converts the traditional data into the graph-structured
data and adopts the graph convolution to generate the robust embedding
for the subsequent subspace clustering. \label{fig:The-motivation}}
\end{figure}
\par\end{center}

To avoid manual data annotation, many works have dedicated to developing
unsupervised HSI classification methods namely HSI clustering. Instead
of using the label information, the HSI clustering aims to find the
intrinsic relationship between data points and automatically determine
labels in an unsupervised manner \cite{HSI_BS-ISSC-SunWW-JSTARS-2015}.
The key to the HSI clustering is to measure the similarity between
data points \cite{Deep-Self-Evolution-Clustering-Chang-TPAMI-2018}.
Traditional clustering methods, e.g., K-means \cite{K-means-multi-view-Zhu-2019},
frequently use the pair-wise distance as the similarity measurement,
such as the Euclidean distance. Owing to the mixed pixel and the redundant
band problem \cite{HSI-band-selection-Hu-GRSL-2019,HSI-band-selection-Zeng-GRSL-2019},
these methods often suffer from unreliable measurement and making
the HSI clustering in great challenges. Compared with the supervised
classification, there are quite fewer studies on the HSI clustering
\cite{HSIClustering-RMMF-ZhangLF-IS-2019} and they are usually uncompetitive
in terms of accuracy. 

Recently, subspace clustering \cite{Subspace-Clustering-Vidal-ISPM-2011}
has drawn increasing attention in the HSI clustering \cite{HSI-Clustering-JSSC-Huang-ICIP-2018,HSI-Clustering-KSSC-Zhai-RS-2017,HSIClustering-L2SSC-ZhanH-GRSL-2017,HSIClustering-S4C-ZhangH-TGRS-2016,HSIClustering-SpectralClustering-GRSL-2017}
due to its ability to handle high-dimensional data and its reliable
performance. Technically, the subspace clustering seeks to express
the data points as a linear combination of a self-expressive dictionary
in the same subspace \cite{subspace-affinty-learning-Li-PR-2018}.
The subspace clustering model typically consists of two steps, i.e.,
self-representation \cite{DeepSC-JPan-NIPS2017} and Spectral Clustering
(SC) \cite{HSIClustering-SpectralClustering-GRSL-2017}. To improve
the performance of the subspace clustering, many works have devoted
to constructing a robust affinity matrix by using various techniques.
For example, Sparse Subspace Clustering (SSC) \cite{SSC-Elhamifar-TPAMI-2013}
uses an $\ell_{1}$-norm to encourage a sparse affinity matrix, while
Low Rank Subspace Clustering (LRSC) \cite{LRSC-1-Vidal-PRLetters-2014}
adopts a nuclear norm to enforce the affinity matrix to be low-rank.
By considering the spectral and spatial properties of HSIs, Zhang
et al. proposed a Spectral--Spatial Sparse Subspace Clustering (S$^{4}$C)
\cite{HSIClustering-S4C-ZhangH-TGRS-2016}. Kernel subspace clustering
\cite{KSC-Patel-ICIP-2014} was proposed as the nonlinear extension
of the subspace clustering model by implicitly mapping data into higher
kernel space. In \cite{HSI-Clustering-KSSC-Zhai-RS-2017}, an improved
kernel subspace clustering was applied to the HSI clustering.

However, the previous subspace clustering models are based on the
Euclidean data and often ignore the inherent graph structure contained
in the data points. On the one hand, the data points are usually corrupted
by noise or can have entries with large errors. On the other hand,
although manifold regularization is useful to incorporate graph information
into the subspace clustering, such as graph regularized LRSC \cite{HSI_BS-Laplacian-Regularized-LRSC-ZhaiH-TGRS-2018,Laplacian-Regularized-LRR-TPAMI-YinM-2016},
it usually needs to add an additional regularization term and a tradeoff
parameter. The recent development of Graph Neural Networks (GNNs)
\cite{GNN-revirew-zhou201-arXiv,GNN-Survey-Wu-TNNLS-2020,GNN-Survey-Zhang-TKDE-2020}
generalizes the powerful CNNs in dealing with the Euclidean data to
modeling the graph-structured data. This allows us to revisit traditional
problems with GNNs \cite{GCN-Deeperinsights-Li-AAAI-2018,gcn-kipf2017-iclr2017}.
However, the subspace clustering that combines graph learning has
not attracted too much attention.

To learn graph embedding and affinity, simultaneously, in this paper,
we present a Graph Convolutional Subspace Clustering (GCSC) framework
that recasts the traditional subspace clustering into the non-Euclidean
domain. Specifically, the GCSC framework calculates the self-representation
coefficients of the subspace clustering by leveraging a graph convolutional
self-representation model combining both graph and feature information.
As a result, the proposed framework can circumvent noise data and
tends to produce a more robust affinity than the traditional subspace
clustering models. Visually, an intuitive description about the motivation
of GCSC is illustrated in Fig. \ref{fig:The-motivation}. 

To sum up, the main contributions of this work are: 

\begin{enumerate} 
\item A robust subspace clustering framework called GCSC is developed for the HSI clustering in which the subspace clustering is recasted into the non-Euclidean domain. Particularly, the traditional subspace clustering can be viewed as the special form of the proposed framework.  

\item Based on the Frobenius norm, two novel and efficient subspace clustering models are proposed under the GCSC framework. We refer to them as Efficient GCSC (EGCSC) and Efficient Kernel GCSC (EKGCSC), respectively. Both EGCSC and EKGCSC have a closed-form solution, making them easier to implement, train, and apply in practice.

\item Our experimental results on several HSI datasets show that the proposed subspace clustering models are effectively better than many existing clustering methods for the HSI clustering. The successful attempt of GCSC offers an alternative orientation for unsupervised learning.

\end{enumerate}

The rest of the paper is structured as follows. We first briefly review
the subspace clustering, graph convolutional networks, and HSI clustering
in Section \ref{sec:Related-Work}. Secondly, we describe the details
of the developed GCSC framework and its two implementations in Section
\ref{sec:method}. In section \ref{sec:Experiments}, we give extensive
experimental results and empirical analysis. Finally, we conclude
with a summary and final remarks in Section \ref{sec:Conclusions}. 

\section{Preliminaries and Related Work \label{sec:Related-Work}}

\subsection{Notations}

Throughout this paper, boldface lowercase italics symbols (e.g., $\boldsymbol{x}$),
boldface uppercase roman symbols (e.g., $\mathbf{X}$), regular italics
symbols (e.g., $x_{ij}$ ), and calligraphy symbols (e.g., $\mathcal{S}$)
denote vectors, matrices, scalars, and sets, respectively. A graph
is represented as $\mathcal{G}=(\mathcal{V},\mathcal{E},\mathbf{A})$,
where $\mathcal{V}$ denotes the node set of the graph with $v_{i}\in\mathcal{V}$
and $\left|\mathcal{V}\right|=N$, $\mathcal{E}$ indicates the edge
set with $\left(v_{i},v_{i}\right)\in\mathcal{E}$, and $\mathbf{A}\in\mathbb{R}^{N\times N}$
stands for an adjacency matrix. We define the diagonal degree matrix
of the graph as $\mathbf{D}\in\mathbb{R}^{N\times N}$, where $D_{ij}=\sum_{j}A_{ij}$.
The graph Laplacian is defined as $\mathbf{L}=\mathbf{D}-\mathbf{A}$,
and its normalized version is given by $\mathbf{L}_{sym}=\mathbf{D}^{-1/2}\mathbf{L}\mathbf{D}^{-1/2}$.
In this paper, $\mathbf{X}^{T}$ denotes the transpose of matrix $\mathbf{X}$
and $\mathbf{I}_{N}$ denotes an identity matrix with the size of
$N$. The Frobenius norm of a matrix is defined as $\mathbf{\left\Vert X\right\Vert }_{F}=\left(\sum_{ij}\left|x_{ij}\right|^{2}\right)^{1/2}$
and the trace of a matrix is denoted as $tr\left(\mathbf{X}\right)$.

\subsection{Subspace Clustering Models}

Let $\mathbf{X}=\left[\boldsymbol{x}_{1},\boldsymbol{x}_{2},\cdots,\boldsymbol{x}_{N}\right]\in\mathbb{R}^{m\times N}$
be a collection of $N$ data points $\left\{ \boldsymbol{x}_{i}\in\mathbb{R}^{m}\right\} _{i=1}^{N}$
drawn from a union of linear or affinity subspaces $\mathcal{S}_{1}\cup\mathcal{S}_{2}\cup\cdots\cup\mathcal{S}_{n}$,
where $N$, $m$, and $n$ denote the number of data points, features,
and subspaces, respectively. The subspace clustering model for the
given data set $\mathbf{X}$ is defined as the following self-representation
problem \cite{SSC-Elhamifar-TPAMI-2013,Laplacian-Regularized-LRR-TPAMI-YinM-2016}:

\begin{equation}
\min\left\Vert \mathbf{W}\right\Vert _{p}~s.t.~\mathbf{XW}=\mathbf{X},~s.t.,~diag\left(\mathbf{W}\right)=0,\label{eq:sc-2}
\end{equation}
where $\mathbf{W}\in\mathbb{R}^{N\times N}$ denotes the self-expressive
coefficient matrix and $diag\left(\mathbf{W}\right)=0$ enforces the
diagonal elements of $\mathbf{W}$ to be zero so that the trivial
solutions are avoided. $\left\Vert \mathbf{W}\right\Vert _{p}$ denotes
a $p$-norm of matrix $\mathbf{W}$, e.g., $\left\Vert \mathbf{W}\right\Vert _{1}$
(SSC) \cite{SSC-Elhamifar-TPAMI-2013}), and$\left\Vert \mathbf{W}\right\Vert _{2}$
($\ell_{2}$-SSC \cite{HSIClustering-L2SSC-ZhanH-GRSL-2017}) . 

In the SSC model, the self-expressive coefficient matrix is assumed
to be sparse and the self-representation problem is often formulated
as 

\begin{equation}
\underset{\mathbf{W}}{\arg\min}\left\Vert \mathbf{XW}-\mathbf{X}\right\Vert _{F}^{2}+\lambda\left\Vert \mathbf{W}\right\Vert _{1},~s.t.,~diag\left(\mathbf{W}\right)=0,\label{eq:ssc}
\end{equation}
Here, the $\ell_{1}$-norm tends to produce a sparse coefficient matrix.
By using a nuclear norm, LRSC \cite{LRSC-2-ZhuX-TKDE-2019,Low-rank-SC-liu-ICML-2010,Laplacian-Regularized-LRR-TPAMI-YinM-2016}
reformulates the self-expressiveness property of data as 

\begin{equation}
\underset{\mathbf{W}}{\arg\min}\left\Vert \mathbf{XW}-\mathbf{X}\right\Vert _{2,1}^{2}+\lambda\left\Vert \mathbf{W}\right\Vert _{*},~s.t.,~diag\left(\mathbf{W}\right)=0,\label{eq:lrsc}
\end{equation}
where $\left\Vert \cdot\right\Vert _{*}$ and $\left\Vert \cdot\right\Vert _{2,1}$
denote the nuclear norm and $\ell_{2,1}$-norm of a matrix. LRSC has
been proven to be effective to incorporate the global structure of
data. Furthermore, subspace clustering can use to model corrupted
data, i.e., $\mathbf{X}=\mathbf{XW}+\mathbf{N},$where $\mathbf{N}$
is arbitrary noise. 

The above problems can be efficiently solved by using convex optimization
methods, such as Alternating Direction Method of Multipliers (ADMM)
\cite{Subspace-Clustering-Vidal-ISPM-2011,SS-Block-Diagonal-Representation-TPAMI-2019}.
Once the coefficient matrix $\mathbf{W}$ is found, the subspace clustering
seeks to segment an affinity matrix $\mathbf{A}=\frac{1}{2}\left(\left|\mathbf{W}\right|+\left|\mathbf{W}\right|^{T}\right)$
by Spectral Clustering (SC) method \cite{HSIClustering-SpectralClustering-GRSL-2017}. 

\begin{center}
\begin{figure*}[tbh]
\begin{centering}
\includegraphics[width=1.8\columnwidth]{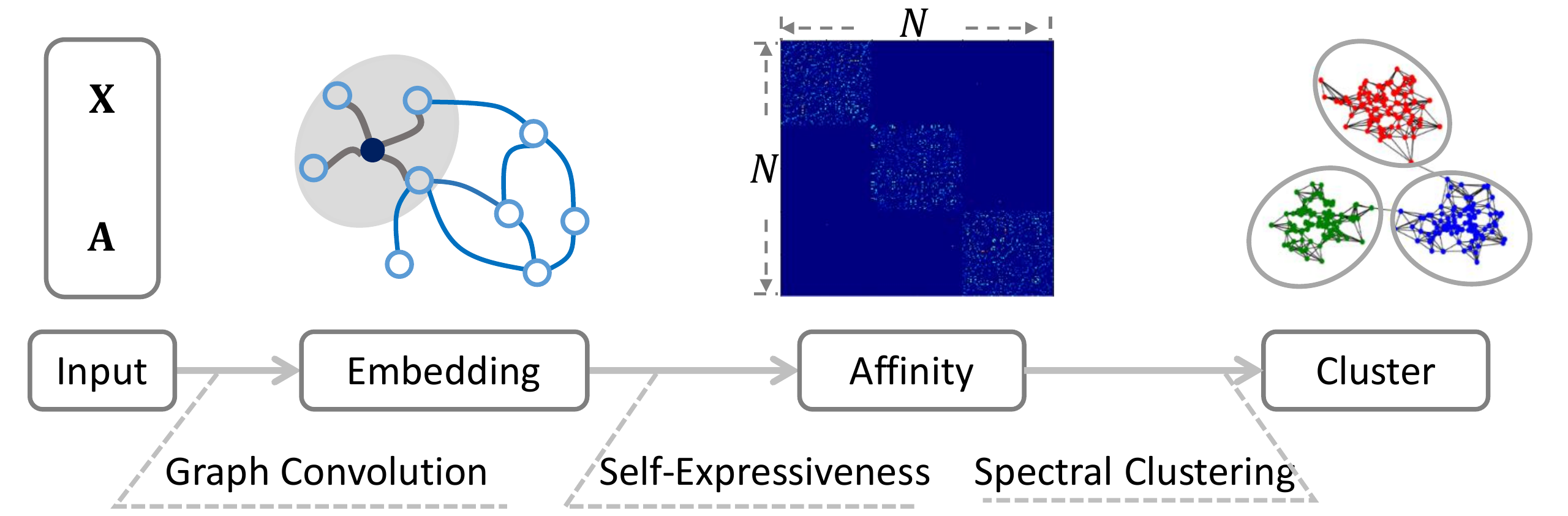}
\par\end{centering}
\caption{The flowchart of the proposed graph convolutional subspace clustering
framework. The framework follows the basic steps of the existing subspace
clustering methods but takes a feature matrix $\mathbf{X}$ and a
adjacency matrix $\mathbf{A}$ of the graph as inputs, which results
in a novel graph convolutional self-representation. The graph convolution
will generate a robust graph embedding that is used as a dictionary
for the subsequent affinity learning. \label{fig:The-framework}}
\end{figure*}
\par\end{center}

\subsection{Graph Convolutional Networks}

There is an increasing interest in generalizing convolutions to the
graph domain \cite{GNN-revirew-zhou201-arXiv,Geometric-GraphNN-Overview-Bronstein-SPM-2017}.
The recent development of GNNs that allows to efficiently approximate
convolution on graph-structured data. GNNs can typically divide into
two categories \cite{GNN-Survey-Zhang-TKDE-2020,GNN-Survey-Wu-TNNLS-2020}:
spectral convolutions, which perform convolution by transforming node
representations into the spectral domain using the graph Fourier transform
or its extensions, and spatial convolutions, which perform convolution
by considering node neighborhoods. Unless otherwise specified, the
graph convolution involved in this paper is the spectral convolution. 

One of the most representative graph convolution models is the Graph
convolutional networks (GCN) developed by Kift et al. \cite{gcn-kipf2017-iclr2017}.
GCN simplifies the spectral convolution by approximating spectral
filters with the $1^{th}$-order Chebyshev polynomials and setting
the largest eigenvalue of the normalized graph Laplacian $\mathbf{L}_{sym}$
to $2$. Formally, GCN defines spectral convolution over a graph as
follows:

\begin{equation}
\mathbf{H}=\sigma\left(\tilde{\mathbf{D}}^{-1/2}\tilde{\mathbf{A}}\tilde{\mathbf{D}}^{-1/2}\mathbf{X}^{T}\mathbf{W}\right).\label{eq:graph-cov}
\end{equation}
Here, $\tilde{\mathbf{A}}=\mathbf{I}_{N}+\mathbf{A}$ is an adjacency
matrix with self-loops, $\tilde{\mathbf{D}}$ denotes the degree matrix
of nodes whose elements are given by $\tilde{D}_{ii}=\sum_{j}\tilde{A_{ij}}$,
$\mathbf{W}\in\mathbb{R}^{m\times l}$ denotes a trainable parameter
matrix, and $\sigma$ is a nonlinear activation function. Specifically,
GCN takes a node feature matrix $\mathbf{X}$ and an adjacency matrix
$\mathbf{\mathbf{A}}$ as inputs, and produce a graph embedding $\mathbf{H}\in\mathbb{R}^{N\times l}$,
where $l$ is the output dimension. 

GCN is originally developed for semi-supervised node classification.
By stacking several graph convolution layers, GCN is possible to learn
deeper graph representation. Similar to traditional deep neural networks,
GCN can be easily trained using gradient descent methods. In different
tasks, the GCN models allow many traditional problems to be revised
in the non-Euclidean domain. 

\subsection{HSI Clustering }

Although supervised methods have achieved great success in HSI classification
\cite{HSIC-New-Frontier-Ghamisi-IGRSM-2018,HSI-Recent-Advances-on-Spectral-Spatial-Classification-He-TGRS-2018},
they are limited by the lack of sufficient labeled data. To avoid
data annotation, HSI clustering has attracted increasing attention.
HSI clustering is based on the fact that the same land-cover object
often shows similar spectral curves. The traditional centroid based
clustering methods, such as K-means \cite{kmeans-1979} and fuzzy
c-means (FCM) \cite{Fuzzy-cluster-Lei-TFS-2019}, are widely used
and easy to implement. However, this kind of method is sensitive to
the random initialization state and thus their clustering results
are hard to reproduce \cite{HSIClustering-RMMF-ZhangLF-IS-2019}. 

Due to the stable performance, subspace clustering is frequently-used
for HSI clustering. Zhang et al. \cite{HSIClustering-L2SSC-ZhanH-GRSL-2017,HSI-Clustering-KSSC-Zhai-RS-2017,HSI-Clustering-JSSC-Huang-ICIP-2018,HSIClustering-S4C-ZhangH-TGRS-2016}
have successfully applied various subspace clustering methods to the
HSI clustering, including Spectral--Spatial Sparse Subspace Clustering
(S$^{4}$C) \cite{HSIClustering-S4C-ZhangH-TGRS-2016}, Kernel Sparse
Subspace Clustering \cite{HSI-Clustering-KSSC-Zhai-RS-2017}, Joint
Sparsity Based Sparse Subspace Clustering (JSSC) \cite{HSI-Clustering-JSSC-Huang-ICIP-2018},
and so on. It is benefited from the ability to exploit the intrinsic
structure of data, subspace clustering has achieved impressive performance.
In recently, evolutionary optimization based clustering method has
attracted increasing interests, e.g., evolutionary multiobjective
optimization based HSI clustering \cite{HSI-Clustering-MPSO-Paoli-TGRS-2009,HSI-clustering-MOEA-SC-Wan-TGRS-2020}.
It is well known that the evolutionary algorithm is powerful to search
the globally optimal solution but it often results in huge computational
cost \cite{EMO-Sparse-Fang-Sensors-2020,Gong-SE2013,Gong-TSMCS2018,Cai-IJCNN-2018}.

It has been proven to be effective to improve HSI clustering performance
by utilizing spectral and spatial information, simultaneously. In
\cite{HSIClustering-RMMF-ZhangLF-IS-2019}, Zhang et al. developed
a state-of-the-art HSI clustering method called Robust Manifold Matrix
Factorization (RMMF) clustering by combining HSI dimensionality reduction
and data clustering, simultaneously. Kong et al. \cite{HSIClustering-UBL-GRSL-2019}
proposed an Unsupervised Broad Learning (UBL) clustering method that
combines clustering with broad representation learning. In our recent
work \cite{HSI-clustering-DSC-MZeng-IGARSS-2019}, we proposed a deep
subspace clustering method for the HSI clustering, which further demonstrates
the potential of combining clustering models with feature learning.

\section{Methodology \label{sec:method}}

In this section, we first introduce the proposed Graph Convolutional
Subspace Clustering (GCSC) framework. Then, we provide the details
of two novel subspace clustering models based on the framework, i.e.,
Efficient Graph Convolutional Subspace Clustering (EGCSC) and Efficient
Kernel Graph Convolutional Subspace Clustering (EKGCSC). We illustrate
a schematic representation of the proposed framework in Fig. \ref{fig:The-framework}
and more details are given in the following subsections.

\subsection{Graph Convolutional Subspace Clustering Framework}

Inspired by the recent development of GCNs, we present a novel subspace
clustering framework by incorporating graph embedding into subspace
clustering. We refer to the framework as GCSC. The goal of the GCSC
framework is to utilize graph convolution to learn a robust affinity.
For this purpose, we first modify the traditional self-representation
as follows:

\begin{equation}
\mathbf{X}=\mathbf{X}\bar{\mathbf{A}}\mathbf{Z},s.t.,diag\left(\mathbf{Z}\right)=0.\label{eq:self-repre}
\end{equation}
Here, $\mathbf{Z}\in\mathbb{R}^{N\times N}$ is the self-representation
coefficient matrix and $\bar{\mathbf{A}}=\tilde{\mathbf{D}}^{-1/2}\tilde{\mathbf{A}}\tilde{\mathbf{D}}^{-1/2}$
denotes the normalized symmetrical Laplacian matrix with self-loops.
Notably, $\mathbf{X}\bar{\mathbf{A}}\mathbf{Z}$ can be treated as
a special linear graph convolution operation (or a special graph auto-encoder)
parameterized by $\mathbf{Z}$. We call Eq. \eqref{eq:self-repre}
graph convolutional self-representation.

Parallel to traditional subspace clustering models, the GCSC framework
can be rewritten as 

\begin{equation}
\underset{\mathbf{Z}}{\arg\min}\frac{1}{2}\left\Vert \mathbf{X}\bar{\mathbf{A}}\mathbf{Z}-\mathbf{X}\right\Vert _{q}+\frac{\lambda}{2}\left\Vert \mathbf{Z}\right\Vert _{p},s.t.,diag\left(\mathbf{Z}\right)=0,\label{eq:gcsc-framework}
\end{equation}
where $q$ and $p$ denote any appropriate matrix norm, such as $p,q=0,\frac{1}{2},1,2$
and nuclear norm, and $\lambda$ is a tradeoff coefficient. It is
easy to prove that the traditional subspace models are the special
cases of our framework, i.e., the traditional subspace models depend
only on data features. For example, when $q$ is the Frobenius norm
and $p=1$, Eq. \eqref{eq:gcsc-framework} becomes the extension of
the classical SSC \cite{SSC-Elhamifar-TPAMI-2013}, while $q$ is
$\ell_{2,1}$ norm and $p$ is the nuclear norm, Eq. \eqref{eq:gcsc-framework}
degenerates to LRSC \cite{LRSC-1-Vidal-PRLetters-2014,Low-rank-SC-liu-ICML-2010}.
Eq. \eqref{eq:gcsc-framework} can be effectively solved by the same
method adopted in the traditional subspace clustering. Once the self-representation
coefficient matrix $\mathbf{Z}$ is obtained, SC can be used to generate
the clustering results.

\subsection{Efficient GCSC}

Basing on the proposed framework, we present the first novel subspace
clustering model, namely Efficient GCSC (EGCSC), by setting both $q$
and $p$ to be the Frobenius norm. Formally, we formulate the EGCSC
model as 

\begin{equation}
\underset{\mathbf{Z}}{\arg\min}\frac{1}{2}\left\Vert \mathbf{X}\bar{\mathbf{A}}\mathbf{Z}-\mathbf{X}\right\Vert _{F}^{2}+\frac{\lambda}{2}\left\Vert \mathbf{Z}\right\Vert _{F}^{2},\label{eq:gcsc}
\end{equation}
In \cite{Efficient-dense-SC-PanJ-WACV-2014}, Pan et al. have proven
that the Frobenius norm will not result in trivial solutions even
without constraint $diag\left(\mathbf{Z}\right)=0$. This leads to
a dense self-representation coefficient matrix and can be denoted
as an efficient closed-form solution. The solution is given by

\begin{equation}
\mathbf{Z}=\left(\bar{\mathbf{A}}^{T}\mathbf{X}^{T}\mathbf{X}\bar{\mathbf{A}}+\lambda\mathbf{I}_{N}\right)^{-1}\bar{\mathbf{A}}^{T}\mathbf{X}^{T}\mathbf{X}.\label{eq:solution-basic}
\end{equation}
The proof of Eq. \eqref{eq:solution-basic} is given in Appendix section
\ref{sec:Appendix}. 

Having obtained $\mathbf{Z}$, we can use it to construct an affinity
matrix $\mathbf{C}$ for the SC. However, there is no globally-accepted
solution for this step in the literature. Most existing works typically
compute the affinity matrix by $\mathbf{C}=\left|\mathbf{Z}\right|+\left|\mathbf{Z}\right|^{T}$
or $\left|\mathbf{Z}\right|$. In this paper, we use the heuristic
adopted by Efficient Dense Subspace Clustering (EDSC) \cite{Efficient-dense-SC-PanJ-WACV-2014}
to enhance the block-structure, which is proved beneficial for clustering
accuracy. The pseudocode of the EGCSC is given in Algorithm \ref{alg:pseudocode-1}.

\begin{algorithm}[tbh]  
\caption{EGCSC} 
\label{alg:pseudocode-1}
\KwIn{$\mathbf{\mathbf{X}}$, $\mathbf{\mathbf{A}}$, $\lambda$, and the number of clusters.}
Compute $\bar{\mathbf{A}}=\tilde{\mathbf{D}}^{-1/2}\tilde{\mathbf{A}}\tilde{\mathbf{D}}^{-1/2}$\;
Compute coefficient matrix: $\mathbf{Z}=\left(\bar{\mathbf{A}}^{T}\mathbf{X}^{T}\mathbf{X}\bar{\mathbf{A}}+\lambda\mathbf{I}_{N}\right)^{-1}\bar{\mathbf{A}}^{T}\mathbf{X}^{T}\mathbf{X}$\;
Construct affinity matrix  $\mathbf{C}$\;
Apply spectral clustering on $\mathbf{C}$\;
\KwOut{Clustering results.}
\end{algorithm}

\subsection{Efficient Kernel GCSC}

We have proposed the EGCSC method. However, the EGCSC model is essentially
modeled on linear subspaces. Due to the complexity and nonlinearity
of HSI, a large number of works have demonstrated that nonlinear models
will yield better performance than their linear counterparts. In this
subsection, we provide a nonlinear extension of EGCSC by using the
kernel trick. The extension is referred to as Efficient Kernel GCSC
(EKGCSC). 

Let $\varPhi:\mathbb{R}^{m}\rightarrow\mathcal{H}$ be a mapping from
the input space to the reproducing kernel Hilbert space $\mathcal{H}$.
We define a positive semidefinite kernel Gram matrix $\mathbf{K}_{\mathbf{XX}}\in\mathbb{R}^{N\times N}$
as 

\begin{equation}
\left[\mathbf{K}_{\mathbf{XX}}\right]_{ij}=\left[\left\langle \varPhi\left(\mathbf{X}\right),\varPhi\left(\mathbf{X}\right)\right\rangle _{\mathcal{H}}\right]=\varPhi\left(\boldsymbol{x}_{i}\right)^{T}\varPhi\left(\boldsymbol{x}_{j}\right)=\kappa\left(\boldsymbol{x}_{i},\boldsymbol{x}_{j}\right)\label{eq:kernel}
\end{equation}
where $\kappa:\mathbb{R}^{m}\times\mathbb{R}^{m}\rightarrow\mathbb{R}$
denotes the kernel function. In this paper, the Gaussian kernel is
used, i.e., $\kappa\left(\boldsymbol{x}_{i},\boldsymbol{x}_{j}\right)=\exp\left(-\gamma\left\Vert \boldsymbol{x}_{i}-\boldsymbol{x}_{j}\right\Vert ^{2}\right)$,
where $\gamma$ is the parameter of the Gaussian kernel function.
Formally, the EKGCSC model is expressed as 

\begin{equation}
\underset{\mathbf{Z}}{\arg\min}\frac{1}{2}\left\Vert \varPhi\left(\mathbf{X}\right)\bar{\mathbf{A}}\mathbf{Z}-\varPhi\left(\mathbf{X}\right)\right\Vert _{F}^{2}+\frac{\lambda}{2}\left\Vert \mathbf{Z}\right\Vert _{F}^{2}.\label{eq:kgcsc}
\end{equation}
By using kernel trick, Eq. \eqref{eq:kgcsc} can be equivalently rewritten
as 

\begin{equation}
\begin{array}{c}
\underset{\mathbf{Z}}{\arg\min}~\frac{1}{2}tr(\mathbf{Z}^{T}\bar{\mathbf{A}}^{T}\mathbf{K}_{\mathbf{XX}}\bar{\mathbf{A}}\mathbf{Z}-2\mathbf{K}_{\mathbf{XX}}\bar{\mathbf{A}}\mathbf{Z}\\
+\mathbf{K}_{\mathbf{XX}}+\lambda\mathbf{Z}^{T}\mathbf{Z})
\end{array},
\end{equation}
The above problem can be solved by calculating the partial derivative
with respect to $\mathbf{Z}$ and set it to be zero (see Appendix
section \ref{sec:Appendix}). The closed-form solution of EKGCSC is
given by 

\begin{equation}
\mathbf{Z}=\left(\bar{\mathbf{A}}^{T}\mathbf{K}_{\mathbf{XX}}\bar{\mathbf{A}}+\lambda\mathbf{I}_{N}\right)^{-1}\bar{\mathbf{A}}^{T}\mathbf{K}_{\mathbf{XX}}.
\end{equation}
The EKGCSC model explicitly maps the original data points onto a higher-dimensional
space, and thus makes a linearly inseparable problem to be a separable
one. We use a manner that is similar to EGCSC to construct the affinity
matrix and obtain the final clustering results by SC. The pseudocode
of EKGCSC is given in Algorithm \ref{alg:pseudocode-2}.

\begin{algorithm}[tbh]  
\caption{EKGCSC} 
\label{alg:pseudocode-2}
\KwIn{$\mathbf{\mathbf{X}}$, $\mathbf{\mathbf{A}}$, $\lambda$, kernel parameters, and the number of clusters.}
Compute $\bar{\mathbf{A}}=\tilde{\mathbf{D}}^{-1/2}\tilde{\mathbf{A}}\tilde{\mathbf{D}}^{-1/2}$\;
Compute kernel matrix $\mathbf{K}_{\mathbf{XX}}$ according to Eq. \eqref{eq:kernel}\;
Compute coefficient matrix: $\mathbf{Z}=\left(\bar{\mathbf{A}}^{T}\mathbf{K}_{\mathbf{XX}}\bar{\mathbf{A}}+\lambda\mathbf{I}_{N}\right)^{-1}\bar{\mathbf{A}}^{T}\mathbf{K}_{\mathbf{XX}}$\;
Construct affinity matrix  $\mathbf{C}$\;
Apply spectral clustering on $\mathbf{C}$\;
\KwOut{Clustering results.}
\end{algorithm}

\subsection{HSI Clustering Using The GCSC Models}

We use the proposed GCSC models for HSI clustering. Two essential
issues need to be tackled before using the GCSC models. First, HSI
data often includes many spectral bands with lots of redundancy, and
thus using only spectral features is hard to achieve good performance.
Second, the GCSC models are based on the graph-structured data, and
however, HSI is typically a Euclidean data.

To remedy the first issue, the following procedures are employed.
We first use Principle Component Analysis (PCA) to reduce the spectral
dimensionality by preserving the top $d$ PCs. On the one hand, PCA
reduces the redundant information contained in HSI data. On the other
hand, it increases computational efficiency when model training. To
take spectral and spatial information consideration, simultaneously,
we represent every data point by extracting 3D patches. Specifically,
every data point is represented by the center pixel and its neighboring
pixels. The manner is widely adopted in different HSI spectral-spatial
classification methods \cite{BS-Net-TGRS-2020-Cai,HSI-Advances-in-Spectral-Spatial-Classification-IEEEProceedings-2013,HSI-Recent-Advances-on-Spectral-Spatial-Classification-He-TGRS-2018}. 

For the second issue, we construct a $k$-nearest neighbor (kNN) graph
to represent the graph structure of the data points. Specifically,
each data point is viewed as a node over the graph and the $k$ nearest
neighbors of $\boldsymbol{x}_{i}$ consists of the edge relationship.
The adjacent matrix $\mathbf{A}$ of a kNN graph  is defined by

\begin{equation}
A_{ij}=\begin{cases}
1 & \boldsymbol{x}_{j}\in\mathcal{N}_{k}\left(\boldsymbol{x}_{i}\right)\\
0 & otherwise
\end{cases},
\end{equation}
where $\mathcal{N}_{k}\left(\boldsymbol{x}_{i}\right)$ indicates
the $k$ nearest neighbors of $\boldsymbol{x}_{i}$. The neighborhood
relationship is obtained by computing the Euclidean distance. 

\subsection{Remarks on The Proposed GCSC}
\begin{center}
\begin{figure}[tbh]
\begin{centering}
\subfloat[]{\includegraphics[width=0.5\columnwidth]{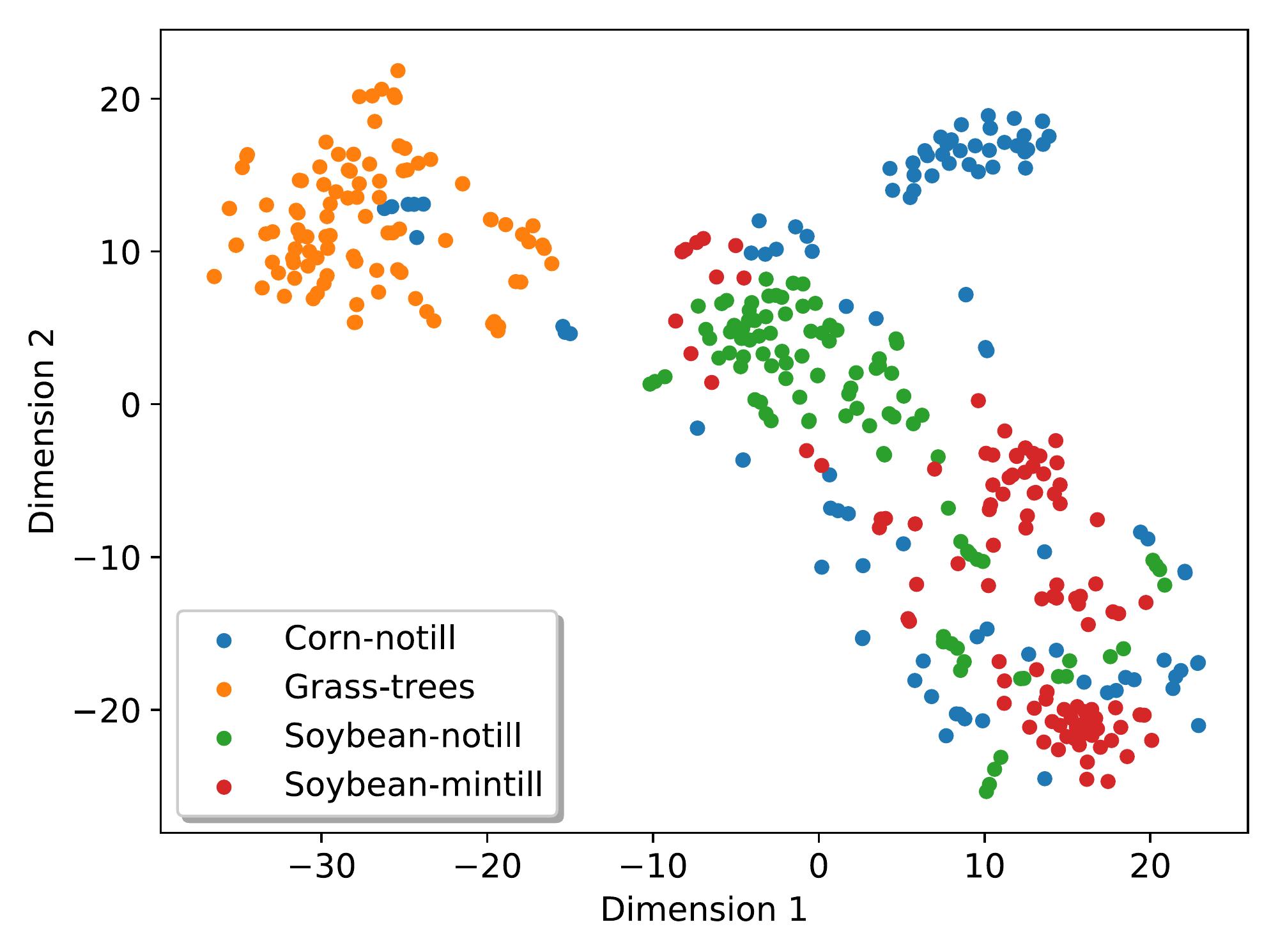}

}\subfloat[]{\includegraphics[width=0.5\columnwidth]{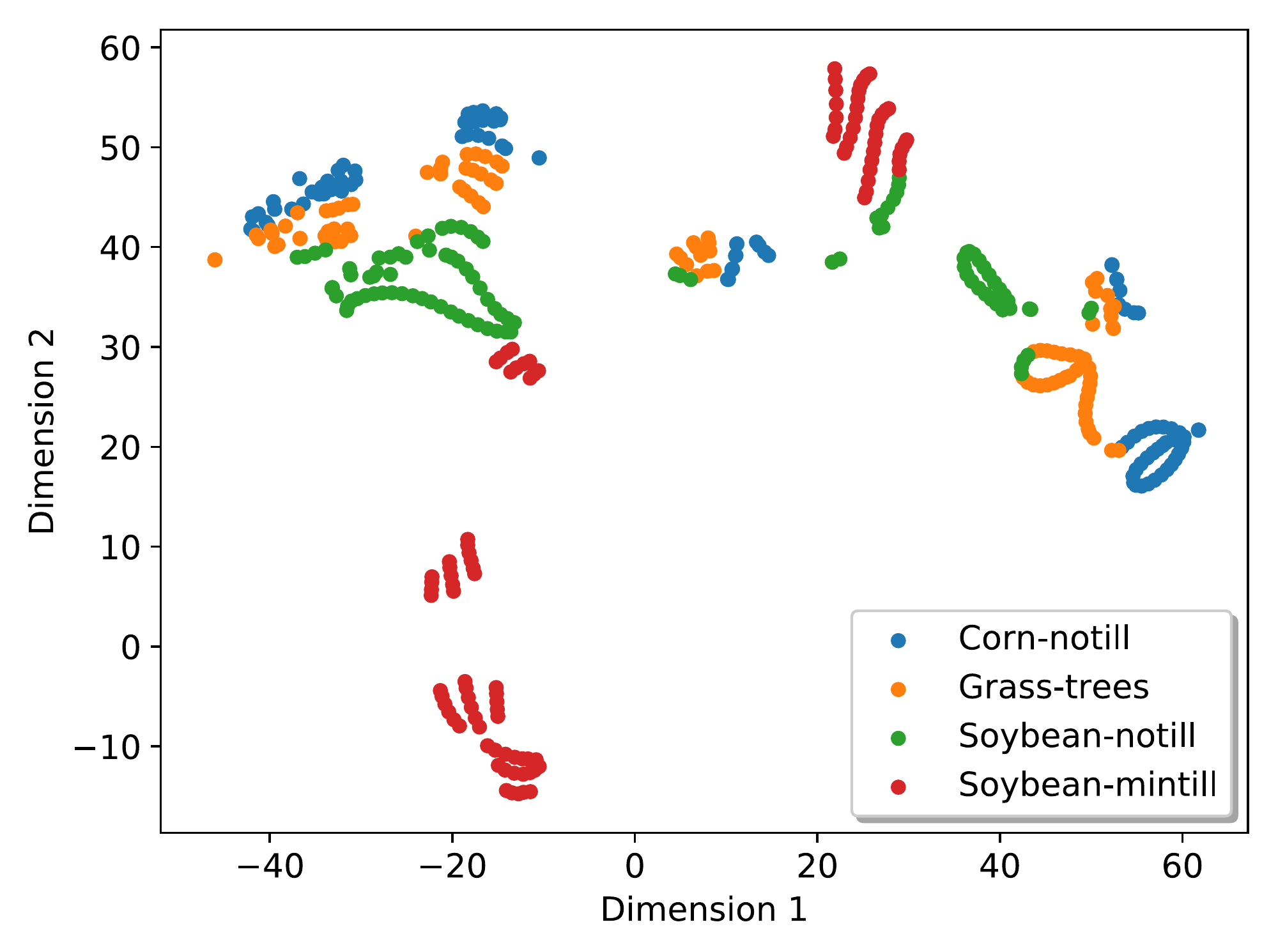}

}
\par\end{centering}
\caption{Visualization of data points selected from Indian Pines data set,
where we randomly select $100$ data points per class and reduce their
feature dimensionality into $2$ with t-SNE \cite{t-SNE-Hinton-JMLR-2008}.
(a) Original data points. (b) Embedding using graph convolution. It
can be seen that corn-till, soybean-notill, and soybean-mintill are
mixed in the original data distribution. By contrast, they show better
separability and are more compact after transformed by graph convolution.
\label{fig:Visualization-of-feature}}
\end{figure}
\par\end{center}

In this subsection, we provide a deeper insight into the GCSC framework
from the following viewpoints. Let $\mathbf{Y}=\mathbf{X}\bar{\mathbf{A}}$
be the graph embedding, thus GCSC can be rewritten as 

\begin{equation}
\underset{\mathbf{Z}}{\arg\min}\frac{1}{2}\left\Vert \mathbf{Y}\mathbf{Z}-\mathbf{X}\right\Vert _{q}+\frac{\lambda}{2}\left\Vert \mathbf{Z}\right\Vert _{p},s.t.,diag\left(\mathbf{Z}\right)=0,\label{eq:gcsc-framework-1}
\end{equation}
From the viewpoint of sparse representation, GCSC aims to use a self-expressive
dictionary matrix $\mathbf{Y}$ to reconstruct the original data.
Since $\mathbf{Y}$ considers the global structure information, those
noise points will be eliminated and a clear dictionary can be obtained,
which is beneficial for producing a robust affinity matrix. It can
be seen from Fig. \ref{fig:Visualization-of-feature}, the resulting
$\mathbf{Y}$ shows better clustering characteristics than the original
$\mathbf{X}$. It can be further explained from the viewpoint of graph
representation learning. Graph convolution essentially is a special
form of the Laplacian smoothing \cite{GCN-Deeperinsights-Li-AAAI-2018},
which combines the features of a node and its nearby neighbors. The
operation makes the features of the node in the same cluster similar,
thus greatly easing the clustering task. 

The main differences between the GCSC model and the traditional subspace
model are as follows. First, GCSC is built in the non-Euclidean domain.
Under the GCSC framework, the traditional subspace clustering models
can be considered as special cases in the Euclidean domain. Second,
GCSC incorporates the graph structure via graph convolution, while
the traditional subspace clustering models do this by manifold regularization.
Therefore, GCSC exploits graph information using a more straightforward
way. 
\begin{center}
\begin{table}[tbh]
\caption{Summary of Salinas, Indian Pines, and Pavia University datasets.\label{tab:Data-sets-descriptions}}

\centering{}%
\begin{tabular}{lccc}
\hline 
Datasets & SalinasA & Indian Pines & Pavia University\tabularnewline
\hline 
Pixels & 83$\times$86 & 85$\times$70 & 140$\times$150\tabularnewline
Channels & 204 & 200 & 103\tabularnewline
Clusters & 6 & 4 & 8\tabularnewline
Samples & 5348 & 4391 & 6445\tabularnewline
Sensor & AVIRIS & AVIRIS & ROSIS\tabularnewline
\hline 
\end{tabular}
\end{table}
\par\end{center}

\section{Experiments \label{sec:Experiments}}

\begin{table}[tbh] 
\caption{The settings of the important hyper-parameters in EGCSC and EKGCSC.}
\label{tab:settings}
\centering{}% 
\begin{tabular}{cccccc} 
\toprule  
\multirow{2}{*}{Method} & \multicolumn{2}{c}{EGCSC} & \multicolumn{3}{c}{EKGCSC}\\
\cmidrule(r){2-3}  \cmidrule(r){4-6}  
& $\lambda$ & $k$ & $\lambda$ & $k$ & $\gamma$\\
\midrule  
SalinasA & 100 & 30 & 100 & 30 & 0.2\\
Indian Pines & 100 & 30 & 100000 & 30 & 6\\
Pavia University & 1000 & 20 & 60000 & 30 & 100\\
\bottomrule 
\end{tabular} 
\end{table}

In this section, we extensively evaluate the clustering performance
of the proposed clustering methods on three frequently used HSI datasets.
The source codes of EGCSC and EKGCEC are released at https://github.com/AngryCai/GraphConvSC.

\subsection{Setup}

\subsubsection{Datasets and Preprocessing}

We conduct experiments on three real HSI images acquired by AVIRIS
and ROSIS sensors, i.e., Salinas, Indian Pines, and Pavia University.
For computational efficiency, we separately take a sub-scene of these
datasets for evaluation as it is done in \cite{HSI_BS-Laplacian-Regularized-LRSC-ZhaiH-TGRS-2018,HSIClustering-SPAHSIC-PanY-ICASSP2019,HSIClustering-UBL-GRSL-2019}.
Specifically, these sub-scenes are located within the original scenes
at $[591-676,158-240]$, $[30-115,24-94]$, and $[150-350,100-200]$,
respectively. Notice that the sub-scene taken from the Salinas dataset
is also known as the SalinasA dataset. The details of the three datasets
are summarized in Table \ref{tab:Data-sets-descriptions}. 

In data preprocessing, we perform PCA to reduce spectral bands into
$4$ by preserving at least $95\%$ of the cumulative percentage of
variance. We construct spectral-spatial samples by setting neighborhood
size to be $9$ for all the datasets. All data points are standardized
by scaling into $[0,1]$ before clustering. 

\subsubsection{Evaluation Metrics}

Three popular metrics \cite{HSIClustering-S4C-ZhangH-TGRS-2016,HSIClustering-RMMF-ZhangLF-IS-2019,HSIClustering-UBL-GRSL-2019}
are used to evaluate the clustering performance of clustering models,
i.e., Overall Accuracy (OA), Normalized Mutual Information (NMI),
and Kappa coefficient (Kappa). These metrics range in $[0,1]$, and
the higher the scores are, the more accurate the clustering results
are achieved. Besides, to evaluate the computational complexity of
our models, running time is compared in the experiment. 

\subsubsection{Compared Methods}

We compare the proposed methods with several existing HSI clustering
methods, including traditional clustering methods and state-of-the-art
methods. Specifically, the compared traditional clustering methods
contain Spectral Clustering (SC) \cite{HSIClustering-SpectralClustering-GRSL-2017},
Sparse Subspace Clustering (SSC) \cite{SSC-Elhamifar-TPAMI-2013},
Efficient Dense Subspace Clustering (EDSC) \cite{Efficient-dense-SC-PanJ-WACV-2014},
Low Rank Subspace Clustering (LRSC) \cite{LRSC-2-ZhuX-TKDE-2019},
and $\ell_{2}$-norm based SSC ($\ell_{2}$-SSC) \cite{HSIClustering-L2SSC-ZhanH-GRSL-2017}.
The compared state-of-the-art HSI clustering methods are Spectral-Spatial
Sparse Subspace Clustering (S$^{4}$C) \cite{HSIClustering-S4C-ZhangH-TGRS-2016},
Unsupervised Broad Learning (UBL) clustering \cite{HSIClustering-UBL-GRSL-2019},
and Robust Manifold Matrix Factorization (RMMF) \cite{HSIClustering-RMMF-ZhangLF-IS-2019}. 

For these HSI clustering methods, i.e., $\ell_{2}$-SSC, S$^{4}$C,
UBL, and RMMF, we follow their settings reported in the corresponding
literature. The hyper-parameters of EGCSC and EKGCSC are given in
Table \ref{tab:settings}. All the compared methods are implemented
with Python 3.5 running on an Intel Xeon E5-2620 2.10 GHz CPU with
32 GB RAM. 
\begin{center}
\begin{table*}[tbph]
\caption{The clustering performance of the compared methods on Indian Pines,
SalinasA, and PaviaU datasets. The best results are highlighted in
bold. \label{tab:The-clustering-performance-three-datasets}}

\centering{}%
\begin{tabular}{cccccccccccc}
\hline 
Data & Metric & SC \cite{HSIClustering-SpectralClustering-GRSL-2017} & SSC \cite{SSC-Elhamifar-TPAMI-2013} & LRSC \cite{LRSC-2-ZhuX-TKDE-2019} & $\ell_{2}$-SSC \cite{HSIClustering-L2SSC-ZhanH-GRSL-2017} & S$^{4}$C \cite{HSIClustering-S4C-ZhangH-TGRS-2016} & UBL \cite{HSIClustering-UBL-GRSL-2019} & RMMF \cite{HSIClustering-RMMF-ZhangLF-IS-2019} & EDSC \cite{Efficient-dense-SC-PanJ-WACV-2014} & EGCSC & EKGCSC\tabularnewline
\hline 
\noalign{\vskip0.1cm}
\multirow{3}{*}{SaA.} & OA & 0.6806 & 0.7666 & 0.5613 & 0.6412 & 0.8631 & 0.9142 & 0.9820 & 0.8702 & 0.9985 & \textbf{1.0000}\tabularnewline
\noalign{\vskip0.1cm}
 & NMI & 0.7464 & 0.7571 & 0.4242 & 0.6971 & 0.7977 & 0.8692 & 0.9483 & 0.9135 & 0.9949 & \textbf{1.0000}\tabularnewline
\noalign{\vskip0.1cm}
 & Kappa & 0.6002 & 0.7138 & 0.4487 & 0.5546 & 0.8312 & 0.8943 & 0.9775 & 0.8384 & 0.9981 & \textbf{1.0000}\tabularnewline
\hline 
\noalign{\vskip0.1cm}
\multirow{3}{*}{InP.} & OA & 0.6841 & 0.4937 & 0.5142 & 0.6645 & 0.7008 & 0.6258 & 0.7121 & 0.7126 & 0.8483 & \textbf{0.8761}\tabularnewline
\noalign{\vskip0.1cm}
 & NMI & 0.5339 & 0.2261 & 0.2455 & 0.3380 & 0.5445 & 0.6680 & 0.4985 & 0.4717 & 0.6422 & \textbf{0.6959}\tabularnewline
\noalign{\vskip0.1cm}
 & Kappa & 0.5055 & 0.2913 & 0.3145 & 0.5260 & 0.5825 & 0.4690 & 0.5609 & 0.5657 & 0.6422 & \textbf{0.8211}\tabularnewline
\hline 
\noalign{\vskip0.1cm}
\multirow{3}{*}{PaU.} & OA & 0.7691 & 0.6146 & 0.4326 & 0.5842 & 0.6509 & 0.7083 & 0.7704 & 0.6175 & 0.8442 & \textbf{0.9736}\tabularnewline
\noalign{\vskip0.1cm}
 & NMI & 0.6784 & 0.6545 & 0.3793 & 0.4942 & 0.7031 & 0.6874 & 0.7388 & 0.5750 & 0.8401 & \textbf{0.9529}\tabularnewline
\noalign{\vskip0.1cm}
 & Kappa & 0.8086 & 0.4886 & 0.2549 & 0.3687 & 0.5852 & 0.6533 & 0.6804 & 0.4250 & 0.7968 & \textbf{0.9653}\tabularnewline
\hline 
\noalign{\vskip0.1cm}
\end{tabular}
\end{table*}
\par\end{center}

\subsection{Results }

\subsubsection{Quantitative Results}

Table \ref{tab:The-clustering-performance-three-datasets} gives the
clustering performance comparison of different methods evaluated on
SalinasA, Indian Pines, and Pavia University datasets. As can be seen
from the results, the proposed GCSC methods achieve the best clustering
performance and significantly outperform the other clustering methods
in terms of OA, NMI, and Kappa. We can further find the following
tendencies from the results. 

First, equipped with graph convolution, the traditional subspace clustering
models can achieve remarkable improvement compared with the traditional
counterpart. For example, EGCSC is significantly better than EDSC.
It signifies that the proposed GCSC framework is beneficial for subspace
clustering. It can be further seen from Table \ref{tab:The-clustering-performance-three-datasets},
few of the compared clustering methods can achieve $80\%$ OA. On
the contrary, OAs yielded by the EGCSC and EKGCSC models are generally
better than $84\%$ on all the datasets. Particularly, on the SalinasA
dataset, EKGCSC achieves a perfect ($100\%$) clustering performance. 

Second, EKGCSC outperforms EGCSC on all the three datasets. Due to
the complexity of HSI, linear models often can not fully exploit the
relationship among data points. By extending EGCSC into the nonlinear
kernel space, EKGCSC's performance can be dramatically enhanced. In
other words, EKGCSC considers the nonlinear relationship between data
points and makes the learned affinity matrix more robust. In the experiment,
EKGCSC achieves $0.15\%$, $2.78\%$, and $12.94\%$ improvement on
SalinasA, Indian Pines, and Pavia University datasets, respectively. 

Third, the results obtained by EGCSC and EKGCSC are comparable with
many supervised HSI classification methods \cite{HSI-Recent-Advances-on-Spectral-Spatial-Classification-He-TGRS-2018,HSIC-New-Frontier-Ghamisi-IGRSM-2018,DL-HSI-overview-ZhangLP-GRSM-2016}.
Specifically, the EKGCSC model achieves $100\%$, $87.61\%$, and
$97.36\%$ in terms of OA on the SalinasA, Indian Pines, and Pavia
University datasets, respectively. The recent development of supervised
HSI classification allows achieving excellent results. However, the
unsupervised classification of HSI is still a challenging task. The
state-of-the-art clustering performance of our methods bridges the
gap between unsupervised HSI classification and supervised HSI classification. 

\subsubsection{Results Visualization}
\begin{center}
\begin{figure*}[tbh]
\begin{centering}
\subfloat[]{\includegraphics[width=0.22\columnwidth]{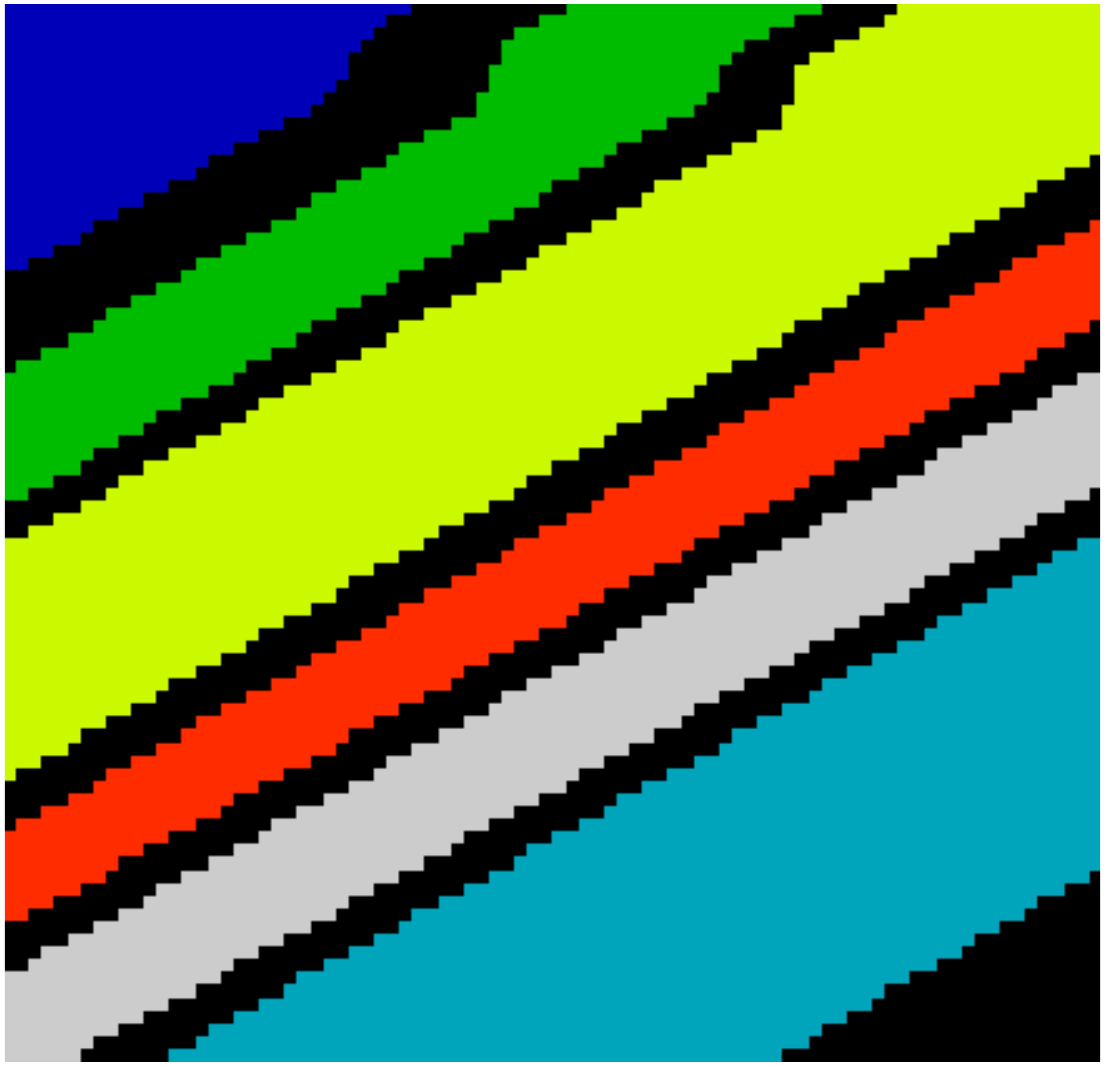}

}\subfloat[]{\includegraphics[width=0.22\columnwidth]{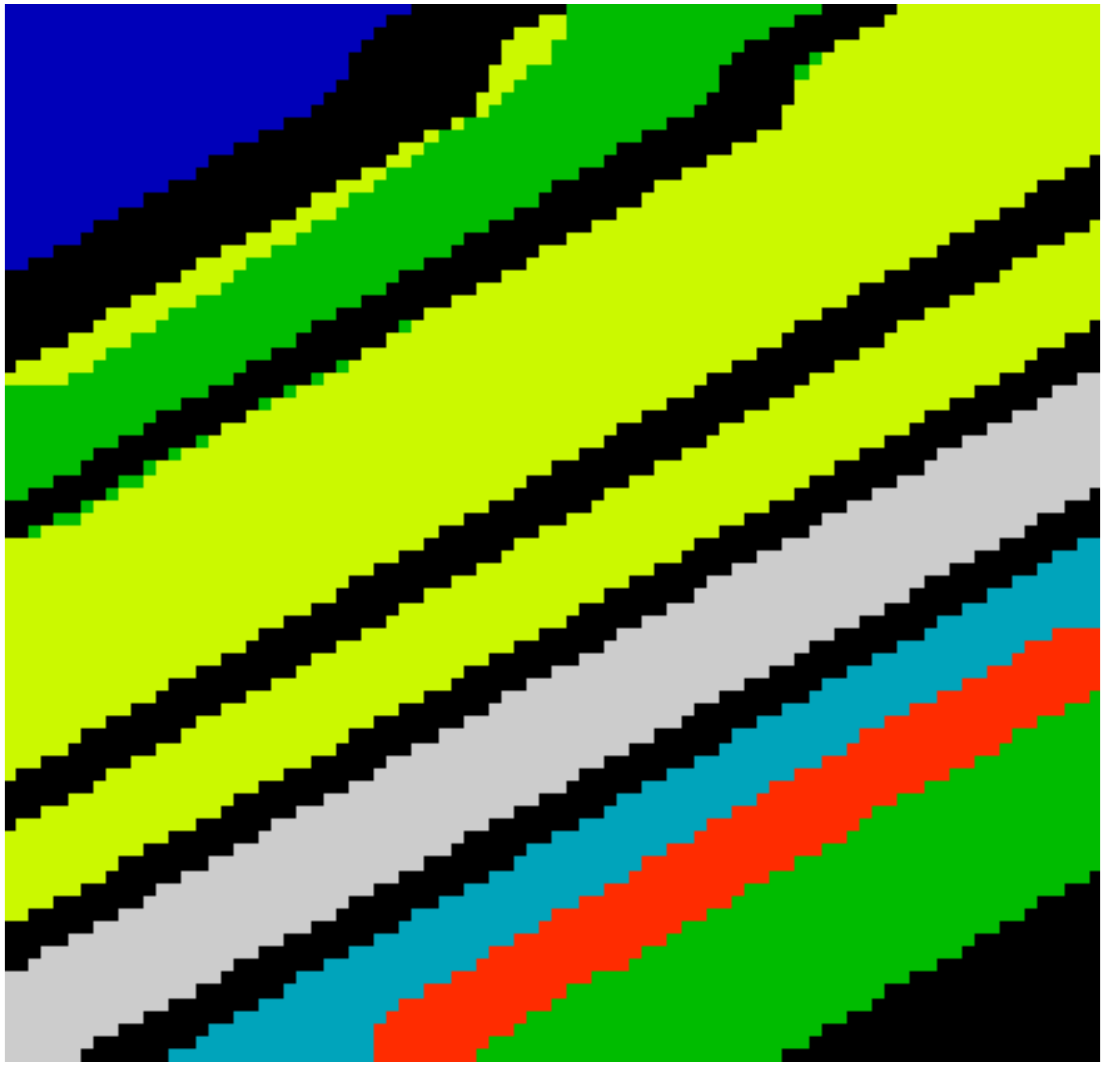}

}\subfloat[]{\includegraphics[width=0.22\columnwidth]{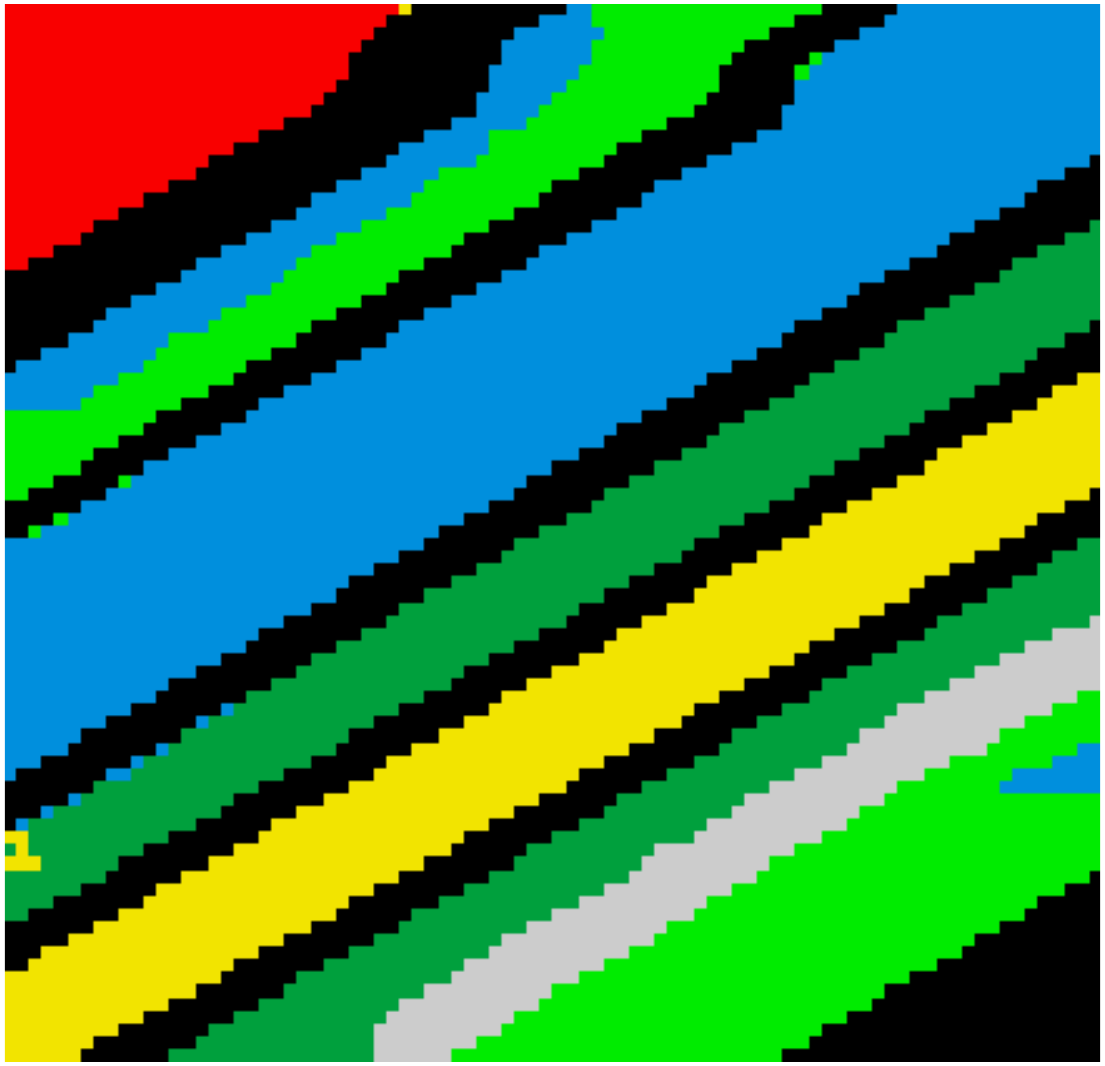}

}\subfloat[]{\includegraphics[width=0.22\columnwidth]{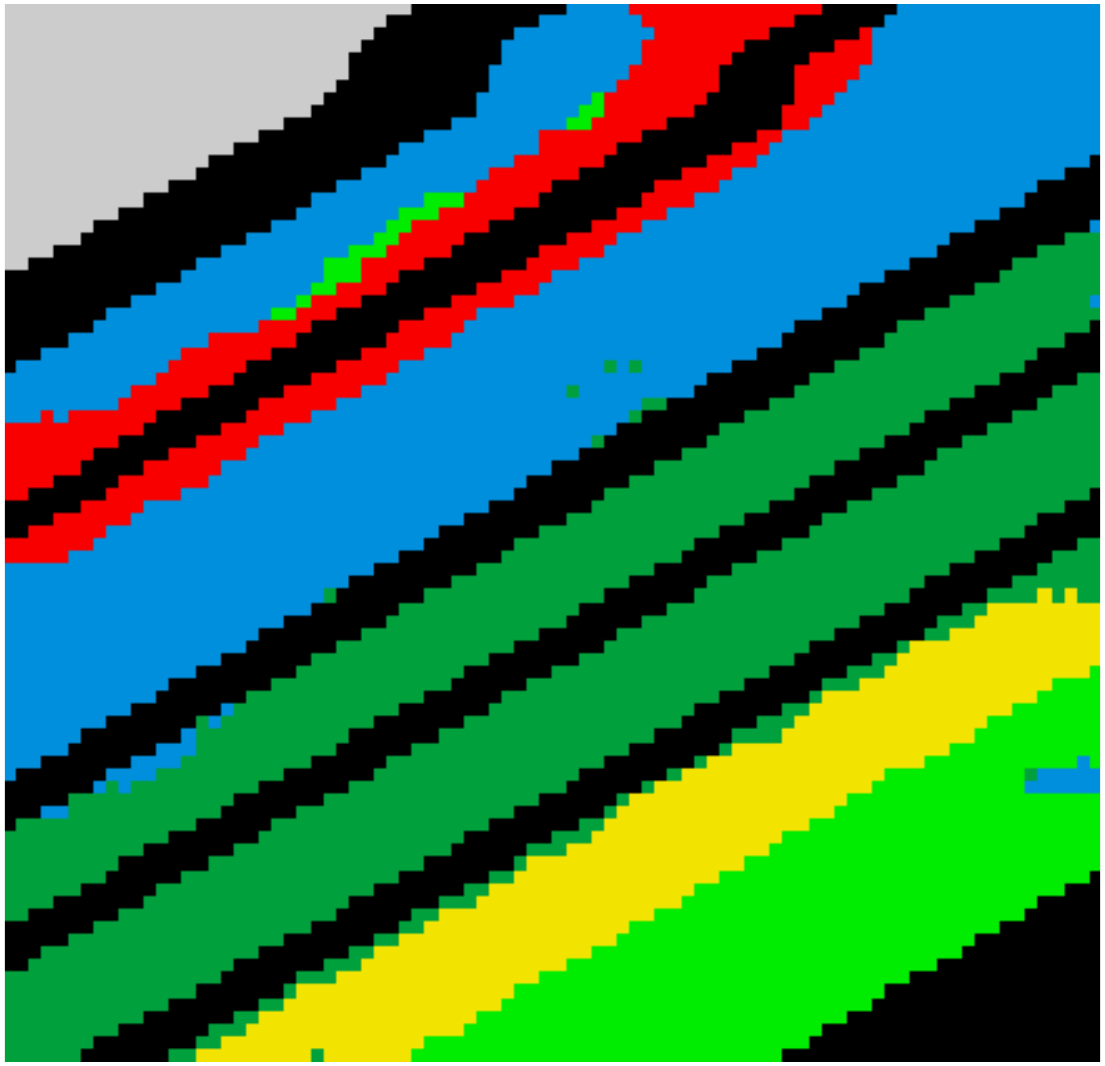}

}\subfloat[]{\includegraphics[width=0.22\columnwidth]{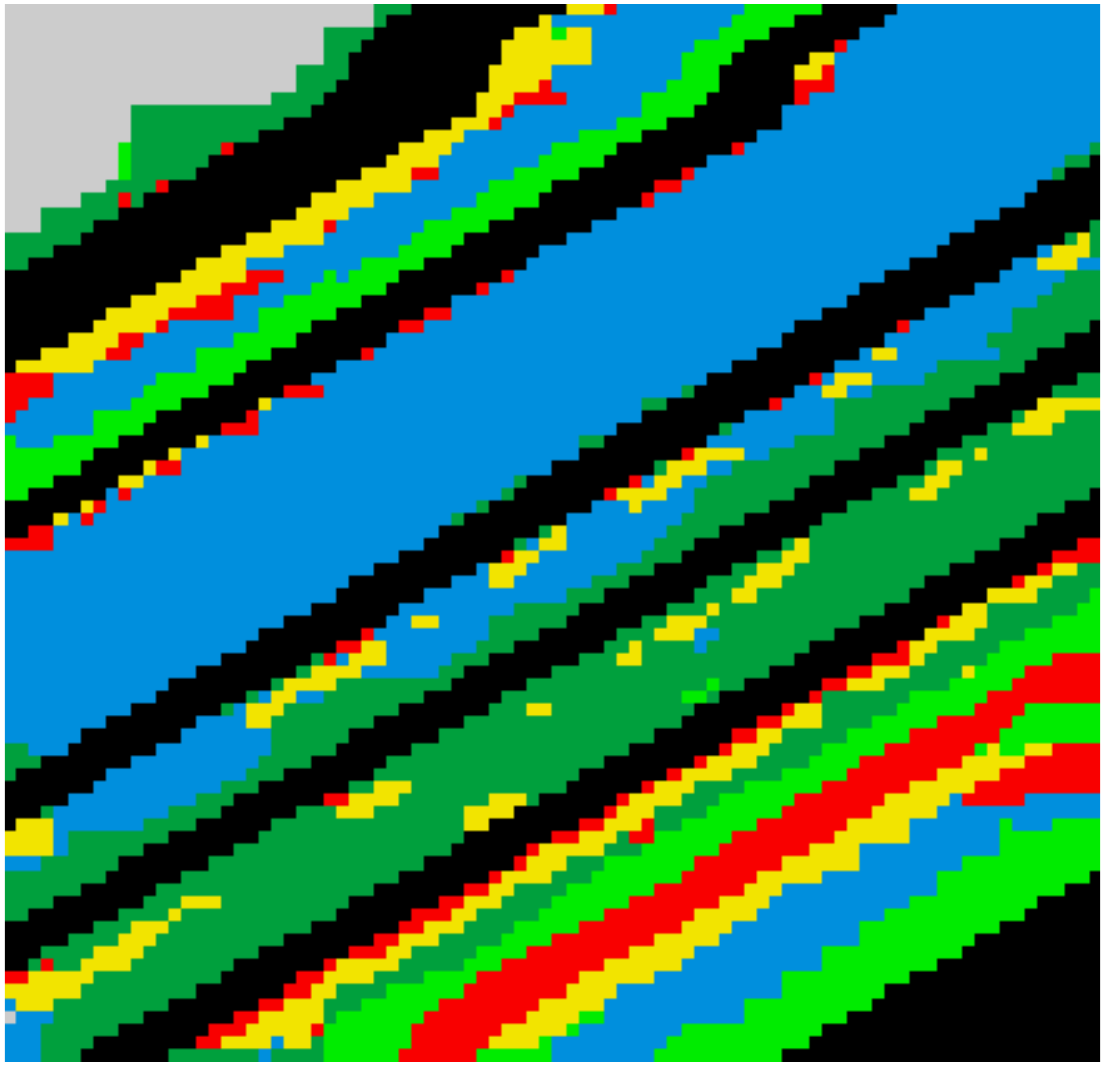}

}\subfloat[]{\includegraphics[width=0.22\columnwidth]{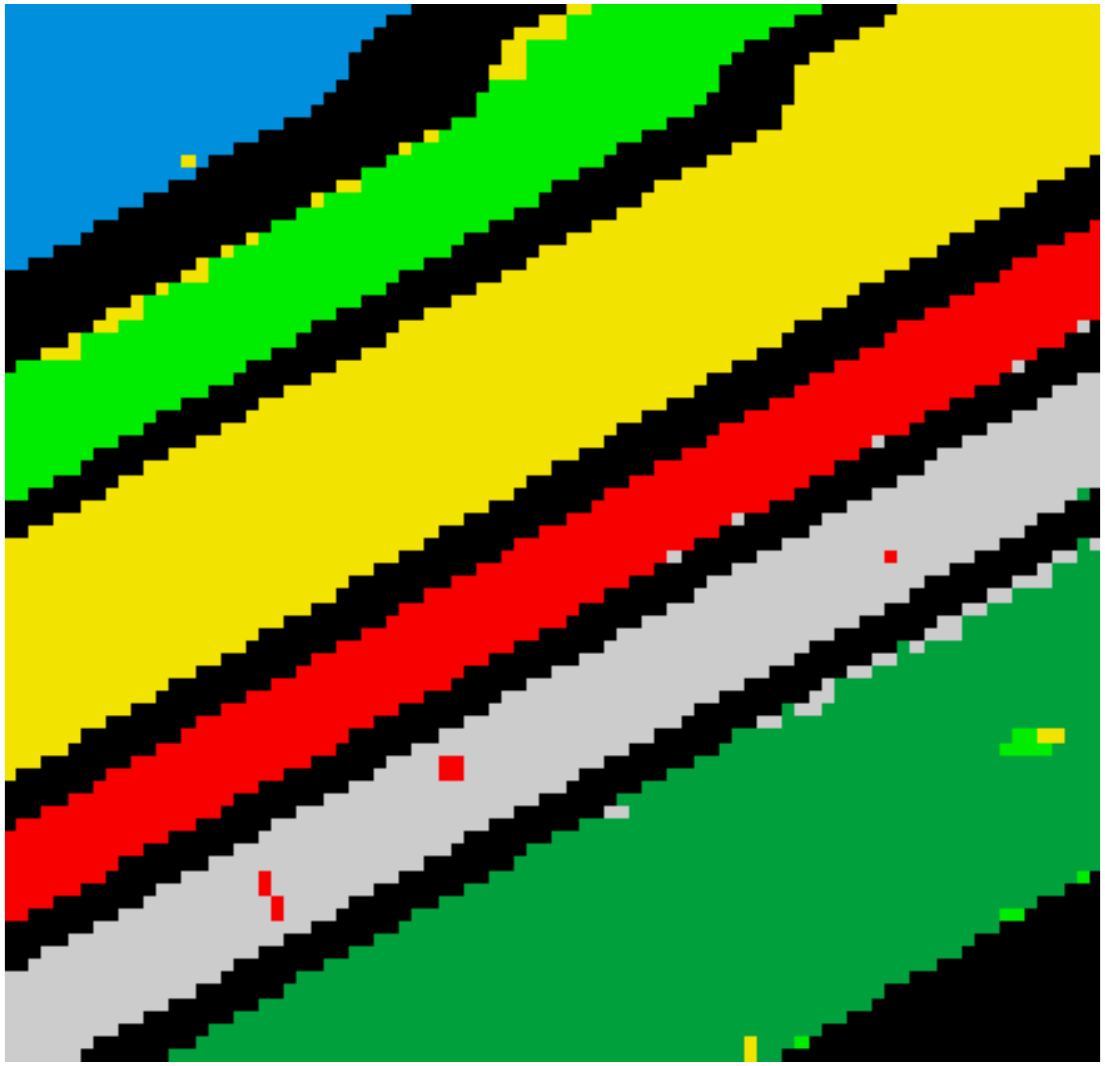}

}\subfloat[]{\includegraphics[width=0.22\columnwidth]{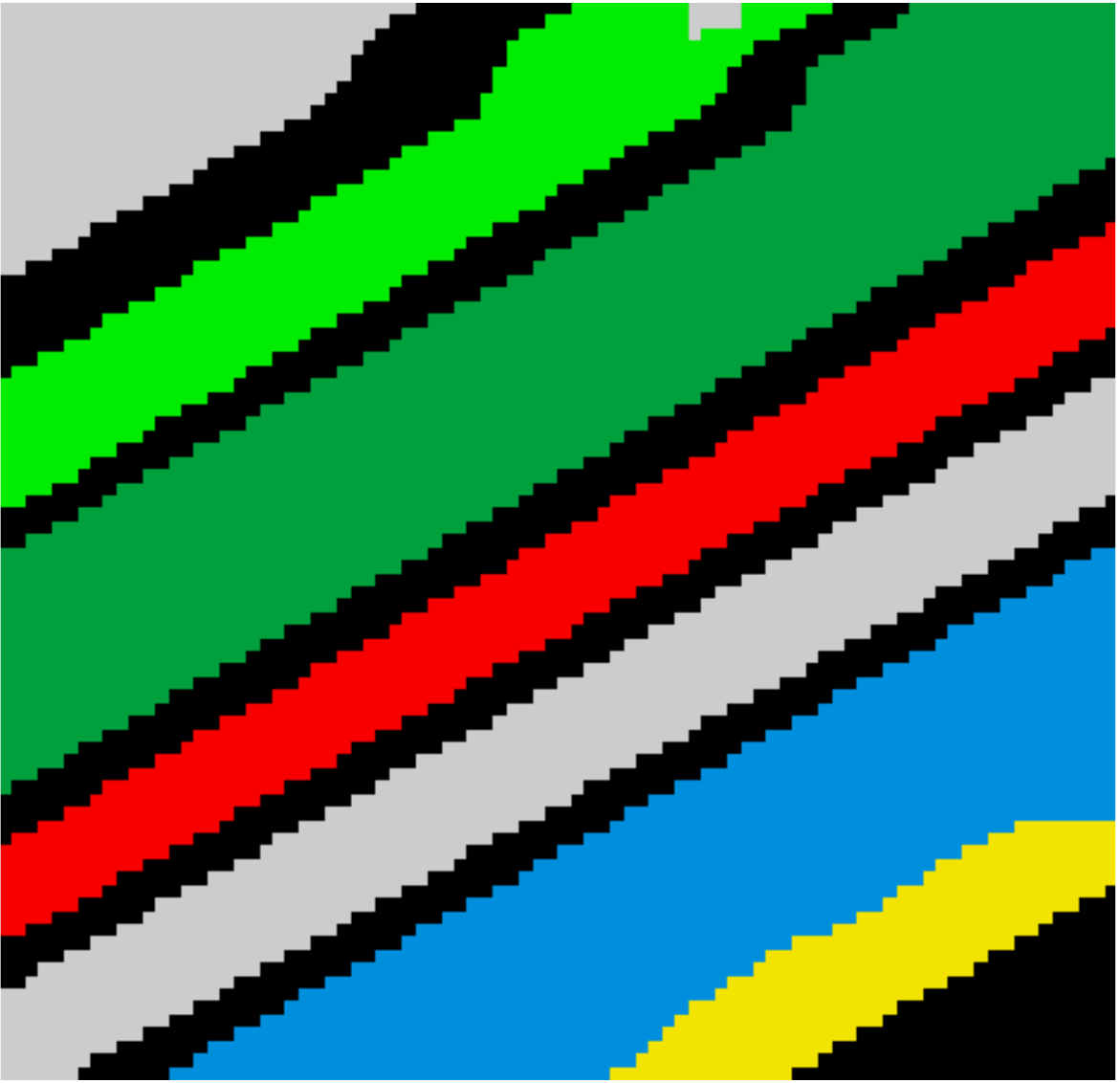}

}\subfloat[]{\includegraphics[width=0.22\columnwidth]{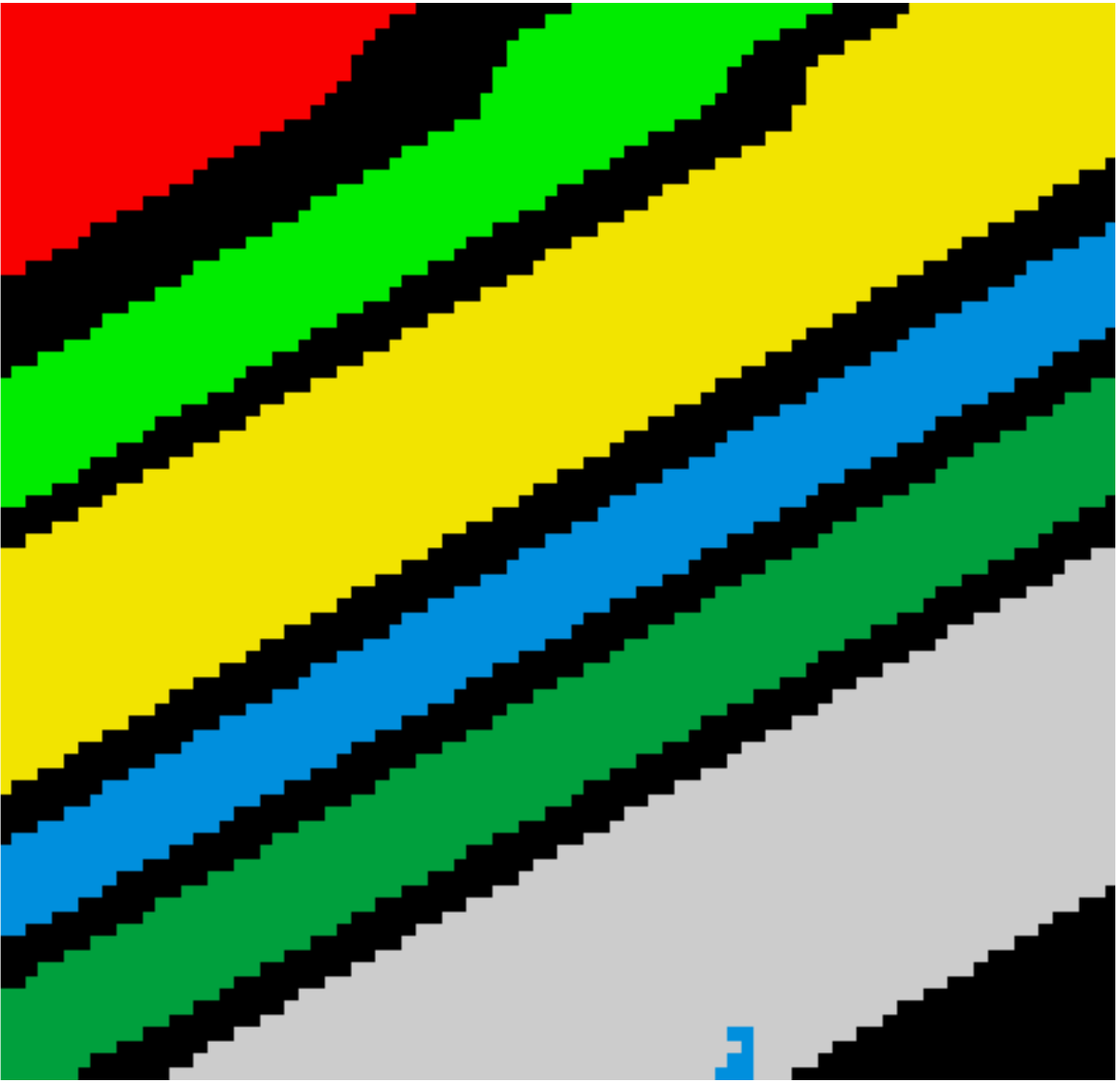}

}\subfloat[]{\includegraphics[width=0.22\columnwidth]{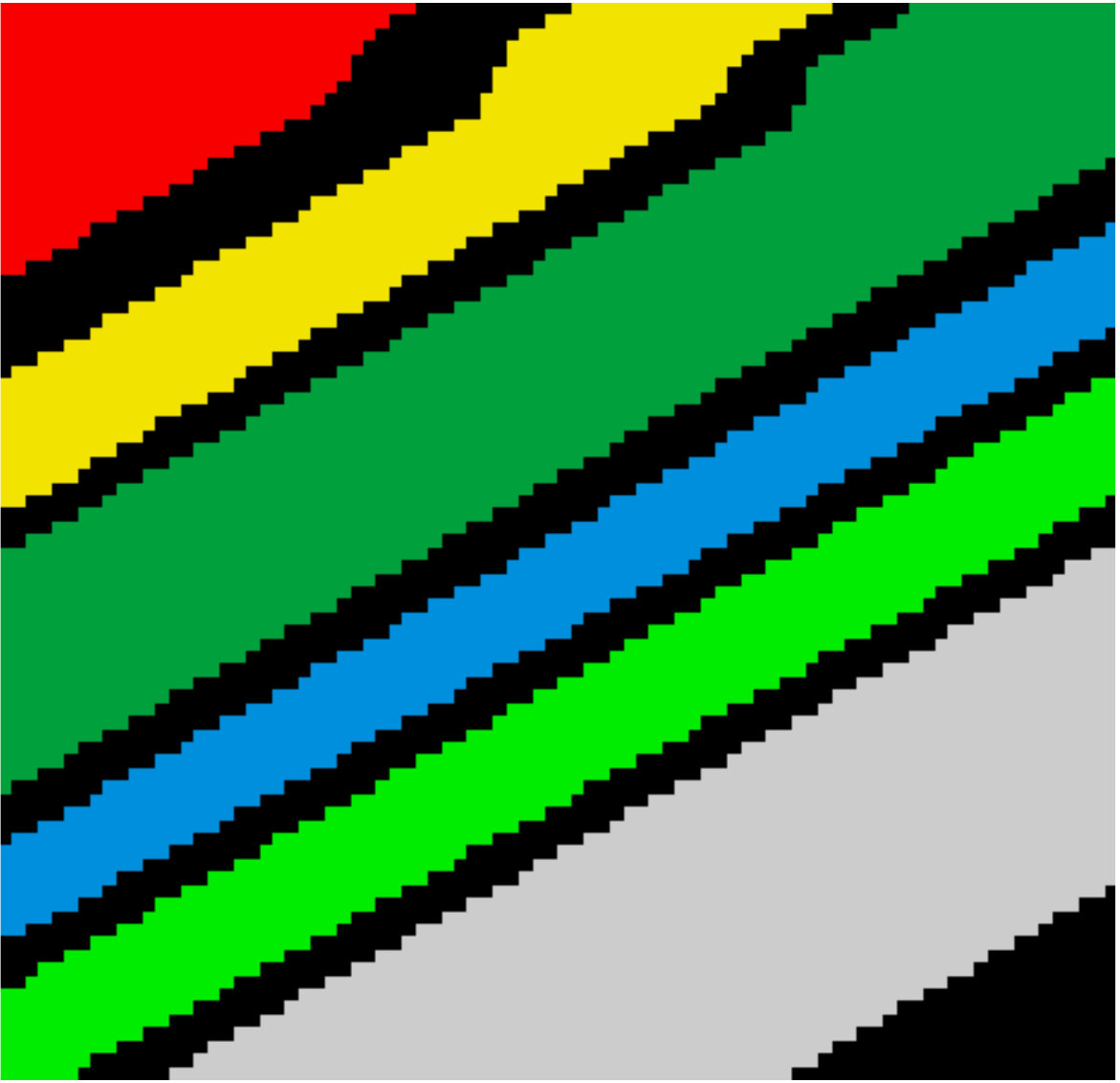}

}
\par\end{centering}
\caption{Clustering results obtained by different methods on the SalinasA dataset:
(a) Ground truth, (b) SC $68.06\%$, (c) SSC $76.66\%$, (d) $\ell_{2}$-SSC
$64.12\%$, (e) LRSC $56.13\%$, (f) RMMF $98.20\%$, (g) EDSC $87.02\%$,
(h) EGCSC $99.85\%$, and (i) EKGCSC $100\%$. \label{fig:Clustering-results-SalinasA-1}}
\end{figure*}
\par\end{center}

\begin{center}
\begin{figure*}[tbh]
\begin{centering}
\subfloat[]{\includegraphics[width=0.22\columnwidth]{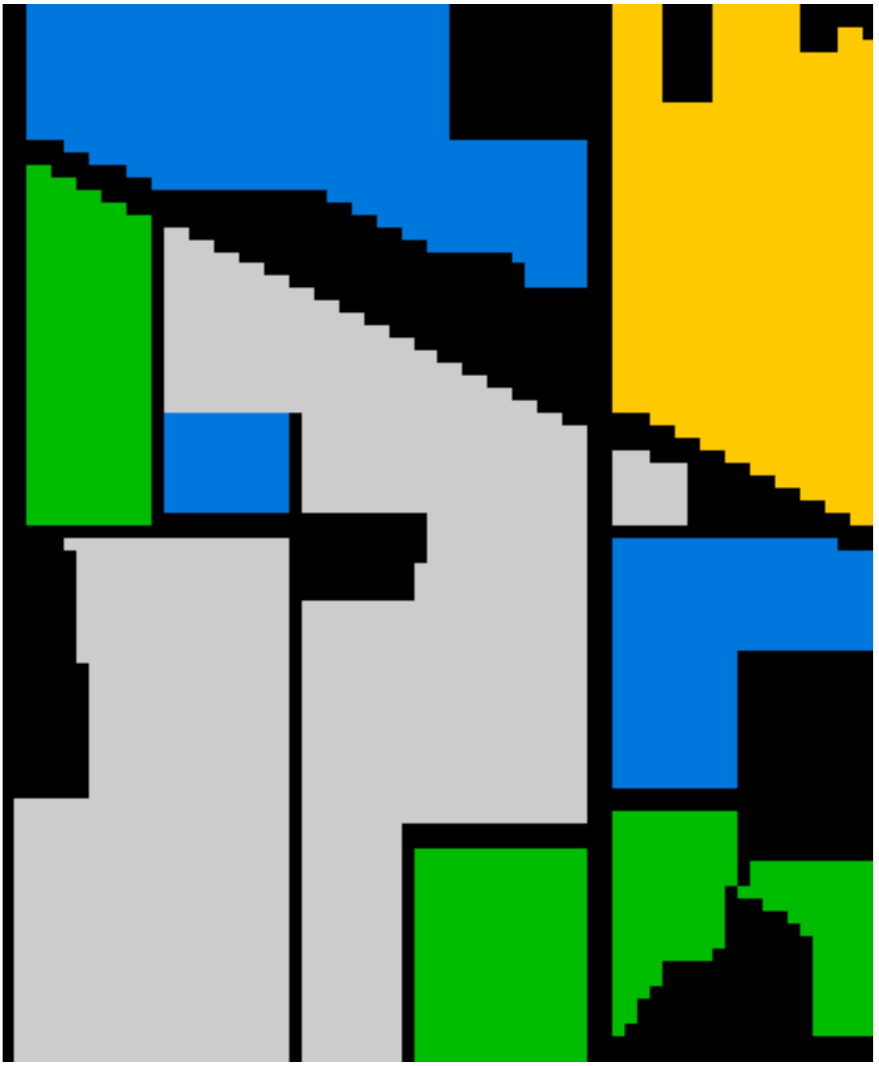}

}\subfloat[]{\includegraphics[width=0.22\columnwidth]{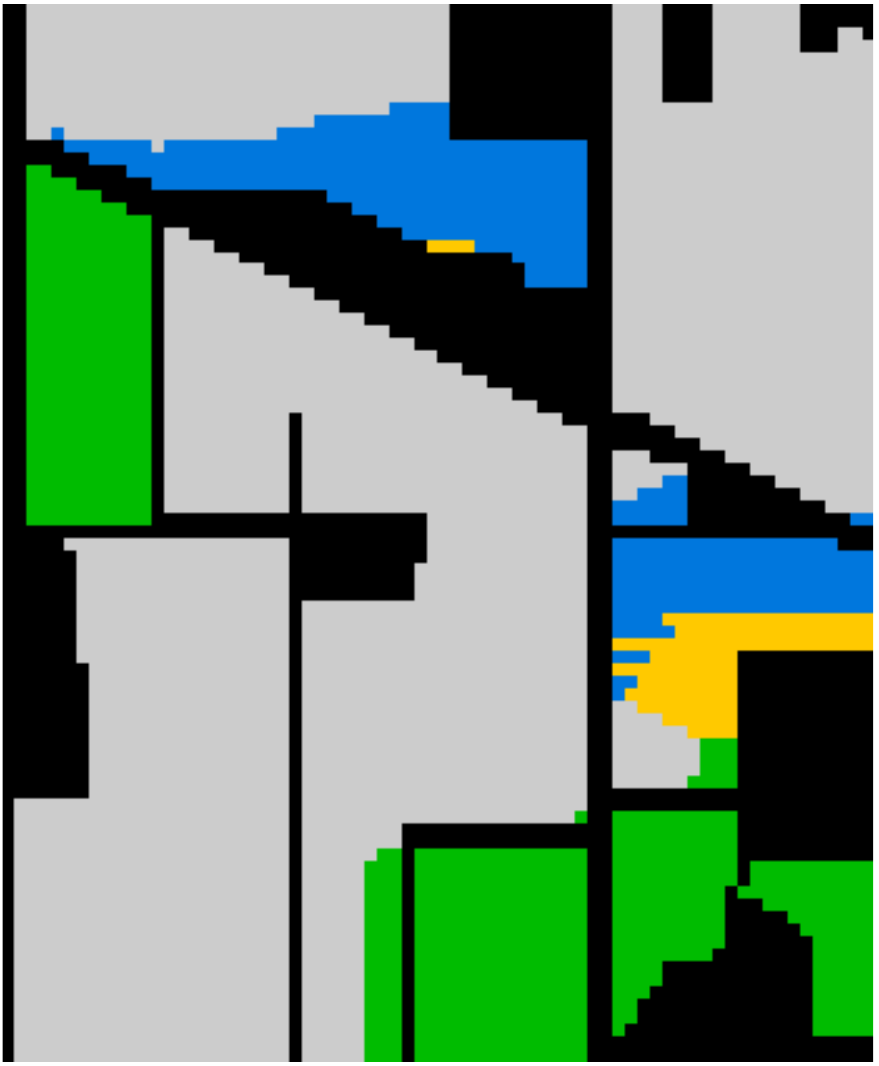}

}\subfloat[]{\includegraphics[width=0.22\columnwidth]{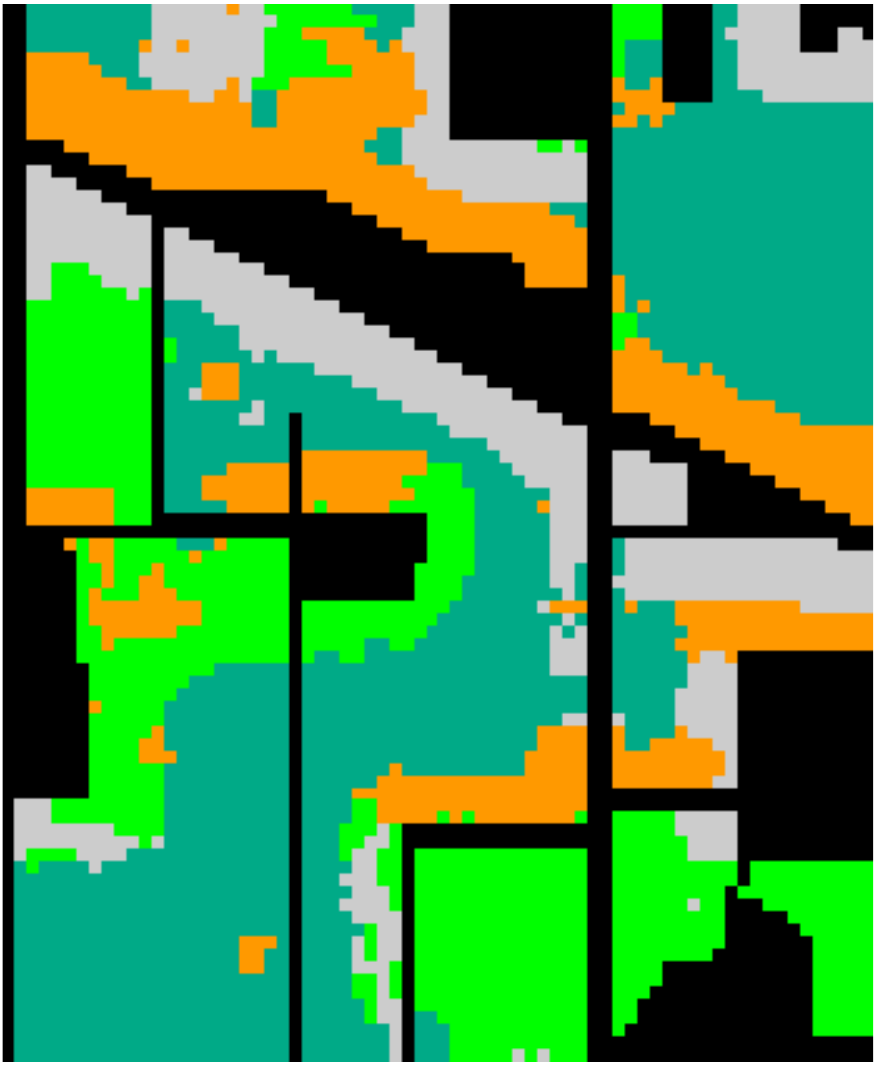}

}\subfloat[]{\includegraphics[width=0.22\columnwidth]{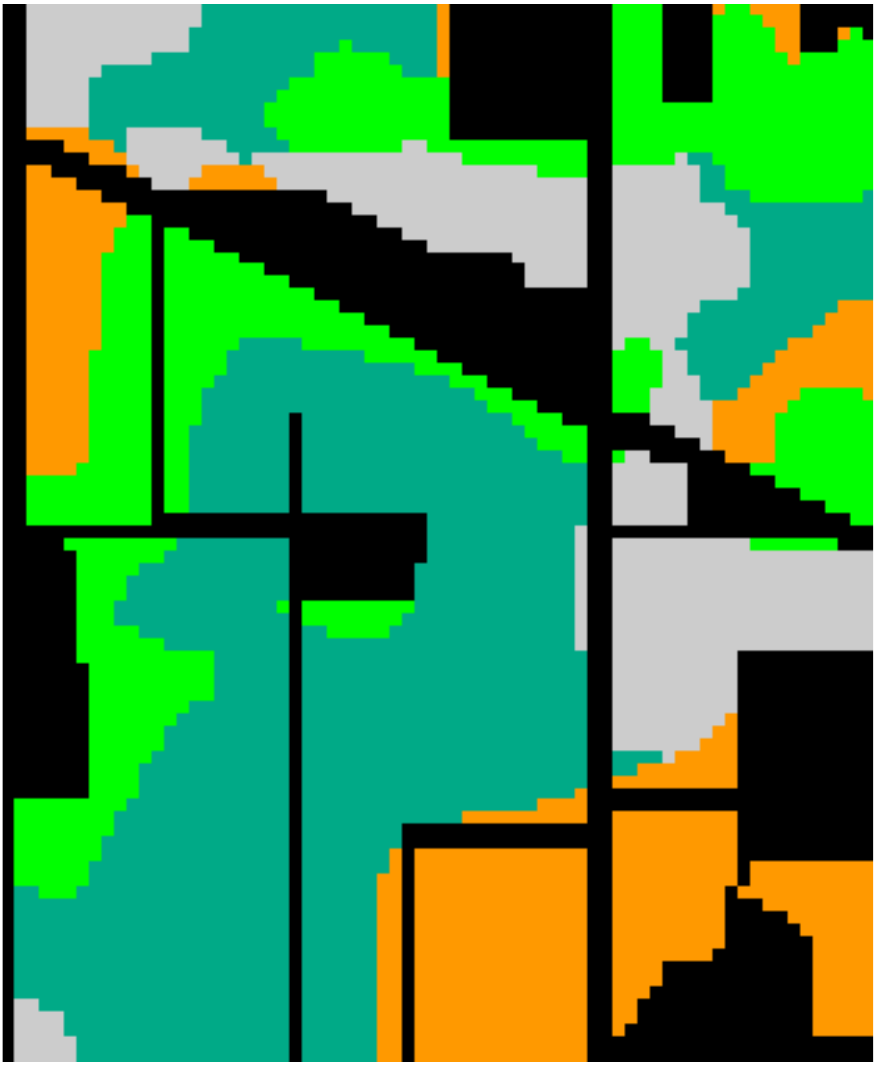}

}\subfloat[]{\includegraphics[width=0.22\columnwidth]{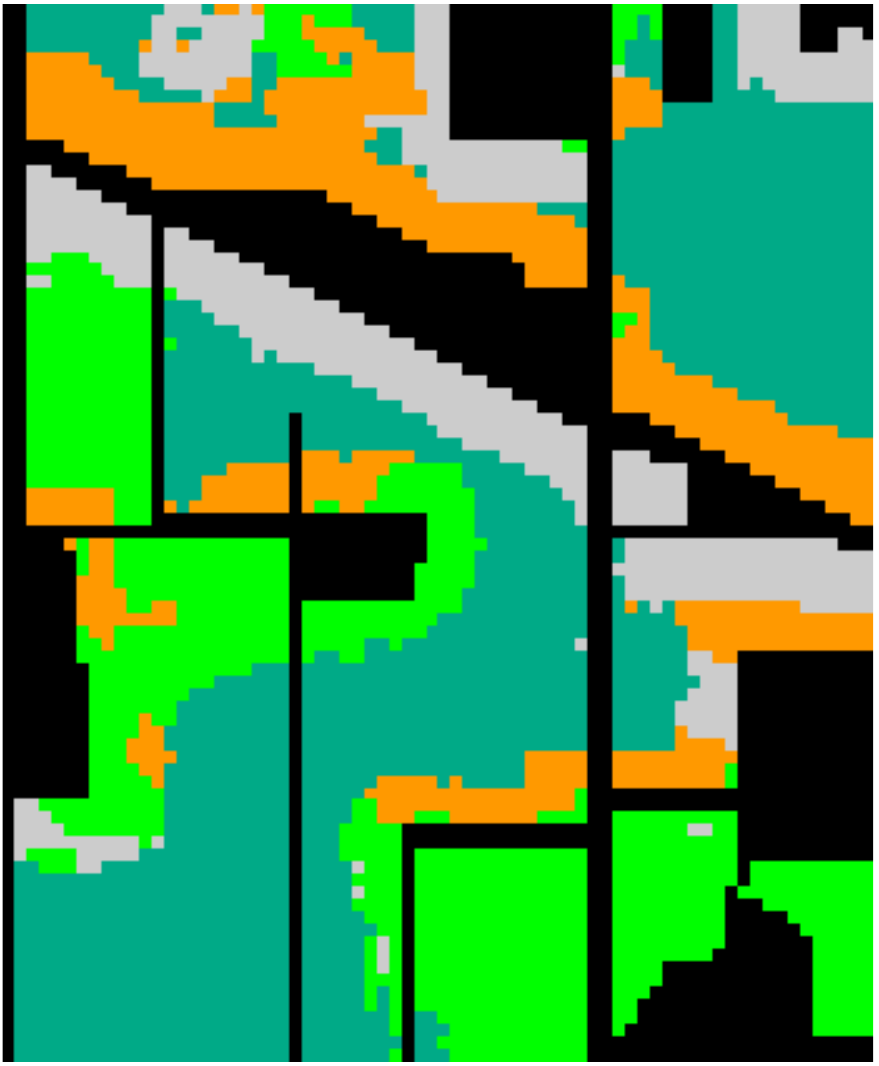}

}\subfloat[]{\includegraphics[width=0.22\columnwidth]{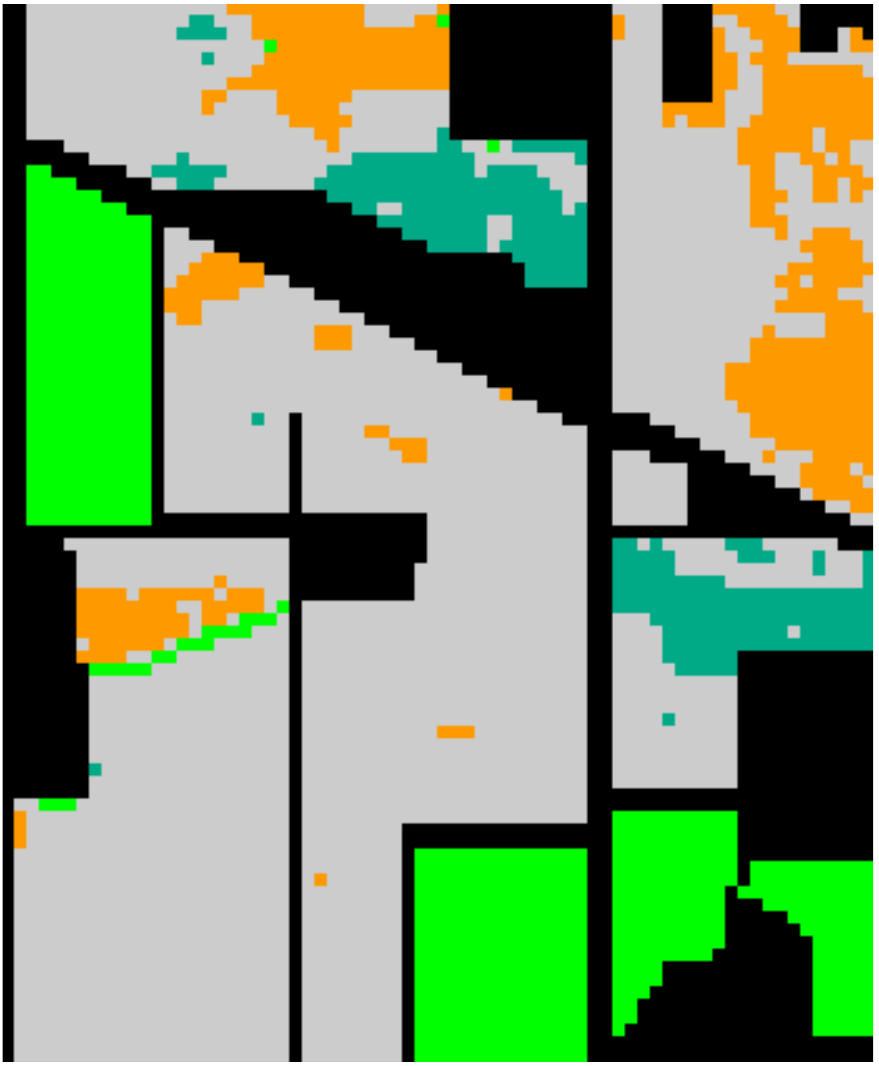}

}\subfloat[]{\includegraphics[width=0.22\columnwidth]{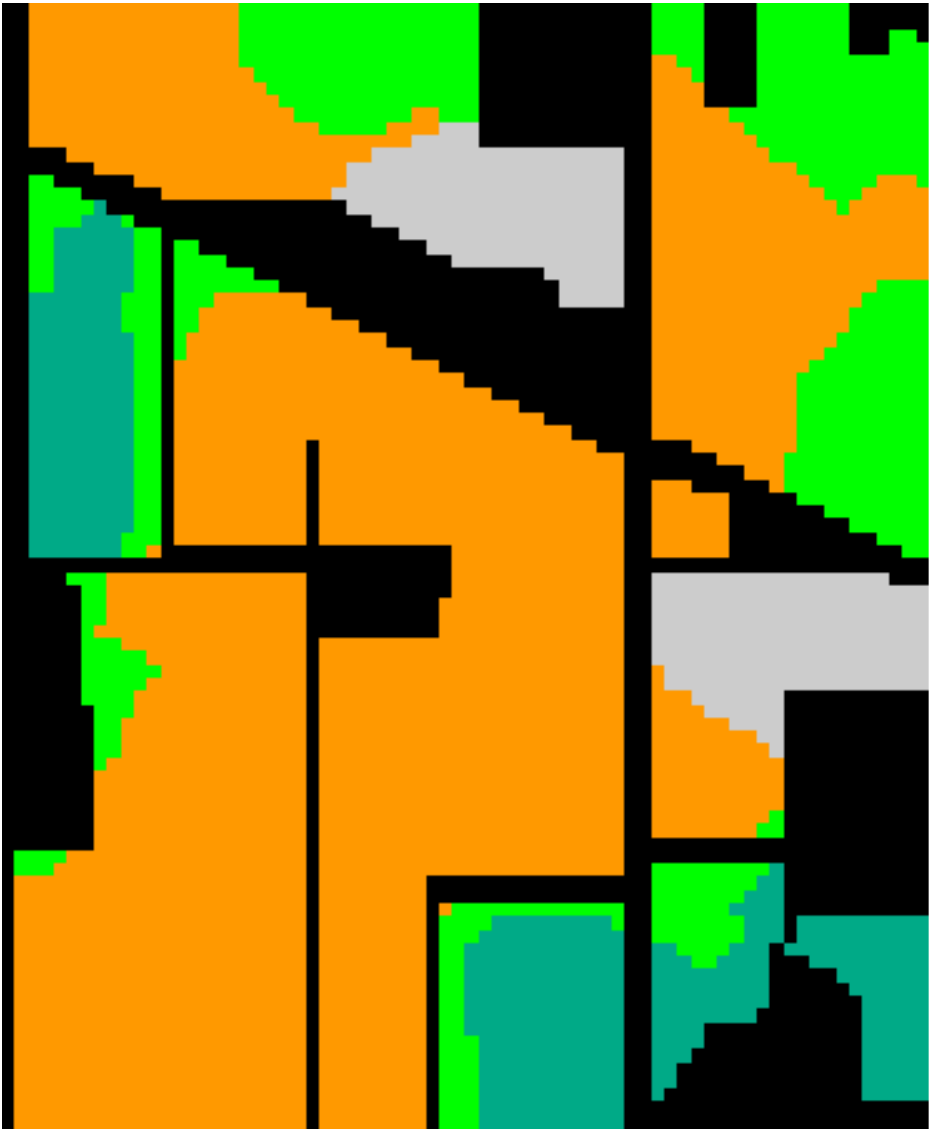}

}\subfloat[]{\includegraphics[width=0.22\columnwidth]{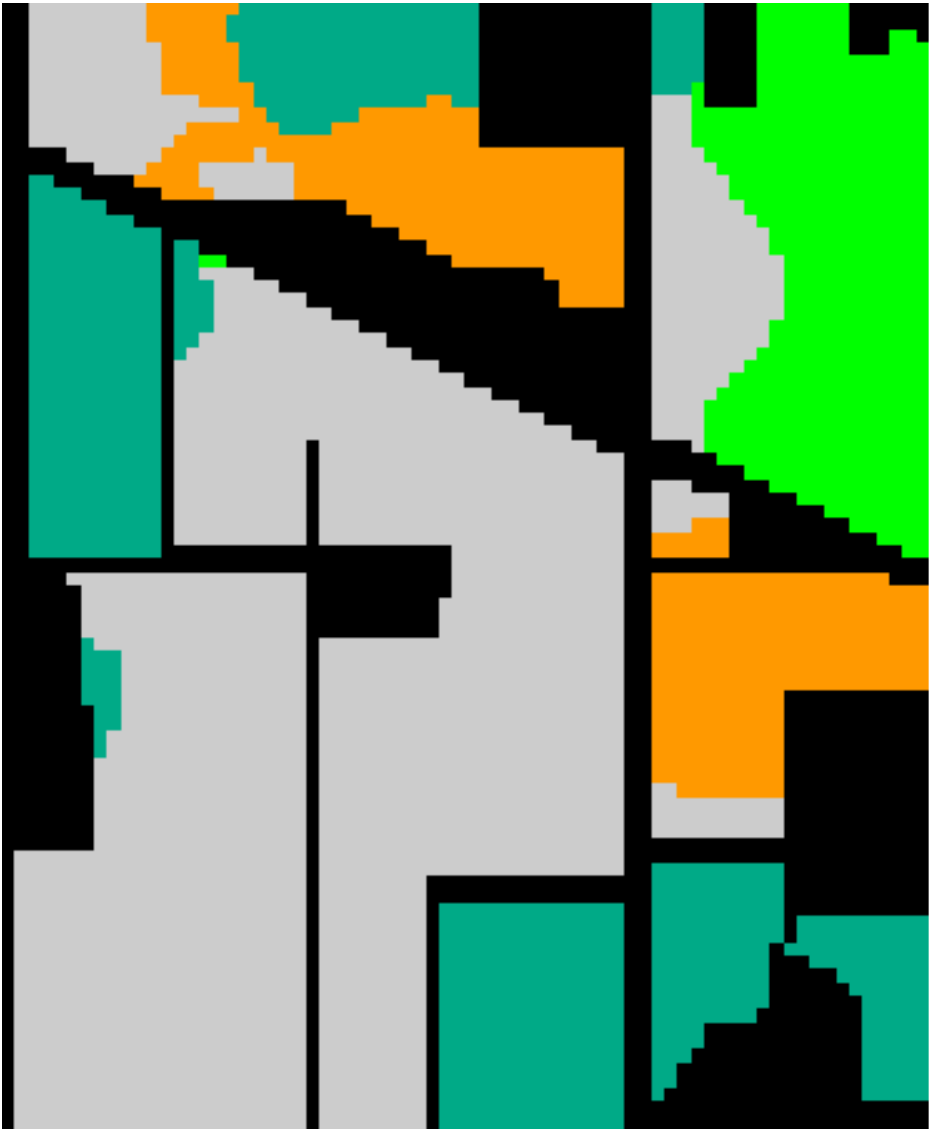}

}\subfloat[]{\includegraphics[width=0.22\columnwidth]{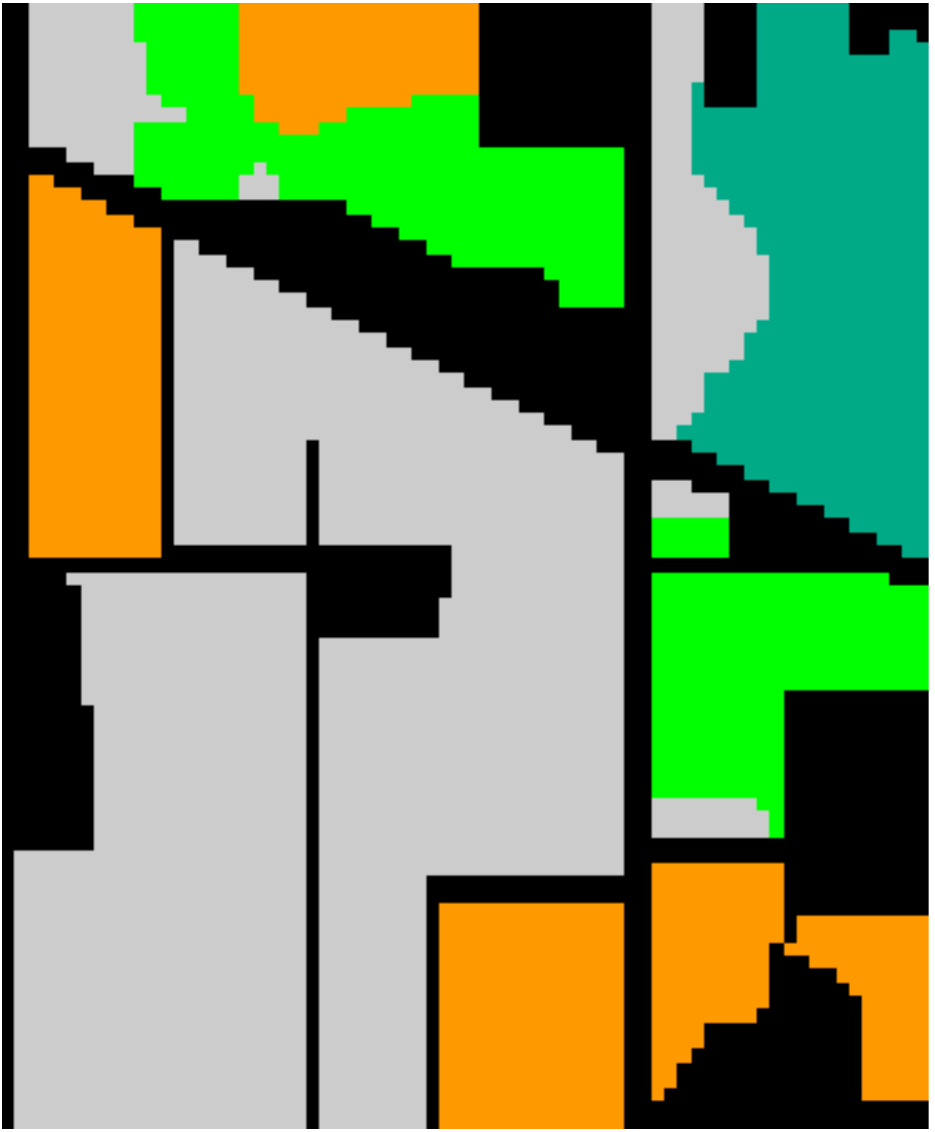}

}
\par\end{centering}
\caption{Clustering results obtained by different methods on the Indian Pines
dataset: (a) Ground truth, (b) SC $68.41\%$, (c) SSC $49.37\%$,
(d) $\ell_{2}$-SSC $66.45\%$, (e) LRSC $51.42\%$, (f) RMMF $71.20\%$,
(g) EDSC $71.26\%$, (h) EGCSC $84.83\%$, and (i) EKGCSC $87.61\%$.
\label{fig:Clustering-results-IndiP-1}}
\end{figure*}
\par\end{center}

\begin{center}
\begin{figure*}[tbh]
\begin{centering}
\subfloat[]{\includegraphics[width=0.22\columnwidth]{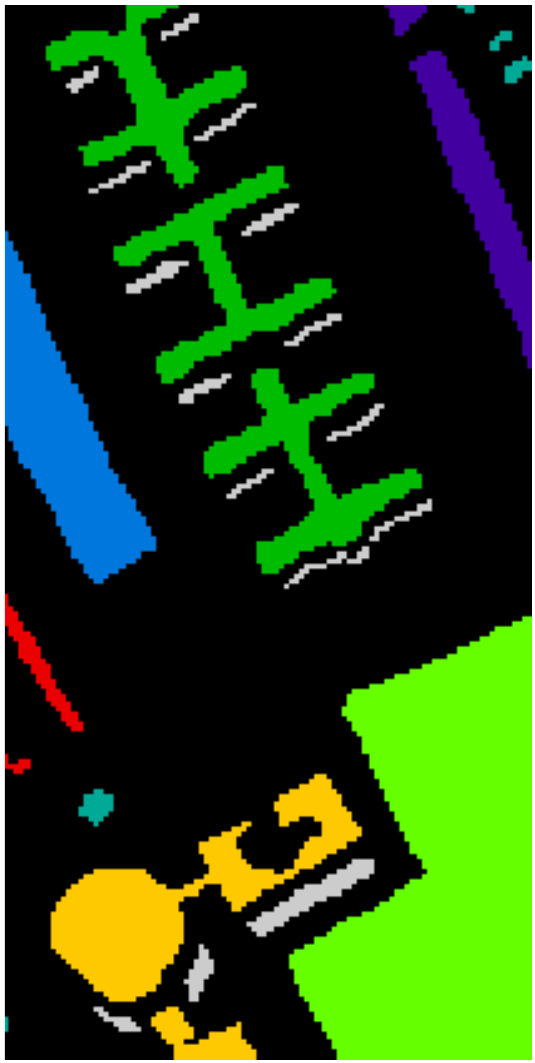}

}\subfloat[]{\includegraphics[width=0.22\columnwidth]{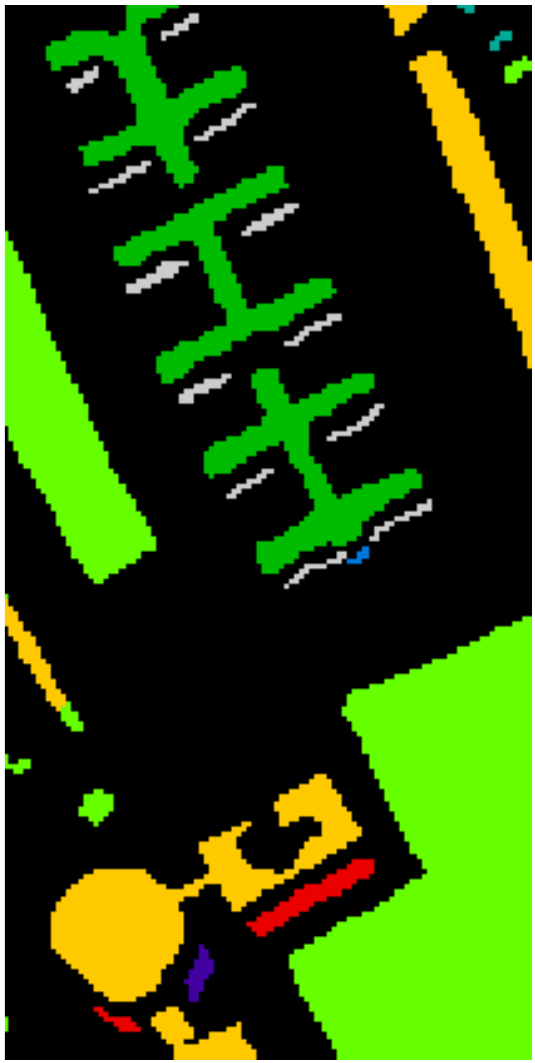}

}\subfloat[]{\includegraphics[width=0.22\columnwidth]{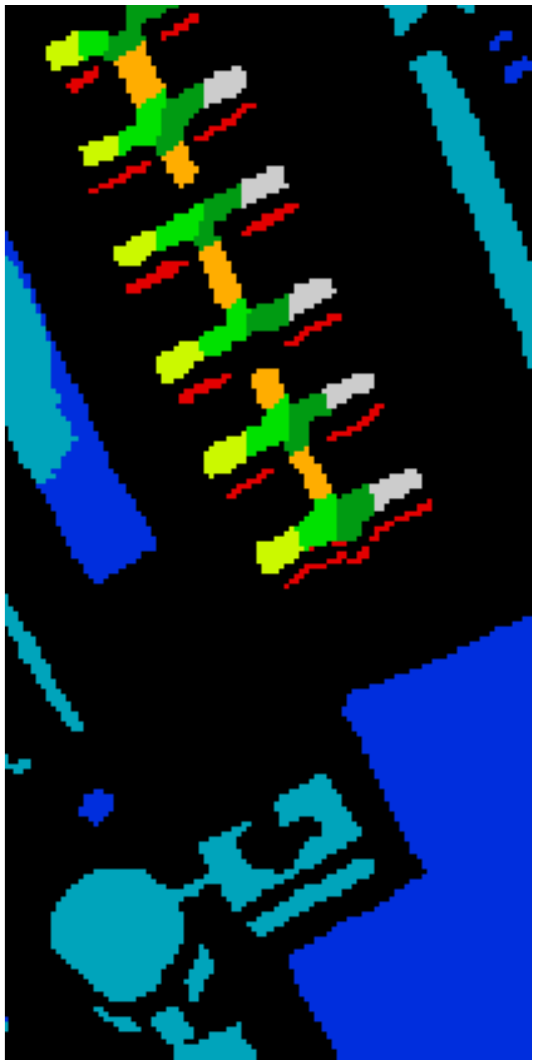}

}\subfloat[]{\includegraphics[width=0.22\columnwidth]{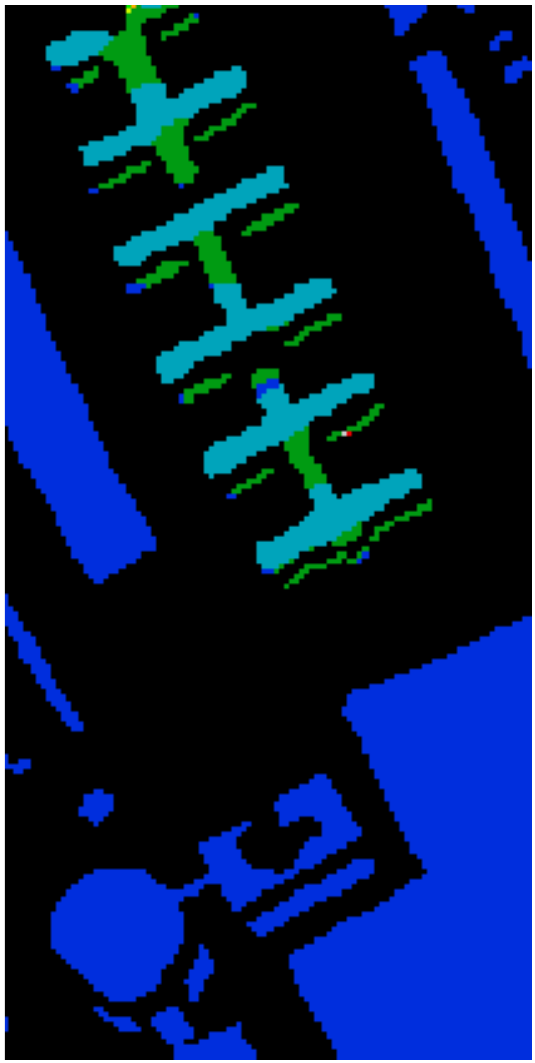}

}\subfloat[]{\includegraphics[width=0.22\columnwidth]{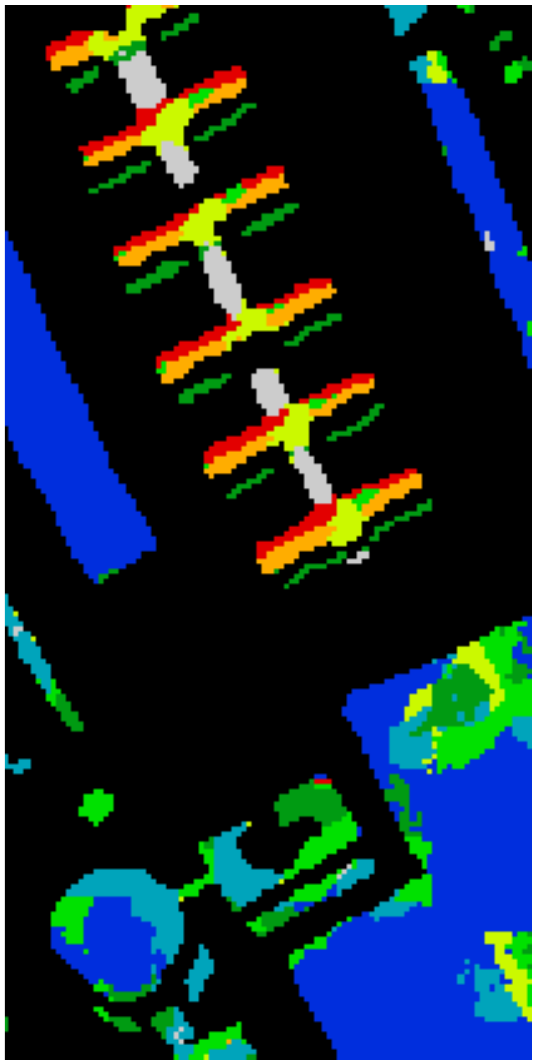}

}\subfloat[]{\includegraphics[width=0.22\columnwidth]{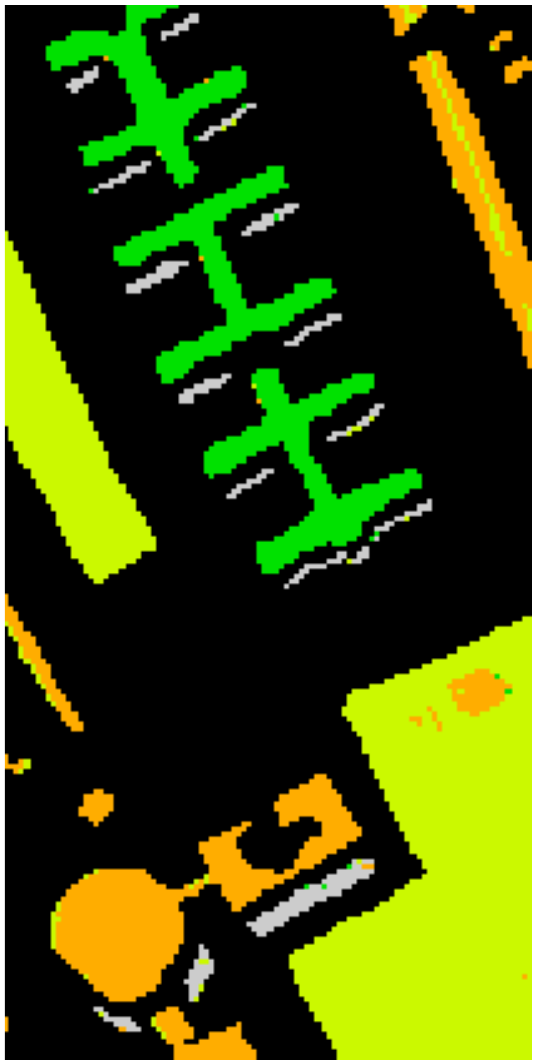}

}\subfloat[]{\includegraphics[width=0.22\columnwidth]{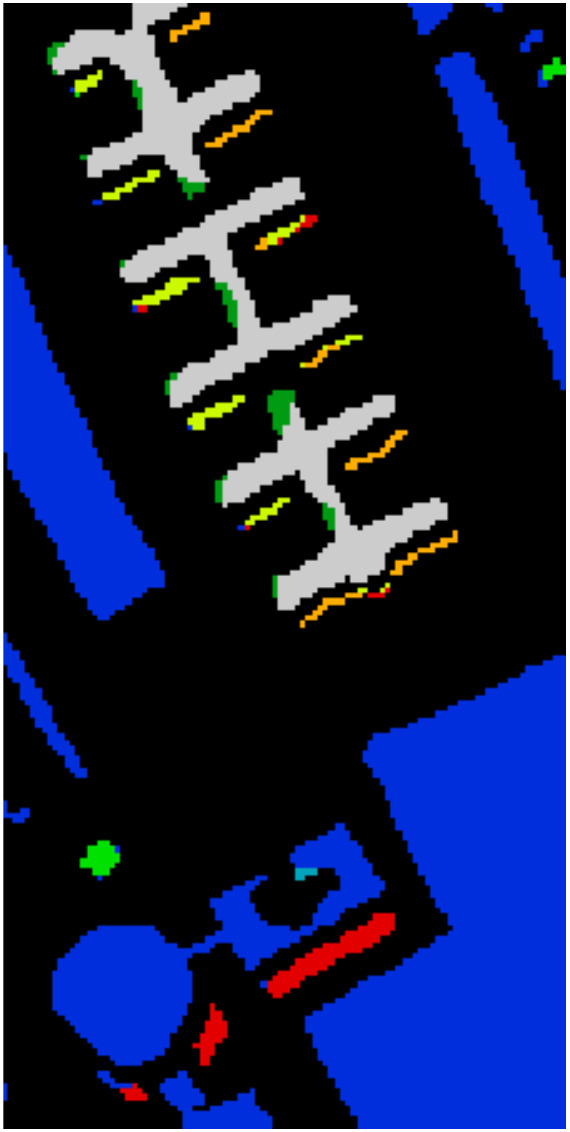}

}\subfloat[]{\includegraphics[width=0.22\columnwidth]{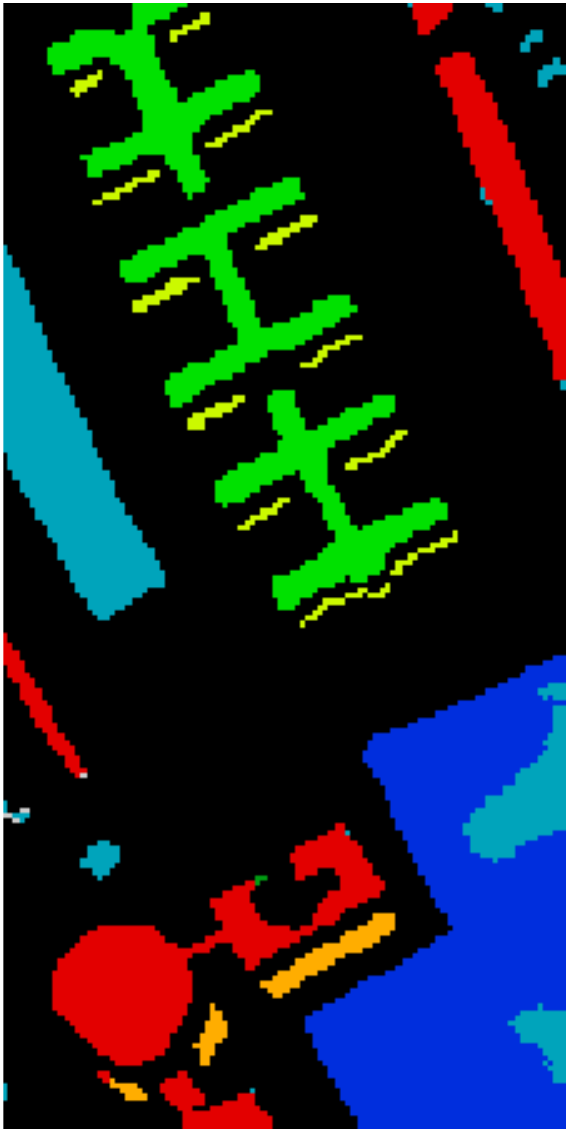}

}\subfloat[]{\includegraphics[width=0.22\columnwidth]{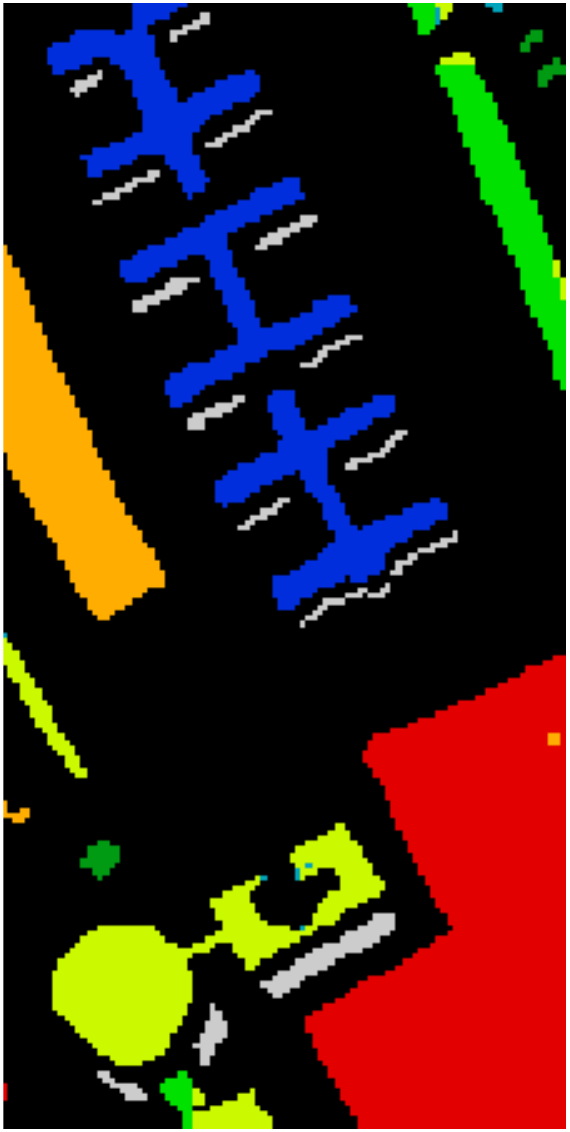}

}
\par\end{centering}
\caption{Clustering results obtained by different methods on the Pavia University
dataset: (a) Ground truth, (b) SC $76.91\%$, (c) SSC $64.46\%$,
(d) $\ell_{2}$-SSC $58.42\%$, (e) LRSC $43.26\%$, (f) RMMF $77.04\%$,
(g) EDSC $61.75\%$, (h) EGCSC $84.42\%$, and (i) EKGCSC $97.36\%$.
\label{fig:Clustering-results-PaviU-1}}
\end{figure*}
\par\end{center}

To visually observe the clustering results, we visualize the clustering
maps of different clustering methods in Fig. \ref{fig:Clustering-results-SalinasA-1}-\ref{fig:Clustering-results-PaviU-1}.
Since the source codes of S$^{4}$C and UBL have not been released,
their class maps are not included in the figures but it does not affect
the analysis. Notice that the color of the same class may be variant
in different class maps, which is because label values may be permuted
by different clustering methods. Observed from Fig. \ref{fig:Clustering-results-SalinasA-1},
the class map obtained by EKGCSC on the SalinasA dataset is in complete
agreement with the ground truth. For the Indian Pines and Pavia University
datasets (i.e., Fig. \ref{fig:Clustering-results-IndiP-1} and Fig.
\ref{fig:Clustering-results-PaviU-1}), EKGCSC shows the best class
maps that are closest to the ground truths. Compared with the other
competitors, EGCSC shows better class maps. While the class maps obtained
by the other methods ( e.g., SSC, LRSC, and EDSC) contain relatively
more noisy points caused by misclassification. Briefly, the results
demonstrate the effectiveness and superiority of the proposed GCSC
framework.

\subsubsection{Visualization of The Learned Affinity Matrix }
\begin{center}
\par\end{center}

\begin{center}
\begin{figure}[tbh]
\begin{centering}
\subfloat[EGCSC]{\includegraphics[width=0.5\columnwidth]{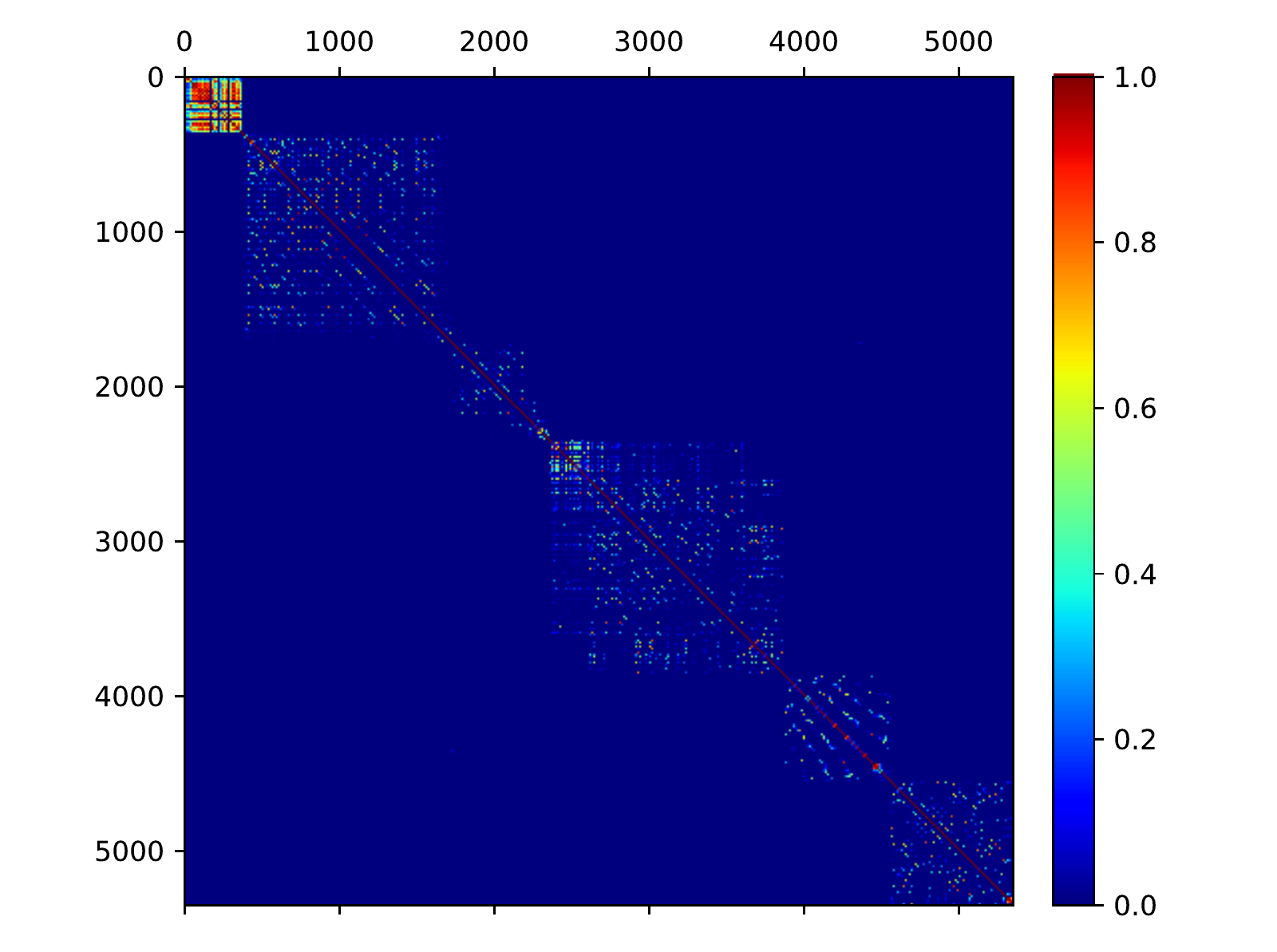}

}\subfloat[EKGCSC]{\includegraphics[width=0.5\columnwidth]{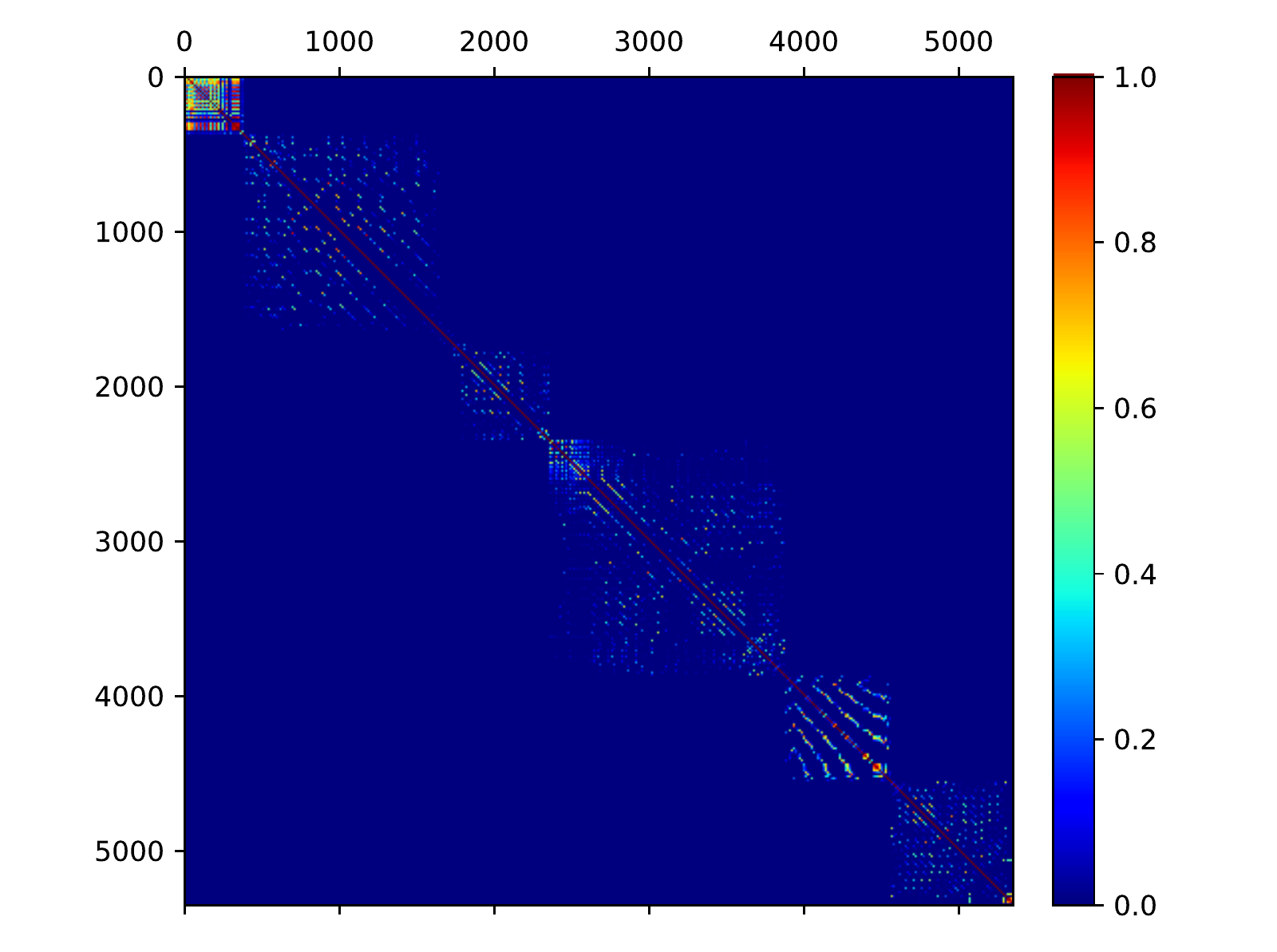}

}
\par\end{centering}
\begin{centering}
\subfloat[EGCSC]{\includegraphics[width=0.5\columnwidth]{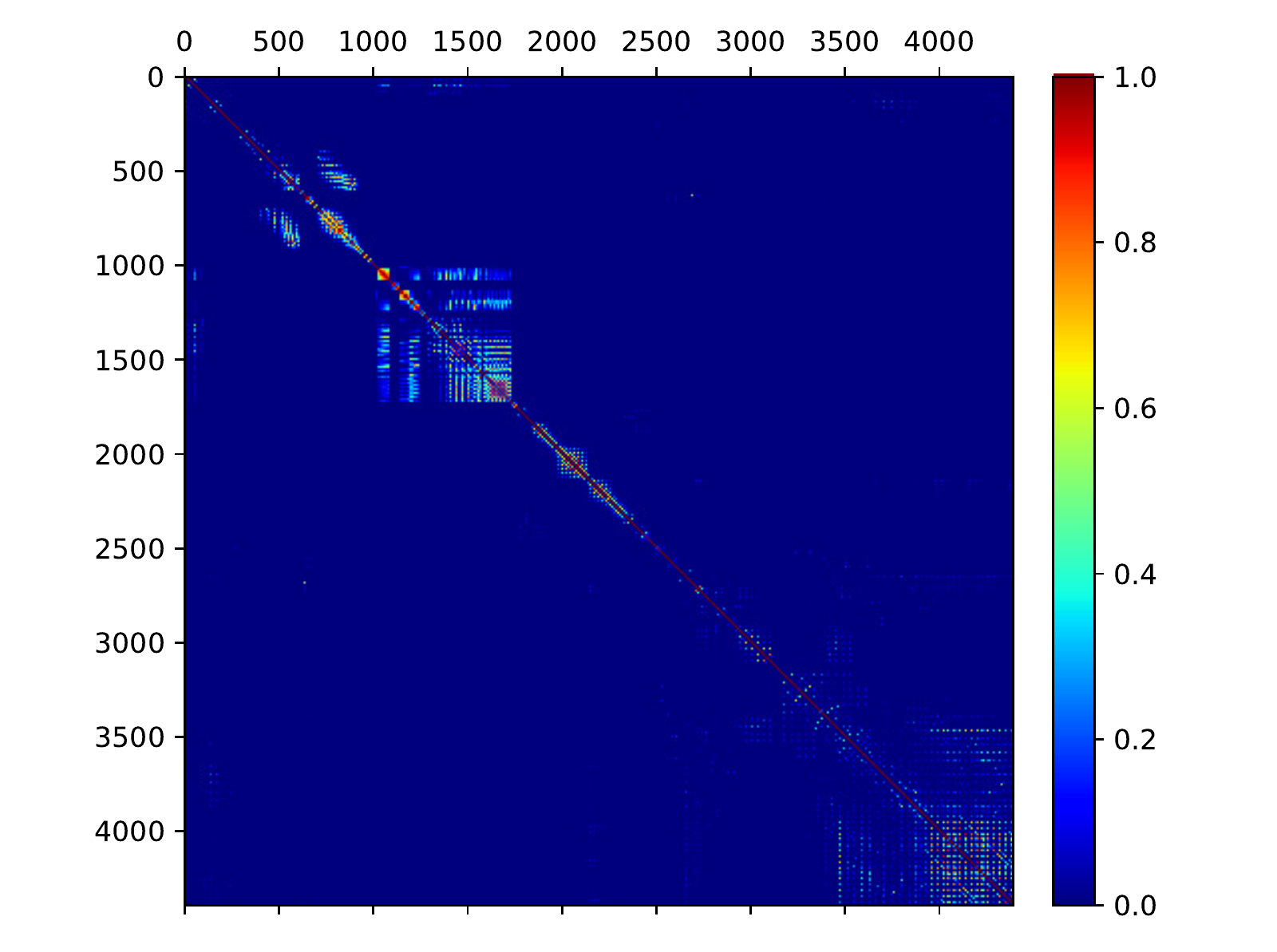}

}\subfloat[EKGCSC]{\includegraphics[width=0.5\columnwidth]{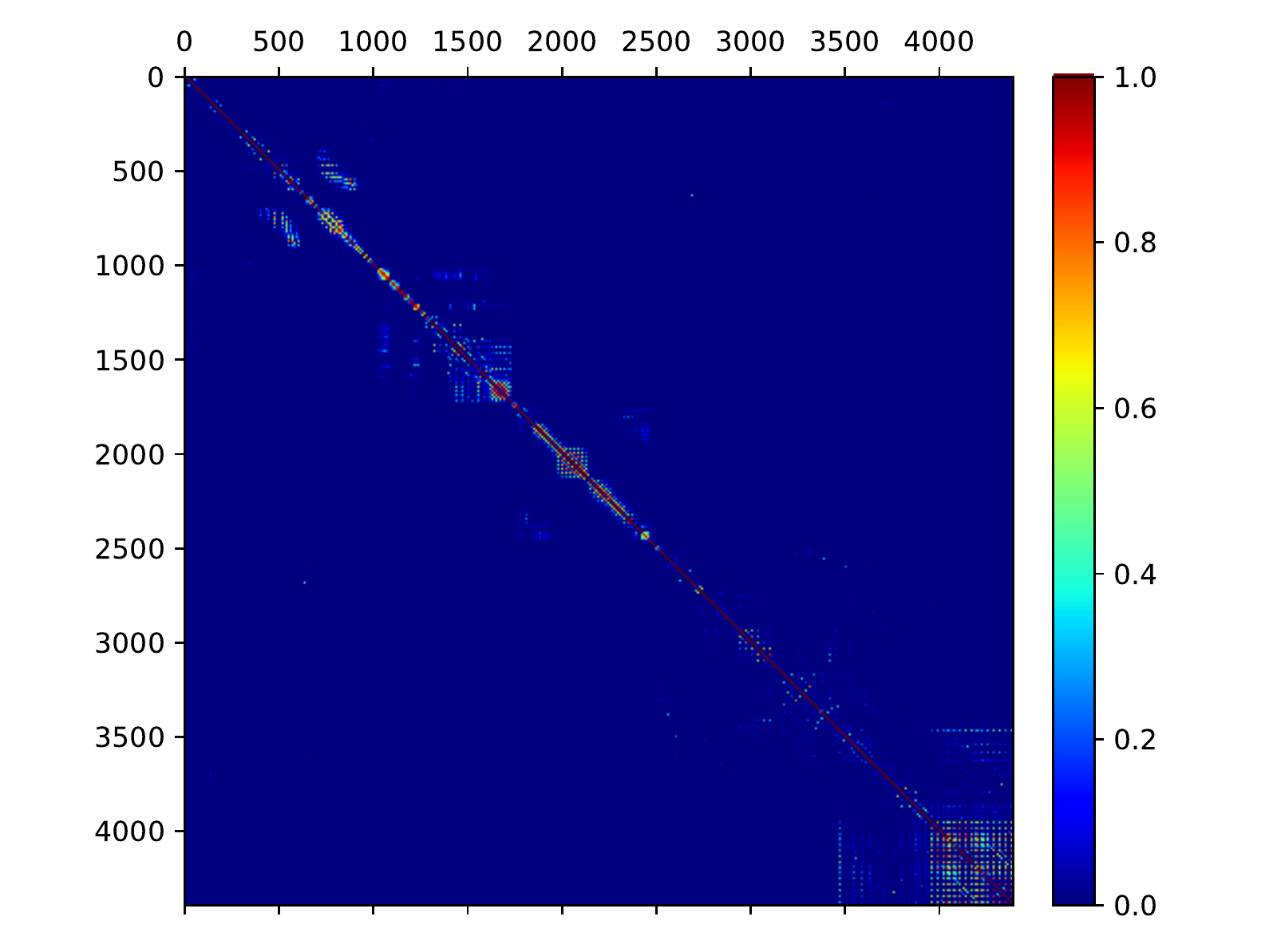}

}
\par\end{centering}
\begin{centering}
\subfloat[EGCSC]{\includegraphics[width=0.5\columnwidth]{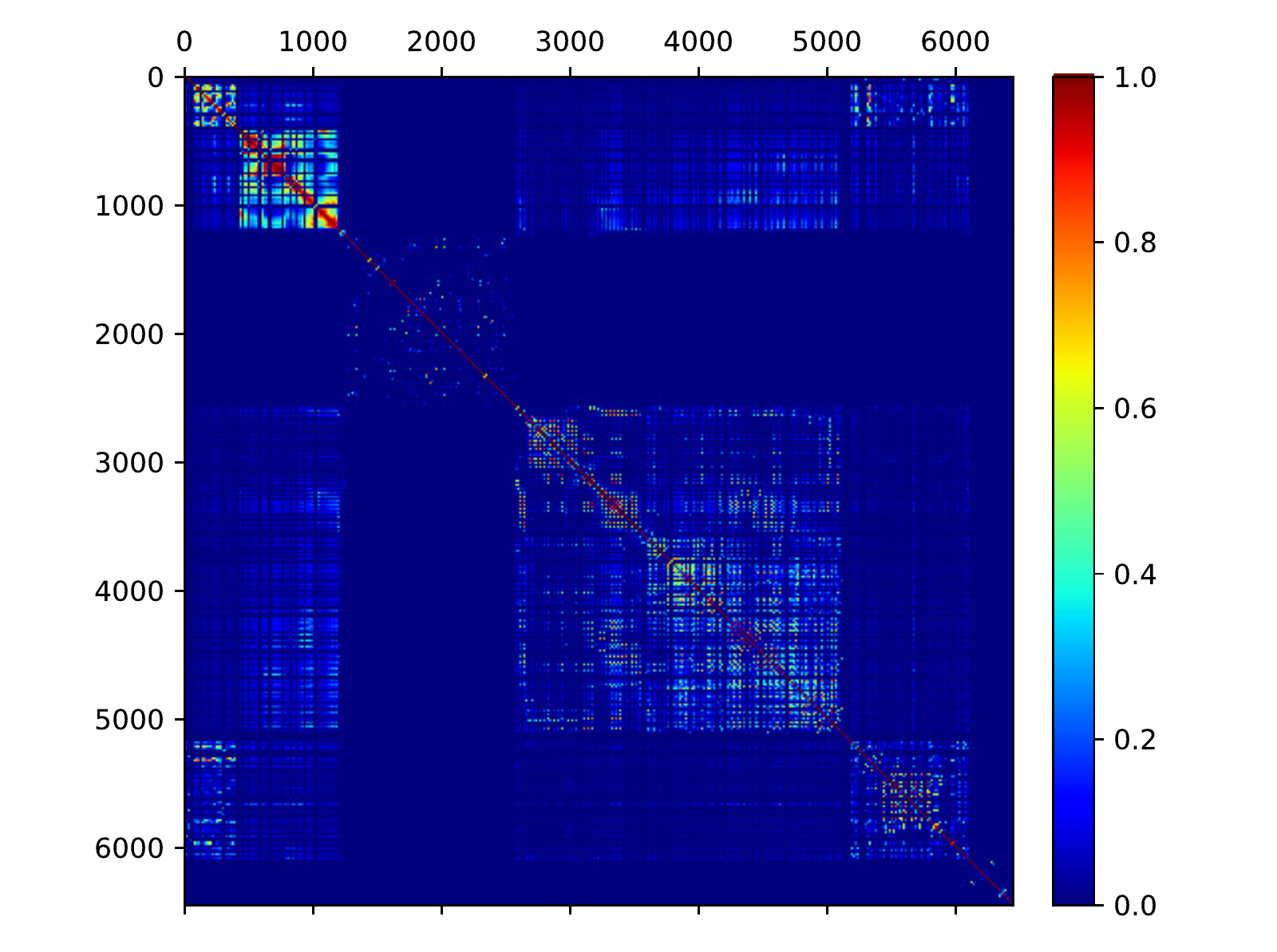}

}\subfloat[EKGCSC]{\includegraphics[width=0.5\columnwidth]{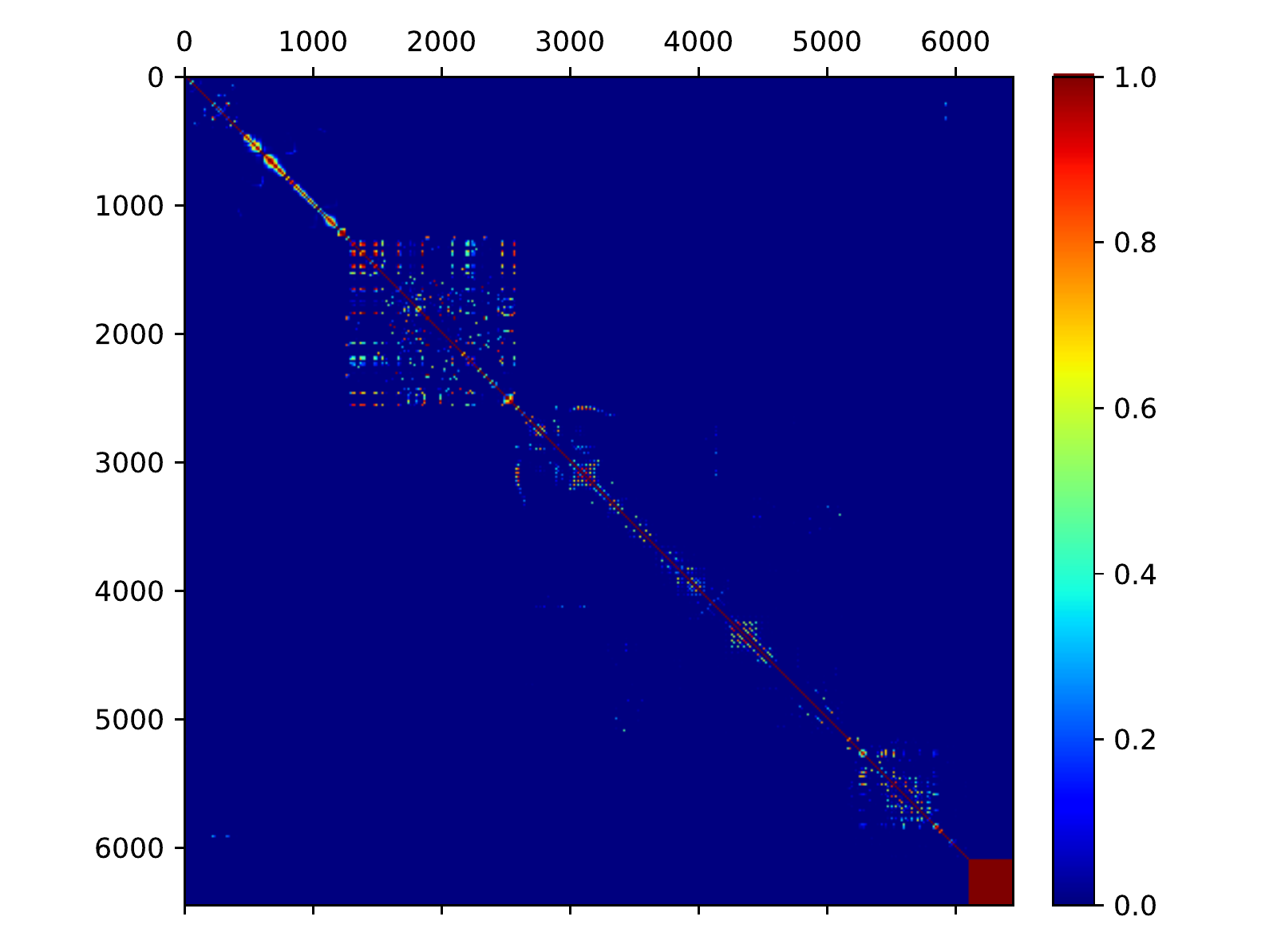}

}
\par\end{centering}
\caption{Visualization of the obtained affinity matrix of EGCSC and EKGCSC
on (a)-(b) SalinasA, (c)-(d)Indian Pines, and (e)-(f) Pavia University
datasets.\label{fig:affinity}}
\end{figure}
\par\end{center}

In Fig. \ref{fig:affinity}, we visualize the affinity matrices learned
by the EGCSC and EKGCSC models. For better presentation, we have re-ordered
data points according to the ground truth before computing the affinity
matrix. In the figures, each column or row of the affinity matrix
denotes the self-representation coefficients that using all data points
to represent the corresponding data point. Therefore, the larger the
coefficient is, the more the corresponding data point contributes
to the reconstruction. Ideally, if a group of data points belongs
to the same cluster, then their self-representation coefficients to
each other will be non-zero, otherwise, they will be zero. Thus, an
ideal affinity matrix is block-diagonal. From Fig. \ref{fig:affinity}
(a)-(f), we can observe that the obtained affinity matrices by both
EGCSC and EKGCSC are sparse and have an apparent block-diagonal structure.
Furthermore, EKGCSC shows better block-structure than EGCSC, which
demonstrates that EKGCSC can more accurately explore the intrinsic
relationships between data points and thus achieve better performance. 

\subsubsection{Impact of $\lambda$ and $k$ }
\begin{center}
\par\end{center}

\begin{center}
\begin{figure}[tbh]
\begin{centering}
\subfloat[EGCSC]{\includegraphics[width=0.5\columnwidth]{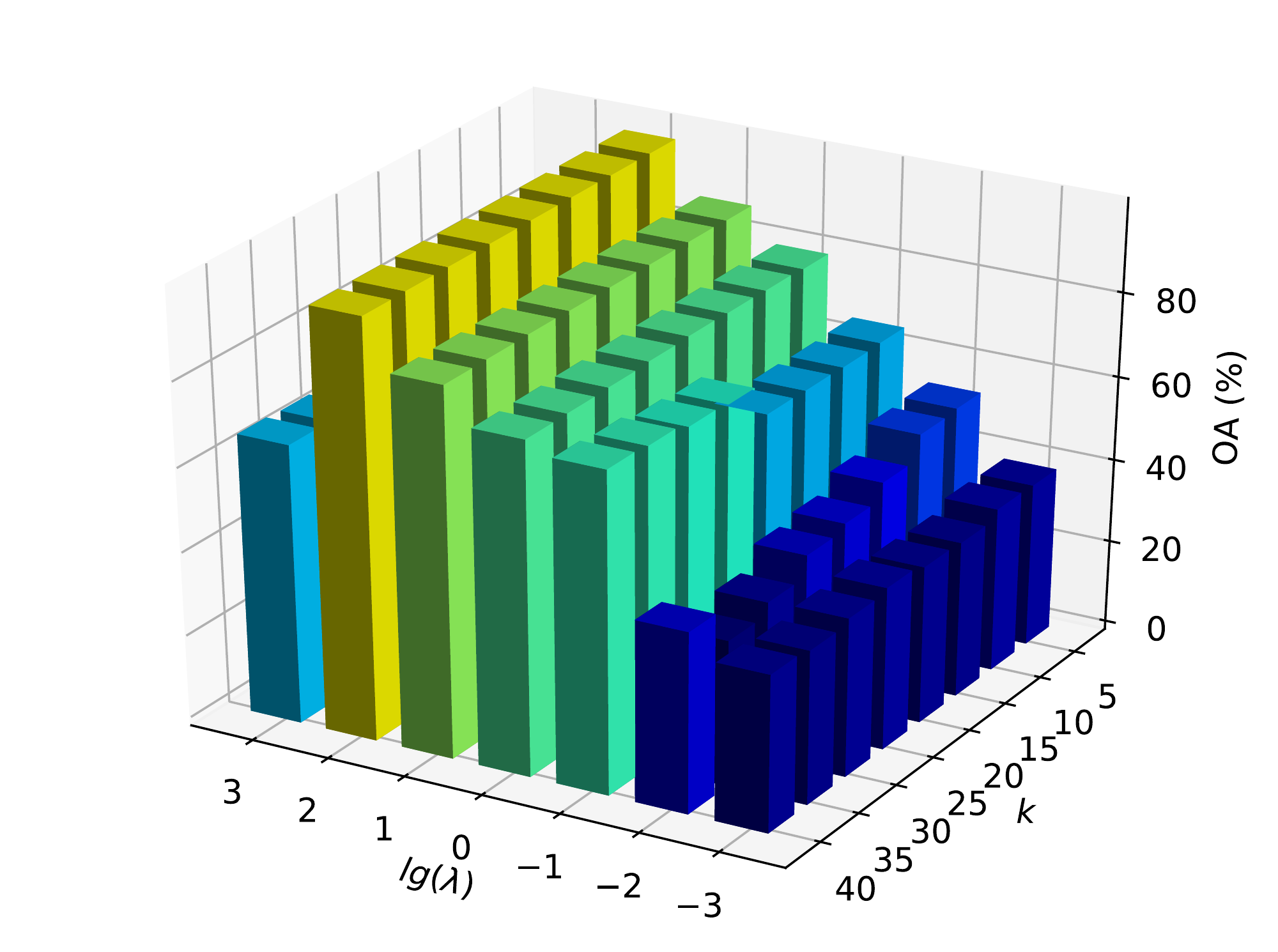}

}\subfloat[EKGCSC]{\includegraphics[width=0.5\columnwidth]{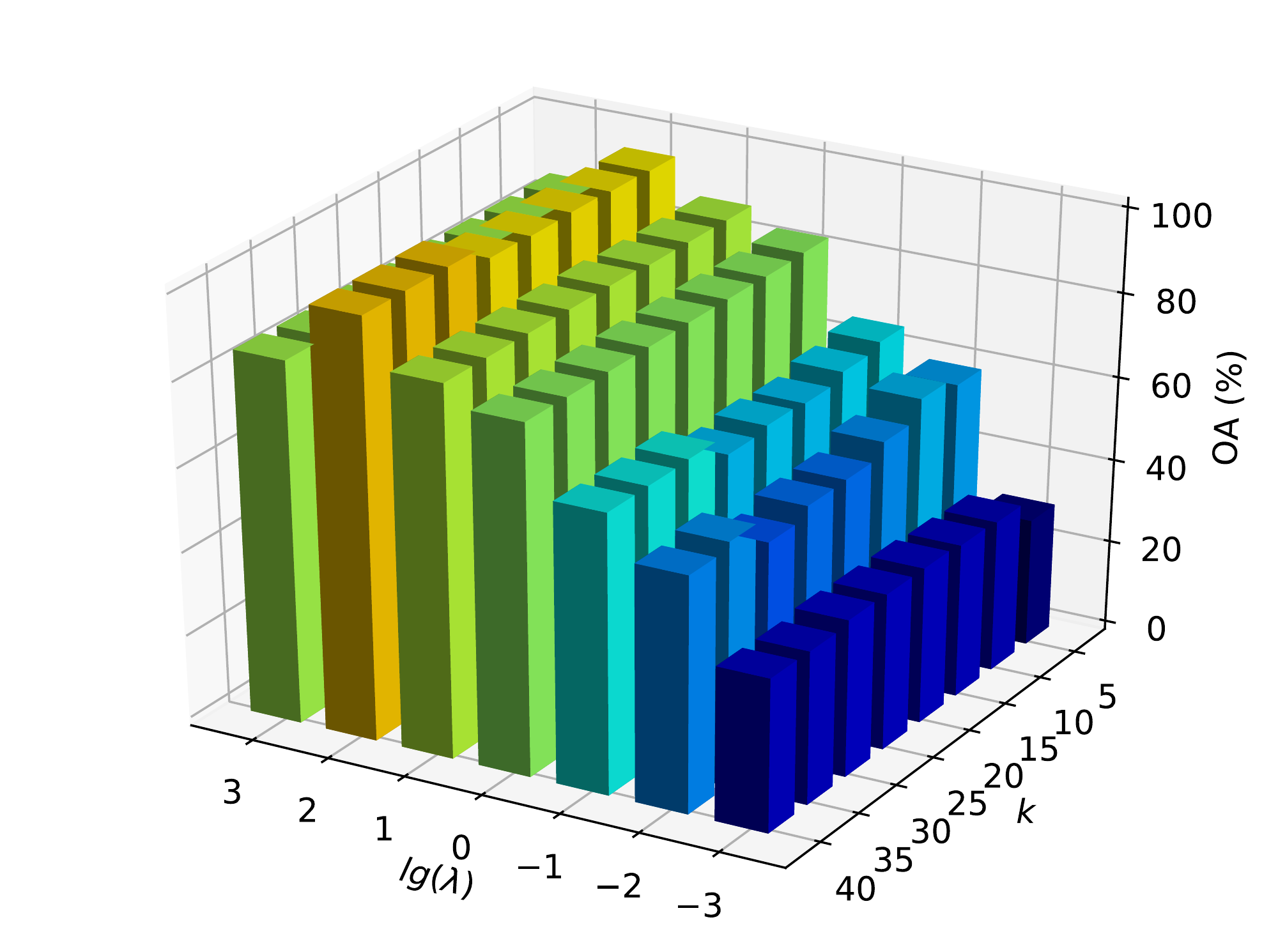}

}
\par\end{centering}
\begin{centering}
\subfloat[EGCSC]{\includegraphics[width=0.5\columnwidth]{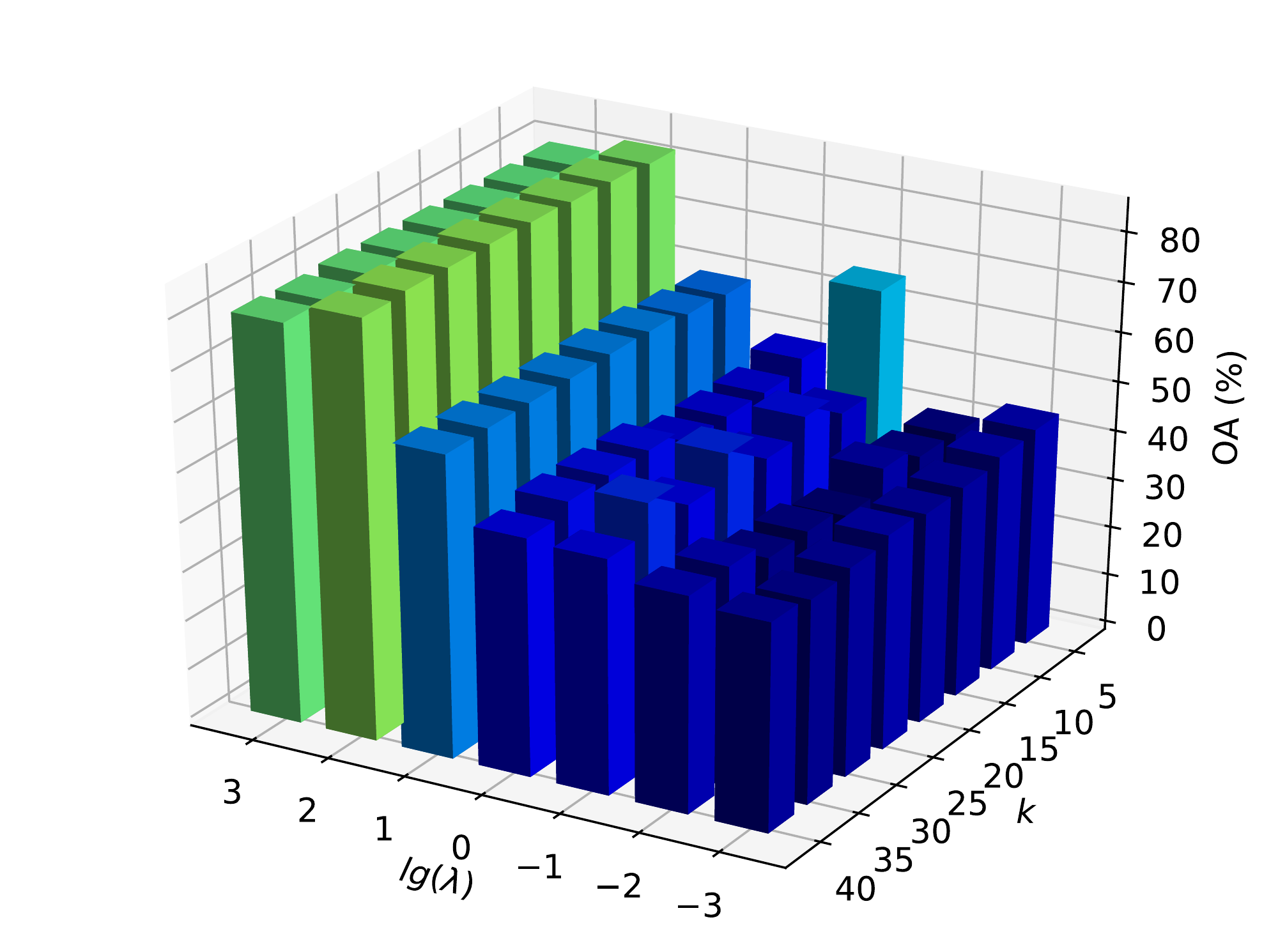}

}\subfloat[EKGCSC]{\includegraphics[width=0.5\columnwidth]{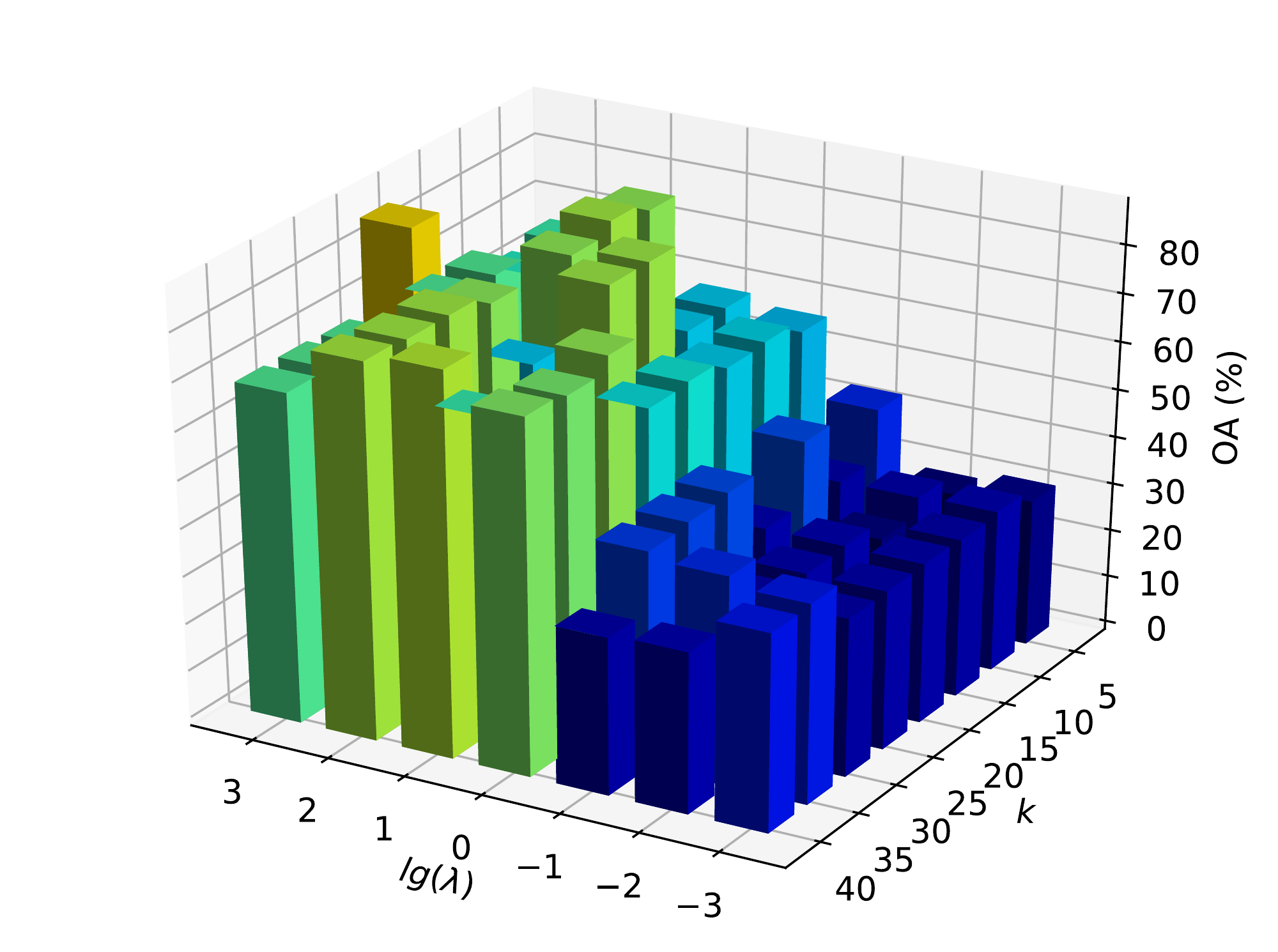}

}
\par\end{centering}
\begin{centering}
\subfloat[EGCSC]{\includegraphics[width=0.5\columnwidth]{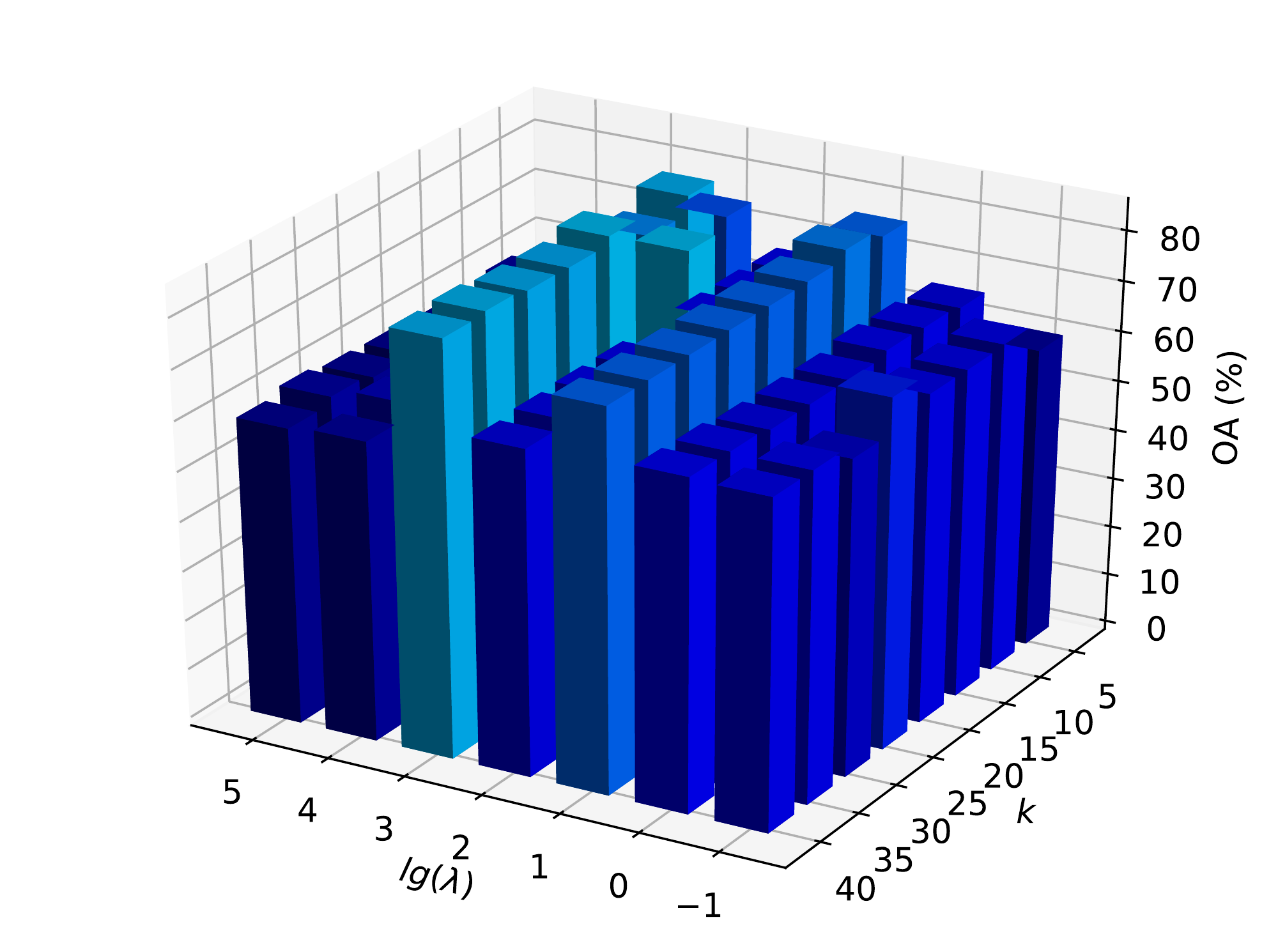}

}\subfloat[EKGCSC]{\includegraphics[width=0.5\columnwidth]{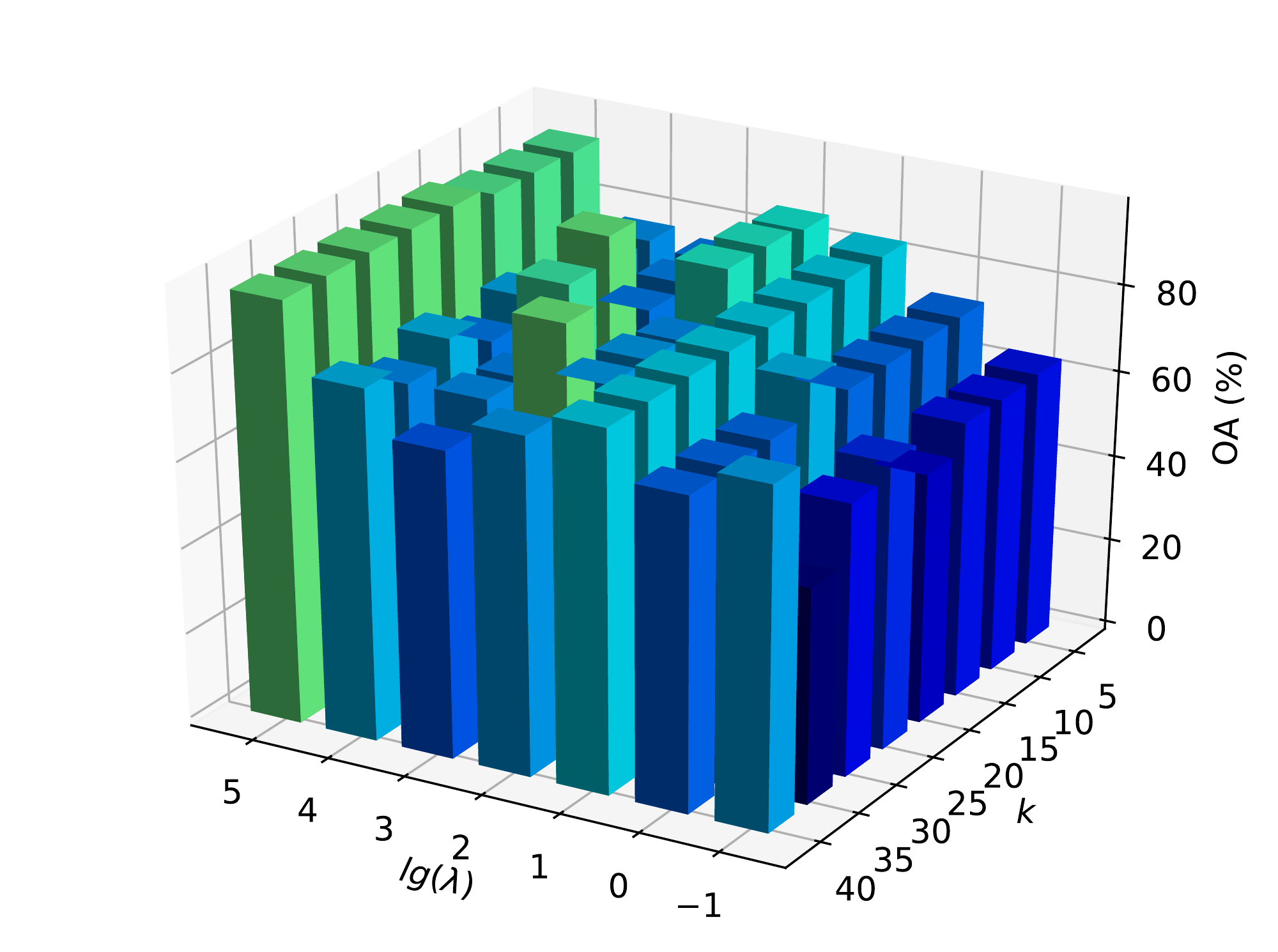}

}
\par\end{centering}
\caption{Influence of $\lambda$ and $k$ for EGCSC and EKGCSC on (a)-(b) SalinasA,
(c)-(d) Indian Pines, and (e)-(f) Pavia University datasets.\label{fig:3d-bar}}
\end{figure}
\par\end{center}

In this experiment, we investigate the impact of the two most important
hyper-parameters involved in the GCSC framework, i.e., the regularization
coefficient $\lambda$ and the number of the nearest neighbors $k$
for the kNN graph. We set $\lambda$ in the range of $\left[10^{-3},10^{3}\right]$
for SalinasA and Indian Pines datasets, and $\left[10^{-1},10^{5}\right]$
for Pavia University dataset. For clarity, we take $lg\left(\lambda\right)$
into account for plotting. For all datasets, we let $k$ vary from
$5$ to $40$ with an interval of $5$. The results are shown in Fig.
\ref{fig:3d-bar}. It can be seen that $\lambda$ has a significant
impact on clustering performance. We can further observe a tendency,
i.e., the clustering performance will increase as $\lambda$ increased.
By contrast, both EGCSC and EKGCSC are insensitive to $k$. However,
when $k$ is too large, graph convolution will result in an over-smoothing
problem. That is, the graph embedding of all the data points will
become similar. Therefore, too large $k$ may negatively affect the
clustering performance. According to the empirical study, we provide
a group of the best hyper-parameter setting in Table \ref{tab:settings}.

\subsubsection{Impact of The Number of PCs}
\begin{center}
\begin{figure*}[tbh]
\begin{centering}
\subfloat[]{\includegraphics[width=0.66\columnwidth]{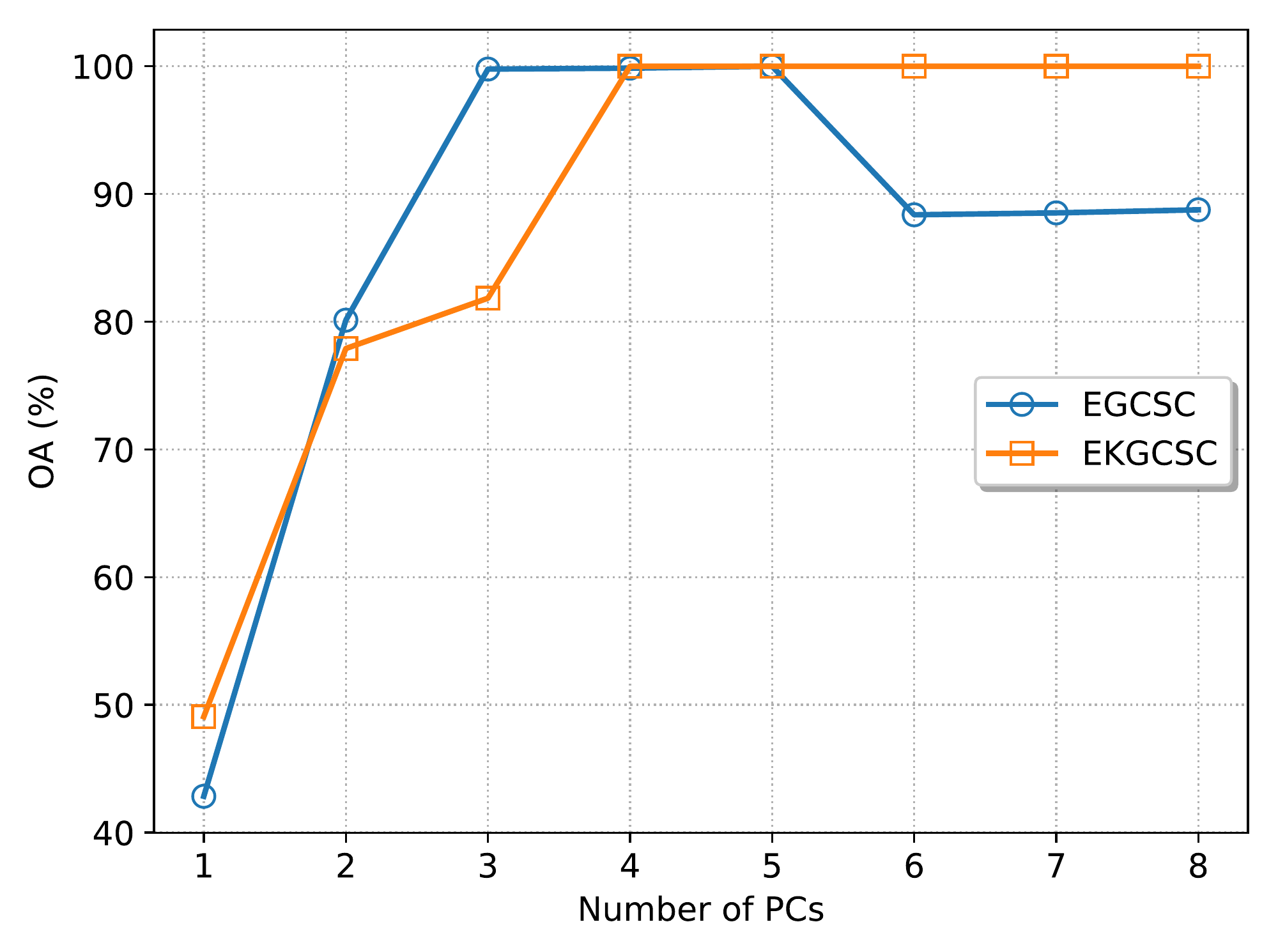}

}\subfloat[]{\includegraphics[width=0.66\columnwidth]{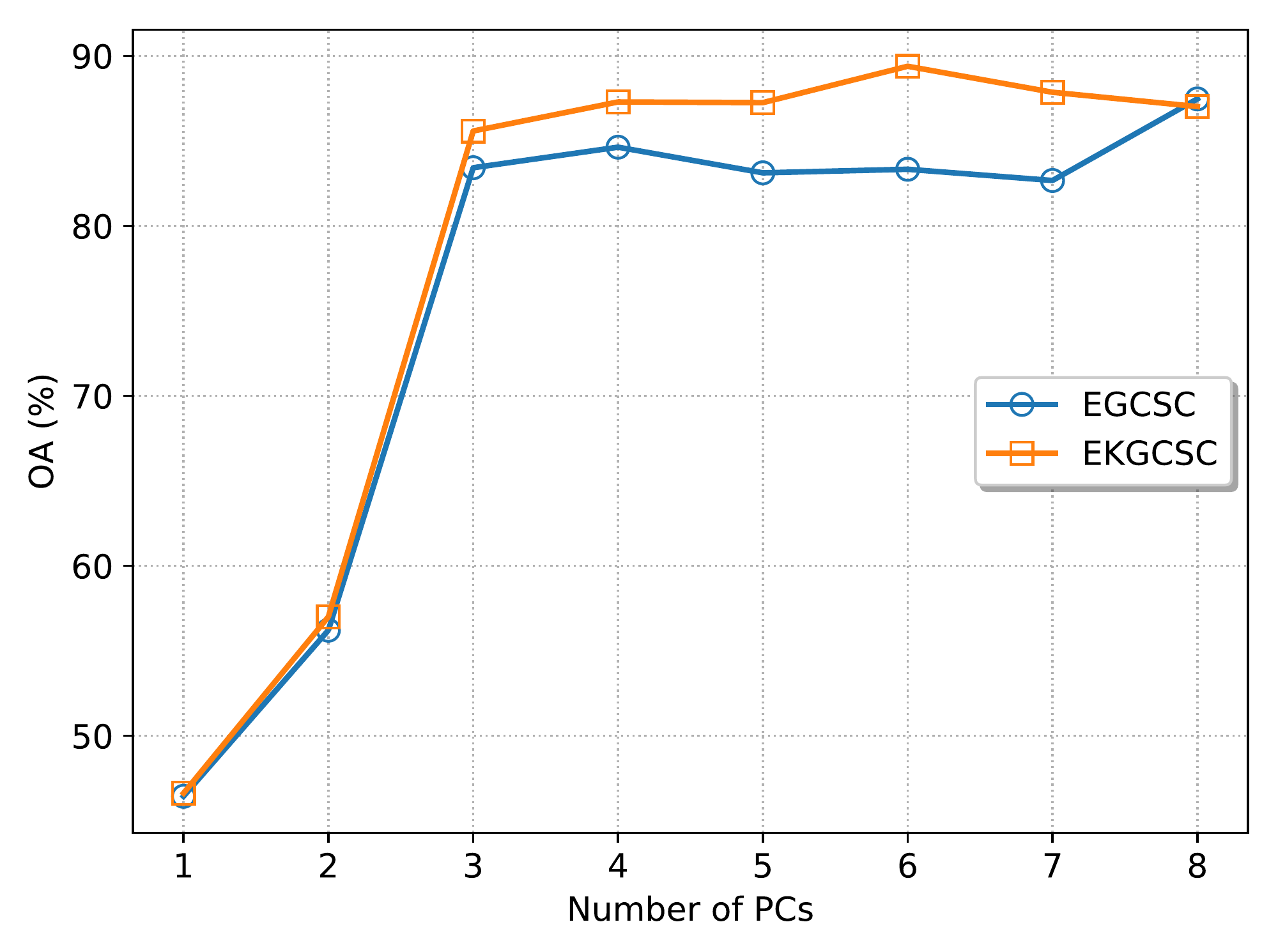}

}\subfloat[]{\includegraphics[width=0.66\columnwidth]{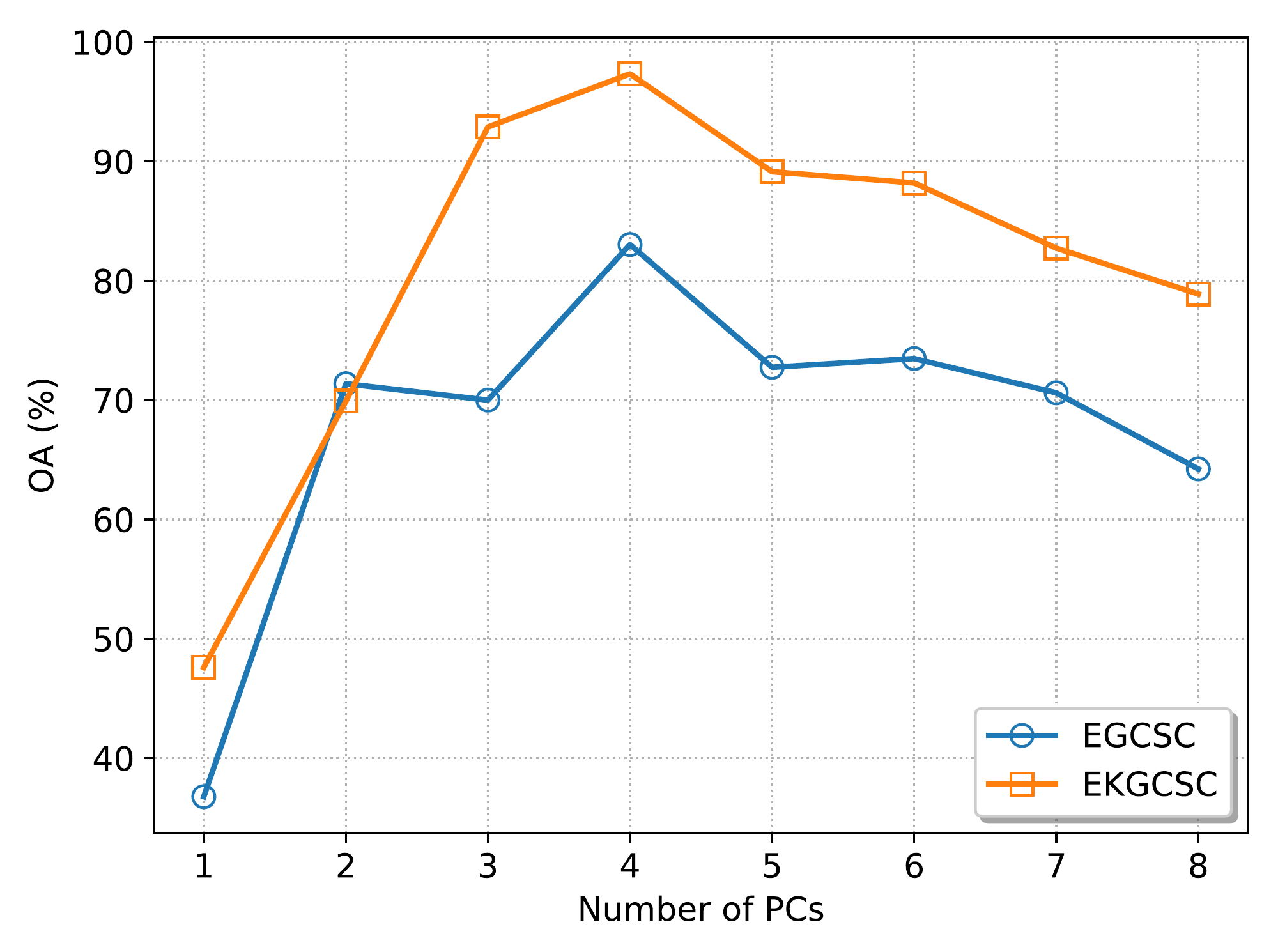}

}
\par\end{centering}
\caption{Clustering OA under varying number of PCs for (a) SalinasA, (b) Indian
Pines, and (c) Pavia University datasets.\label{fig:pc}}

\end{figure*}
\par\end{center}

To empirically study the influence of the number of PCs, we perform
the proposed methods with varying PCs from $1$ to $8$. We show the
results in Fig. \ref{fig:pc}. As shown in the figures, the clustering
performance of the GCSC models increases with PCs, which is because
more spectral information will be included when more PCs are considered.
However, it does not always enhance model performance since more redundancy
might be increased. For example, Fig. \ref{fig:pc} (c) shows the
best number of PCs is $4$ instead of $8$. Although dimensionality
reduction is an optional step in our framework, it achieves a good
balance between the computational efficiency and model performance. 

\subsubsection{Comparison of Running Time}
\begin{center}
\begin{table*}[tbh]
\caption{Running time of different methods (in second). \label{tab:Running-time}}

\centering{}%
\begin{tabular}{ccccccccccc}
\hline 
Data & SC \cite{HSIClustering-SpectralClustering-GRSL-2017} & SSC \cite{SSC-Elhamifar-TPAMI-2013} & LRSC \cite{LRSC-2-ZhuX-TKDE-2019} & $\ell_{2}$-SSC \cite{HSIClustering-L2SSC-ZhanH-GRSL-2017} & S$^{4}$C \cite{HSIClustering-S4C-ZhangH-TGRS-2016} & UBL \cite{HSIClustering-UBL-GRSL-2019} & RMMF \cite{HSIClustering-RMMF-ZhangLF-IS-2019} & EDSC \cite{Efficient-dense-SC-PanJ-WACV-2014} & EGCSC & EKGCSC\tabularnewline
\hline 
\noalign{\vskip0.1cm}
SaA. & 13.203 & 855.663 & 7030.710 & 4.717 & 9363.5 & 2509.68 & 9.448 & 42.762 & 92.951 & 131.485\tabularnewline
\noalign{\vskip0.1cm}
InP. & 8.587 & 653.998 & 3980.004 & 3.272 & 1567.9 & 104.90 & 1.494 & 24.093 & 69.366 & 99.052\tabularnewline
\noalign{\vskip0.1cm}
PaU. & 15.640 & 1022.382 & 15861.621 & 15.677 & 7398.3 & 237.44 & 4.310 & 98.533 & 124.333 & 264.649\tabularnewline
\hline 
\noalign{\vskip0.1cm}
\end{tabular}
\end{table*}
\par\end{center}

We compare our methods with the other competitors in terms of running
time. Table \ref{tab:Running-time} lists the running time of different
clustering methods. Since EGCSC and EKGCSC have closed-form solutions
without needing an iterative operation, they are significantly faster
than SSC, LRSC, S$^{4}$C, and UBL. Although SC, $\ell_{2}$-SSC,
and RMMF take less running time, they cannot achieve better performance
than our methods. Compared with EDSC, both of our methods take relatively
more running time, which is because the proposed GCSC framework needs
to construct the graph from data points. Furthermore, EKGCSC needs
to compute the kernel matrix, thus its running time will be increased
compared with EGCSC. To sum up, our proposed EGCSC and EKGCSC models
achieve a good balance between time cost and clustering accuracy. 

\section{Conclusions \label{sec:Conclusions}}

We have proposed a novel HSI clustering framework, termed as GCSC,
based on introducing graph convolution into subspace clustering. The
key to the proposed framework is to utilize a graph convolutional
self-representation to incorporate the intrinsic structure information
of data points. Traditional subspace clustering models can be treated
as the special forms of the GCSC framework built on the Euclidean
data. Benefiting from the graph convolution, the GCSC model tends
to use a clear dictionary to learn a robust affinity matrix. We design
two efficient subspace clustering models (i.e., EGCSC and EKGCSC)
based on the proposed GCSC framework by using the Frobenius norm.
The experimental results on three HSI data sets demonstrate that the
proposed GCSC models can achieve state-of-the-art performance with
significant margins compared with many existing clustering models.
Particularly, the EKGCSC model achieves $100\%$, $87.61\%$, and
$97.36\%$ clustering OA on SalinasA, Indian Pines, and Pavia University
datasets, respectively.

The successful attempt of the GCSC model signifies that considering
the intrinsic graph structure among data set is important for clustering,
which offers an alternative orientation for unsupervised learning.
The proposed GCSC framework also enables us to revisit traditional
clustering models in the non-Euclidean domain. There are many promising
ways to improve the GCSC model. For example, one can consider deep
graph embedding into GCSC. These issues will be further studied in
our future works. 

\section{Appendix\label{sec:Appendix}}
\begin{IEEEproof}
The solution of EGCSC.

Let $\mathcal{L}$ be the loss function of EGCSC and Eq. \eqref{eq:gcsc}
can be rewritten as 

\begin{equation}
\begin{aligned}\mathcal{L}\left(\mathbf{Z}\right)= & \frac{1}{2}\left\Vert \mathbf{X}\bar{\mathbf{A}}\mathbf{Z}-\mathbf{X}\right\Vert _{F}^{2}+\frac{\lambda}{2}\left\Vert \mathbf{Z}\right\Vert _{F}^{2}\\
= & \frac{1}{2}tr\left[\left(\mathbf{X}\bar{\mathbf{A}}\mathbf{Z}-\mathbf{X}\right)^{T}\left(\mathbf{X}\bar{\mathbf{A}}\mathbf{Z}-\mathbf{X}\right)+\lambda\mathbf{Z}^{T}\mathbf{Z}\right]\\
= & \frac{1}{2}tr\left(\mathbf{Z}^{T}\bar{\mathbf{A}}^{T}\mathbf{X}^{T}\mathbf{X}\bar{\mathbf{A}}\mathbf{Z}+\mathbf{X}^{T}\mathbf{X}-2\mathbf{X}^{T}\mathbf{X}\bar{\mathbf{A}}\mathbf{Z}+\lambda\mathbf{Z}^{T}\mathbf{Z}\right)
\end{aligned}
.\label{eq:eq1}
\end{equation}
According to the properties of matrix trace and matrix derivatives,
the partial derivative of $\mathcal{L}$ with respect to $\mathbf{Z}$
can be presented as

\begin{equation}
\begin{aligned}\frac{\partial\mathcal{L}}{\partial\mathbf{Z}}= & \bar{\mathbf{A}}^{T}\mathbf{X}^{T}\mathbf{X}\bar{\mathbf{A}}\mathbf{Z}-\bar{\mathbf{A}}^{T}\mathbf{X}^{T}\mathbf{X}+\lambda\mathbf{Z}\\
= & \left(\bar{\mathbf{A}}^{T}\mathbf{X}^{T}\mathbf{X}\bar{\mathbf{A}}+\lambda\mathbf{I}_{N}\right)\mathbf{Z}-\bar{\mathbf{A}}^{T}\mathbf{X}^{T}\mathbf{X}
\end{aligned}
.\label{eq:eq2}
\end{equation}
Let $\frac{\partial\mathcal{L}}{\partial\mathbf{Z}}=0$, we get 

\begin{equation}
\left(\bar{\mathbf{A}}^{T}\mathbf{X}^{T}\mathbf{X}\bar{\mathbf{A}}+\lambda\mathbf{I}_{N}\right)\mathbf{Z}=\bar{\mathbf{A}}^{T}\mathbf{X}^{T}\mathbf{X}.
\end{equation}
Finally, $\mathbf{Z}$ can be expressed as 

\begin{equation}
\mathbf{Z}=\left(\bar{\mathbf{A}}^{T}\mathbf{X}^{T}\mathbf{X}\bar{\mathbf{A}}+\lambda\mathbf{I}_{N}\right)^{-1}\bar{\mathbf{A}}^{T}\mathbf{X}^{T}\mathbf{X}.
\end{equation}
Due to $\bar{\mathbf{A}}^{T}\mathbf{X}^{T}\mathbf{X}\bar{\mathbf{A}}+\lambda\mathbf{I}$
is positive semidefinite, its reversible always exists.
\end{IEEEproof}
\begin{IEEEproof}
The solution of EKGCSC.

Similar to the above proof, the loss function of EKGCSC (Eq. \eqref{eq:kgcsc})
can be expressed as 

\begin{equation}
\begin{aligned}\mathcal{L}\left(\mathbf{Z}\right)= & \frac{1}{2}tr(\mathbf{Z}^{T}\bar{\mathbf{A}}^{T}\mathbf{K}_{\mathbf{XX}}\bar{\mathbf{A}}\mathbf{Z}-2\mathbf{K}_{\mathbf{XX}}\bar{\mathbf{A}}\mathbf{Z}+\\
 & \mathbf{K}_{\mathbf{XX}}+\lambda\mathbf{Z}^{T}\mathbf{Z})
\end{aligned}
.
\end{equation}
The partial derivative of $\mathcal{L}$ with respect to $\mathbf{Z}$
is then given by 

\begin{equation}
\begin{aligned}\frac{\partial\mathcal{L}}{\partial\mathbf{Z}}= & \bar{\mathbf{A}}^{T}\mathbf{K}_{\mathbf{XX}}\bar{\mathbf{A}}\mathbf{Z}-\bar{\mathbf{A}}^{T}\mathbf{K}_{\mathbf{XX}}+\lambda\mathbf{Z}\\
= & \left(\bar{\mathbf{A}}^{T}\mathbf{K}_{\mathbf{XX}}\bar{\mathbf{A}}\mathbf{Z}+\lambda\mathbf{I}_{N}\right)\mathbf{Z}-\bar{\mathbf{A}}^{T}\mathbf{K}_{\mathbf{XX}}
\end{aligned}
.
\end{equation}
By setting $\frac{\partial\mathcal{L}}{\partial\mathbf{Z}}=0$, we
finally get the optimal solution of $\mathbf{Z}$ as follows: 

\begin{equation}
\mathbf{Z}=\left(\bar{\mathbf{A}}^{T}\mathbf{K}_{\mathbf{XX}}\bar{\mathbf{A}}+\lambda\mathbf{I}_{N}\right)^{-1}\bar{\mathbf{A}}^{T}\mathbf{K}_{\mathbf{XX}}
\end{equation}
\end{IEEEproof}

\section*{Acknowlegment}

The authors would like to thank the anonymous reviewers for their
constructive suggestions and criticisms. We would also like to thank
Prof. Lefei Zhang who provided the source codes of the RMMF algorithm. 

\bibliographystyle{IEEEtran}
\bibliography{Ref-A}

\end{document}